\documentclass{article}

\usepackage{arxiv}

\usepackage[utf8]{inputenc} 
\usepackage[T1]{fontenc}    
\usepackage{hyperref}       
\usepackage{url}            
\usepackage{booktabs}       
\usepackage{amsfonts}       
\usepackage{nicefrac}       
\usepackage{microtype}      
\usepackage{lipsum}
\usepackage{graphicx}
\usepackage{caption}
\usepackage{subcaption}
\usepackage{graphicx}
\usepackage{float}
\usepackage{multirow}
\usepackage[font=small,labelfont=bf,justification=justified]{caption}
\usepackage{booktabs}
\usepackage{multirow}
\usepackage{float}
\usepackage{xcolor}
\usepackage{amsmath}
\usepackage{subcaption}
\usepackage{cite}
\usepackage[export]{adjustbox}
\usepackage{amsmath,amssymb,amsfonts}
\usepackage{soul}
\usepackage[normalem]{ulem}

\newcommand*{\duy}{\textcolor{blue}}

\title{High Accurate and Explainable Multi-Pill Detection Framework with Graph Neural Network-Assisted Multimodal Data Fusion}

\author{
 Anh Duy Nguyen \\
  VinUni-Illinois  Smart Health Center, \\ VinUniversity, Vietnam \\
  \texttt{duy.na@vinuni.edu.vn} \\
   \And
 Huy Hieu Pham \\
 College of Engineering \& Computer Science, \\ VinUni-Illinois Smart Health Center, VinUniversity, Vietnam\\
  \texttt{hieu.ph@vinuni.edu.vn} \\
  \And
 Huynh Thanh Trung \\
  École Polytechnique Fédérale de Lausanne,  \\ Switzerland \\
  \texttt{thanh.huynh@epfl.ch} \\
  \And
Quoc Viet Hung Nguyen \\
  Griffith University, \\ Australia  \\
  \texttt{henry.nguyen@griffith.edu.au} \\
  \And
 Thao Nguyen Truong \\
 National Institute of Advanced Industrial Science \\ and Technology, Japan \\
  \texttt{nguyen.truong@aist.go.jp} \\
  \And
 Phi Le Nguyen \\
  School of Information and Communication Technology,  \\ Hanoi University of Science
and Technology, Vietnam\\
  \texttt{lenp@soict.hust.edu.vn}
}

\begin{document}
\maketitle
\begin{abstract}
Due to the significant resemblance in visual appearance, pill misuse is prevalent and has become a critical issue, responsible for one-third of all deaths worldwide. Pill identification, thus, is a crucial concern needed to be investigated thoroughly. 
Recently, several attempts have been made to exploit deep learning to tackle the pill identification problem. 
However, most published works consider only single-pill identification and fail to distinguish hard samples with identical appearances. 
Also, most existing pill image datasets only feature single pill images captured in carefully controlled environments under ideal lighting conditions and clean backgrounds.
In this work, we are the first to tackle the multi-pill detection problem in real-world settings, aiming at localizing and identifying pills captured by users in a pill intake. Moreover, we also introduce a multi-pill image dataset taken in unconstrained conditions. 
To handle hard samples, we propose a novel method for constructing heterogeneous a priori graphs incorporating three forms of inter-pill relationships, including co-occurrence likelihood, relative size, and visual semantic correlation. 
We then offer a framework for integrating a priori with pills' visual features to enhance detection accuracy. 
Our experimental results have proved the robustness, reliability, and explainability of the proposed framework. Experimentally, it outperforms all detection benchmarks in terms of all evaluation metrics. 
Specifically, our proposed framework improves COCO mAP metrics by $\mathbf{9.4\%}$ over Faster R-CNN, and $\mathbf{12.0\%}$ compared to vanilla YOLOv5. Our study opens up new opportunities for protecting patients from medication errors using an AI-based pill identification solution.  
\end{abstract}


\section{Introduction}
\label{sec:introduction}
\noindent \textbf{Background.} 
{Oral pill is one of the most popular and commonly used methods in healthcare due to their efficacy and simplicity.
Pills usually exhibit various visual features in terms of shape, color, and imprinted text.
Despite this, erroneously taking pills is exceptionally prevalent due to the significant similarity in pill appearances. According to a WHO report\cite{who}, drug misuse rather than illness is responsible for one-third of all deaths. Moreover, according to Yaniv et al.~\cite{yaniv}, around six to eight thousand people are killed annually by prescription errors. Recently, the US National Centers for Biomedical Computing (NCBCs) stated that taking this country alone, each year, $7,000$ to $9,000$ people die due to a medication error. This circumstance necessitates the invention of solutions to protect users/patients from taking incorrect pills. This need is more stringent than ever, given the aging of the global population and the rising prevalence of chronic diseases requiring continuous medication.}

\noindent \textbf{Image-based pill recognition.} 
{In the early stage, pill recognition was handled through a variety of online systems that allow users to identify pills by manually entering multiple attributes, such as shape, color, and imprinted text \cite{PillIdentifier1}. However, these methods are time-consuming and may not be reliable, as the predefined features may not encompass all real-world cases. Recently, Artificial Intelligence (AI) has made tremendous achievements and has emerged as a powerful tool in resolving various problems.
Although still in its infancy, AI has been used to recognize pills from images, helping prevent incorrect medication. An early effort to classify pills using a Deep Convolution Network (DCN) was introduced in \cite{wong2017development}. 
In \cite{usuyama2020epillid}, the authors provided ePillID, a large pill image dataset comprising $13K$ images representing $8,184$ appearance classes.
Additionally, they conducted experiments to evaluate various baseline models on the proposed dataset.
Even with the best baseline, the experimental findings demonstrated that it fails to discriminate confusing classes.
The problem of few-shot pill recognition was addressed in \cite{ling2020few}. 
The authors also provided a new pill image data named CURE. 
Recently, there have been a few works considering the multi-pill detection problem \cite{kwon_pill}. The authors adopted the two-step deep neural networks consisting of an object localization and a classifier. }

\begin{figure}[bt]
    \centering
    \subfloat[Pills with similar shapes, colors, and different sizes]{
    \includegraphics[width=0.45\columnwidth]{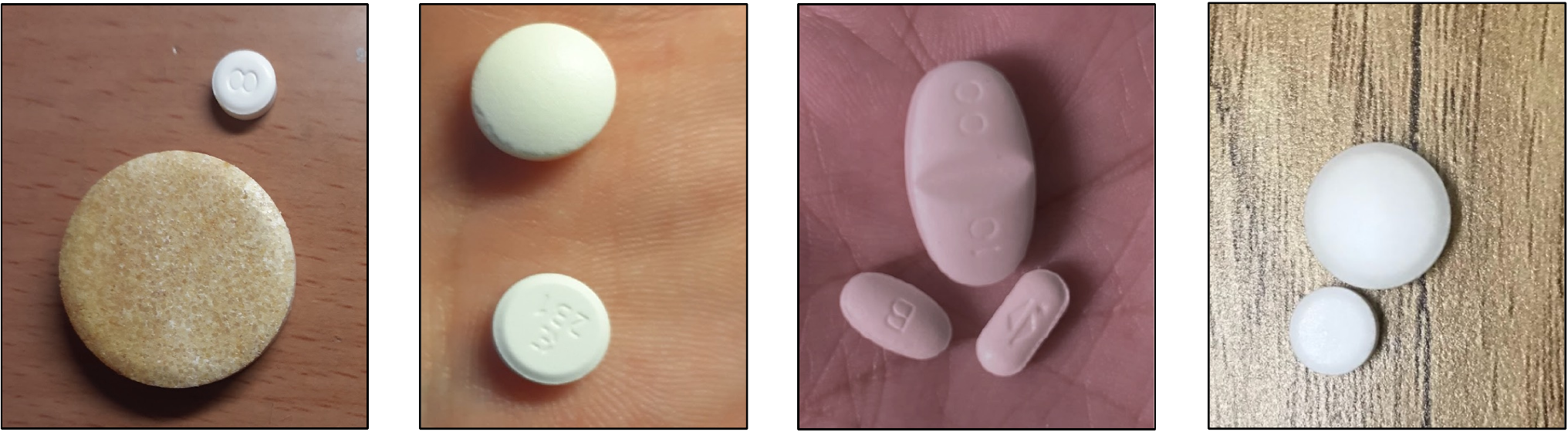}
     \label{fig:hard_case_diff_size}
    }
    \hfill
    \subfloat[Pills with similar shapes, sizes and colors]{
    \label{fig:hard_case_all} 
    \includegraphics[width=0.45\columnwidth]{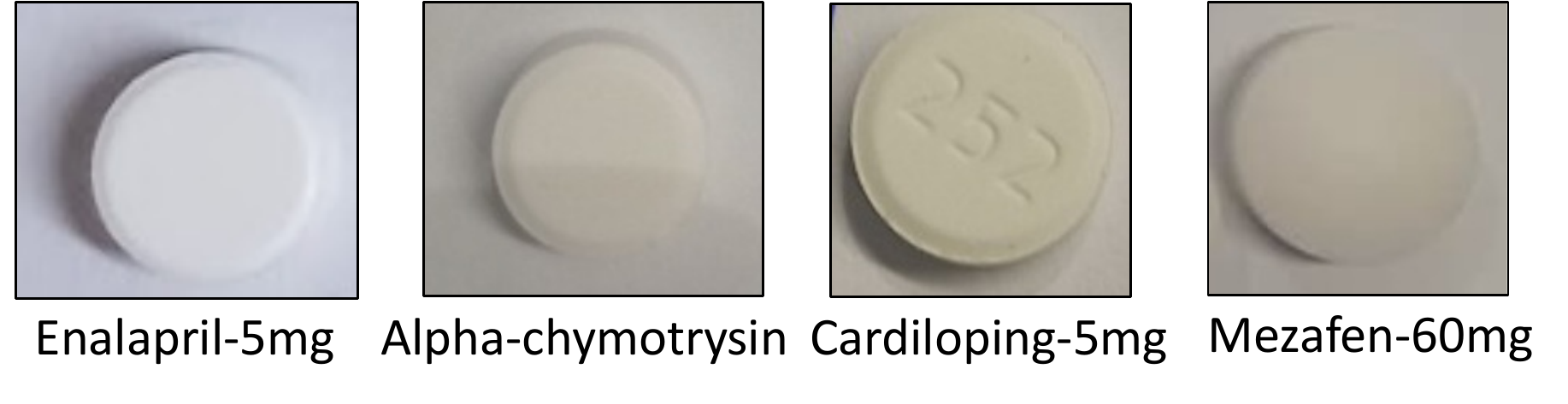}
    }
    \caption{\textbf{Hard samples with high similarity in terms of shape, color and size (examples taken from our handcrafted dataset).} The existence of hard examples has rendered the pill identification problem complicated and challenging.\label{fig:hard_case} }
    \vspace{-15pt}
\end{figure}

\noindent  \textbf{Problem statement.} Despite several efforts that have been made, existing solutions for pill identification reveal the following critical shortcomings. 
\begin{itemize}
    \item Most existing works have been restricted to the classification of single-pill images. This constraint limits the solutions' application capacity, as in practice, users usually take multiple pills simultaneously, resulting in multi-pill images in most cases. 
    \item Most of the current pill image datasets (e.g., ePillID, CURE) are limited to single-pill images. Moreover, all of them were collected in tightly-controlled settings under ideal illumination and background conditions, leading to the lack of diversity.
    \item No prior work has studied the explainability of the model. This insufficiency diminishes the trustworthiness of the solutions, hence restricting their practical applications.
\end{itemize} 
We are, to the best of our knowledge, the first to tackle the multi-pill detection problem in real-world settings.
Specifically, we focus on a practical application that recognizes pills in patients’ pill intake pictures. Our targeted problem can be formulated as follows. 
\emph{Given an image capturing multiple pills in patients' pill intake, we aim to determine each pill's location and identity. }
In addition to developing a novel pill detection framework with high reliability and explainable capacity, we build a dataset of multi-pill images captured under unconstrained real-world conditions. 

\begin{figure}[bt]
    \centering
    \includegraphics[width=0.8\columnwidth]{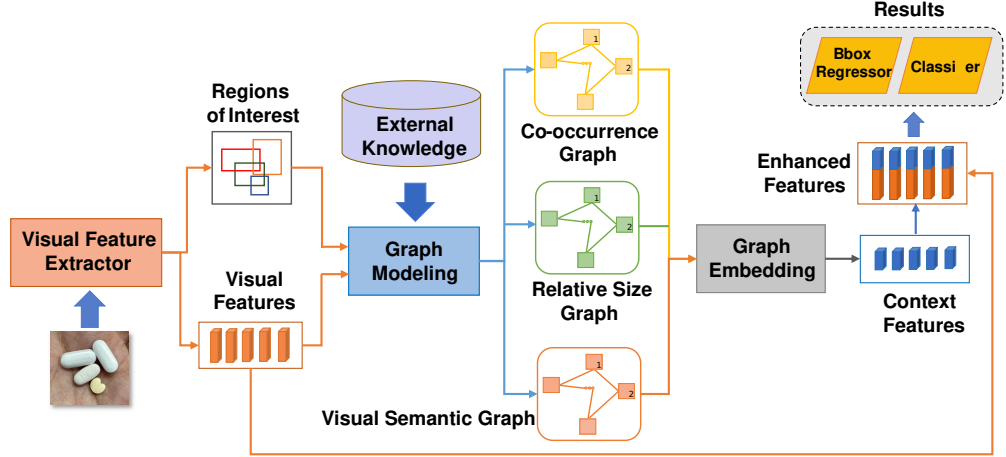}
    \caption{\textbf{Overview of our proposed solution.}
    We leverage external knowledge from prescriptions and training datasets to build the co-occurrence and relative size graph. The visual features of the pills are exploited to construct the visual semantic graph. Using the graph embedding module, the three graphs are transformed into the vector space, then fused with the visual features to provide enhanced feature vectors, which are then utilized to create the final results.
    \label{fig:intro_overview}}
    \vspace{-15pt}
\end{figure}

\noindent \textbf{Our motivation and key ideas.} 
{One of the most significant obstacles in the pill detection problem is the existence of numerous pills with similar shapes, colors, and sizes (Fig.~\ref{fig:hard_case}).
We call these hard samples whose occurrence renders the pill identification problem complicated and challenging to solve by generic object detection.
We argue that relying merely on pills' appearance is insufficient to improve pill detection accuracy, if not possible.
We discovered that besides the challenge (e.g., localizing pills in hard cases such as overlapping pills), the multi-pill detection problem, on the other hand, provides us with an opportunity to improve the pill recognition accuracy. 
Motivated by the human tendency and ability to integrate different data sources while making decisions, our proposed solution seeks to utilize external knowledge to improve detection accuracy.
Specifically, we provide a priori graphs incorporating three kinds of pill relationships: co-occurrence, relative size, and visual semantic correlation.
The first a priori, or co-occurrence graph, demonstrates the frequency with which medications are prescribed for the same diseases; thus, it reflects the likelihood that pills appear in the same image. 
By utilizing this knowledge, we can enhance the accuracy in dealing with hard samples by leveraging the high accuracy in detecting easy samples in the same image.
The second one, i.e., the relative size graph, gives us the relative size information of the pills, thus, improving our model's capacity to distinguish pills of identical shape and color but differing only in size.
Finally, the visual semantic graph learns pills' latent semantic link embedded in pills' visual appearance.
The overview of our proposed model is illustrated in Fig.\ref{fig:intro_overview}.
}
{
\noindent \textbf{Our contributions.} Our main contributions are as follows.
\begin{itemize}
    \item We introduce the first real-world multi-pill image dataset consisting of $9,426$ images representing $96$ pill classes. The images were taken with ordinary smartphones in various settings.
    The dataset will assist in the advancement of research in the field.
    \item We propose a novel pill detection framework named PGPNet (which stands for a Priori Graph-assisted Pill detection Network), which leverages three-fold graph-based a priori, including co-occurrence likelihood, relative pill size, and visual semantic correlation to tackle hard pill samples. 
    In addition, we provide a method for constructing these heterogeneous a priori graphs from given prescriptions and the training pill image dataset. Furthermore, we offer a multi-modal fusion method for incorporating graph-based inter-pill relational information with intra-pill visual features to enhance the detection result. 
    \item We conduct thorough experiments to evaluate the efficacy of the proposed solution and compare it to existing state-of-the-art (SOTA). The experimental findings demonstrated that our approach enhances the object detection accuracy by at least $9.4\%$ for COCO mAP metric compared to generic SOTA in object detection. 
\end{itemize}

The remainder of the paper is divided into four sections. We briefly summarize the literature on pill detection and pill image datasets in Section ~\nameref{sec:related_work}. In Section \nameref{sec:method}, we describe our methodology in detail.
Section \nameref{sec:result} evaluates the performance of our proposed PGPNet and compares it with the other methods. 
Finally, we conclude the paper in Section \nameref{sec:conclusion}.}

\section*{Related Works}
\label{sec:related_work}

\textbf{Pill Classification.}
Many studies have employed machine learning to tackle the pill recognition challenge~\cite{WONG2017130,8962044}. 
The authors in~\cite{WONG2017130} first utilized the Manifold ranking-based method to filter out the foreground mask from the input pill image and then used an AlexNet-based network for identifying the label. In~\cite{Ting17}, Ting et al. combined the Enhanced Feature Pyramid Networks and Global Convolution Networks to improve pill localization accuracy. 
Ling {et al.}~\cite{Ling_2020_CVPR} tackled the few-shot pill detection problem with a Multi-Stream (MS) deep learning model. 
In~\cite{Proma19}, the authors integrated three handcrafted features, namely shape, color, and imprinted text, to identify pills. 

Recently, a few efforts have leveraged the two-stage object detection approach to solve the multi-pill detection challenge~\cite{Ou2020, kwon_pill}. 
In the first stage, object localization techniques are applied to determine the pills' bounding boxes. 
These bounding boxes are then fed into a classifier in the second stage to identify the pills.
Specifically, in \cite{Ou2020}, an enhanced feature pyramid network based on the ResNet-50 backbone has been built for pill localization. 
After that, the pill bounding boxes are fed into an Inception-ResNet v2 for classification. 
Authors in \cite{kwon_pill} exploited the Mask-RCNN framework to solve the problem.

Multi-pill detection solutions are still in their infancy.
All current works only investigate images acquired in laboratories under optimal lighting and background conditions, with each pill arranged separately.
In fact, existing techniques only use specific object localization models to crop the pills and then treat the issue as a typical single-pill classiﬁcation problem.

{\textbf{Pill Image Datasets.} One of the most widely used pill image datasets is NIH Pill Image Dataset \cite{nih_dataset} released by the U.S. National Library of
Medicine (NLM). This dataset consists of 4,000 high-quality reference pills and 133,000 pictures captured by digital cameras on mobile phones.
In \cite{ling2020few}, the authors provided the CURE pill dataset consisting of 8,973 single-pill images representing 196 classes. 
Although taken under various backgrounds and lighting conditions, all of these images are carefully captured from a top-down view and focus on the pills.
Authors in \cite{wong2017development} contributed a pill dataset capturing about 400 commonly used tablets and capsules.
Ten to twenty-five pictures were taken for each pill, resulting in 5,284 images. 

Unfortunately, all of these datasets provide only single-pill images.
Most images were captured under quite ideal conditions, e.g., pills were put on a clean background, and the images were taken from the top-down view with the camera focused on the pills.} 

\section*{Methodology}
\label{sec:method}

\begin{figure*}[bth]
    \centering
    \includegraphics[width=1\textwidth]{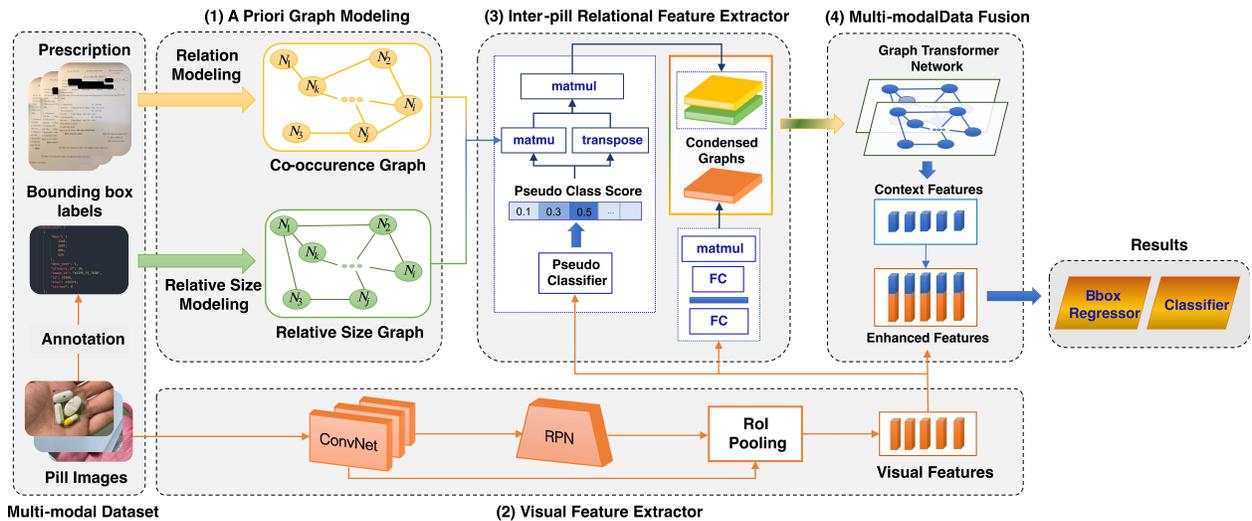}
    \caption{ \textbf{Workflow of PGPNet.}
    First, the \textbf{A Priori Graph Modeling} Module leverages given prescriptions to create a non-directed Medical Co-occurrence Graph $\mathcal{G}_c$ and then leverages $\mathcal{G}_c$ in conjunction with bounding box annotation information from the pill image dataset to construct a Relative Size Graph $\mathcal{G}_s$.
    Second, the pill images are passed through the \textbf{Visual Feature Extractor} to determine Regions of Interest and retrieve their visual representations. This information, together with the graphs serve as the input for the \textbf{Inter-Pill Relational Feature Extractor} to generate a heterogeneous relational graph expressing three types of relationships between pills, namely, co-occurrence, relative size, and visual semantic relation.
    Finally, the heterogeneous graph-based information and visual features are combined by \textbf{Multi-modal Data Fusion} module to form final enhanced vectors, which will be used to determine the final detection results.\label{fig:generalflow}.
    \vspace{-15pt}
     }
\end{figure*}

In this section, we propose a novel pill detection framework named PGPNet (i.e., a \textbf{P}riori \textbf{G}raph-assisted \textbf{P}ill Detection \textbf{Net}work). 

\subsection*{PGPNet Overview}
\label{sec:method_overview}

{
We focus on a practical application that recognizes pills in
patient intake pictures. Our model receives a multiple-pill picture as input and generates both the bounding box and the identification of each pill. Here, we incur a critical challenge: \emph{how to distinguish pills with identical appearances (i.e., shape, color, and size)}. We believe that relying solely on the visual features of pills is insufficient to address this
issue. Moreover, employing the correlation between pills, rather than counting on each pill individually, may enhance recognition accuracy. In light of this, we propose introducing two types of a priori, the first indicating the co-occurrence likelihood and the second modeling the relative size of pills. The a-priori is extracted from a given
prescription and pill image training dataset and represented
as heterogeneous graphs. In summary, the proposed model
comprises four components: \textbf{A priori graph modeling}, \textbf{visual
feature extractor}, \textbf{inter-pill relational feature extractor}, and
\textbf{multi-modal data fusion}, as illustrated in Fig.\ref{fig:generalflow}. The overall
flow is as follows.
}

\begin{itemize}
    \item [$\bullet$] \textbf{Step 1 - A Priori Graph Modeling}.
    {We construct two generic graphs, namely \emph{Prescription-based Medical Co-occurrence Graph} (or Co-graph for
short) and \emph{Relative Size Graph} (Size-graph for short) that represent
the relationship between all the pills in terms of co-
occurrence and relative size, respectively. Concerning the
former, we leverage a given set of prescriptions from
which we can model the interaction between pills (i.e.,
which pills are likely to be used to treat the same diseases).
Based on this information, we developed the Co-graph, whose nodes represent the pill classes and whose
edge weights reflect the co-occurrence likelihood between
the two vertices. In the meantime, using the coordinates
of the bounding boxes from our training dataset for the
pill detection task, we determine the area of each box
and model the relative size ratios of all the pill classes in
the given images. This information is then aggregated to
formulate the Size-graph. Section~\nameref{sec:method_graph_model} covers the details of this algorithm.}
    \item [$\bullet$] \textbf{Step 2 - Visual Feature Extraction}. {The original image containing multiple pills is passed through a Convolutional Network (ConvNet) for extracting visual features and a Region Proposal Network (RPN) for detecting potential Regions of Interest (RoI). The outputs of the two modules are fed into an RoI pooling layer to filter out all visual presentations of pills (i.e., RoIs). It is worth noting that the \textit{Visual Feature Extractor} described here follows the architecture of the two-step object detection architecture (e.g., Faster RCNN \cite{fasterrcnn}). However, PGPNet can also be implemented with one-step detection architecture.}
    \item [$\bullet$] \textbf{Step 3 - Inter-pill Relational Feature Extraction}.
    {The two a priori graphs are aggregated with the pills' visual features to yield condensed versions of the Co-graph and Size-graph that highlight the relationship between only those pills that are likely to appear in the image. Besides, the pills' visual features are leveraged to construct a so-called \emph{Visual semantic graph} that captures the pills' relationships encapsulated under their appearances.}
    \item [$\bullet$] \textbf{Step 4 - Multi-modal Data Fusion}.
    {Now, the inter-pill relational and intra-pill visual features are fused to obtain enhanced feature vectors, each of which encapsulates the characteristics of a pill standalone and its relationship with other pills. These enhanced feature vectors are used to offer the final results.}
\end{itemize}

\subsection*{A Priori Graph Modeling}
\label{sec:method_graph_model}
{In this section, we describe our method to construct the two generic graphs, namely the co-occurrence Graph (i.e., Co-graph) and relative size graph (i.e., Size-graph) in Sections \nameref{subsec:method_coocurent_modeling} and \nameref{subsec:method_size_modeling}, relatively.}
\subsubsection*{Prescription-based Co-occurrence Graph Modeling}
\label{subsec:method_coocurent_modeling}
{We propose to leverage an external source, namely prescriptions, to build the co-occurrence graph. 
The rationale behind our idea is that as most pills are intended to cure or alleviate certain diseases or symptoms, there is a significant likelihood that pills meant to treat the same diseases will appear concurrently.
Thus, the implicit relationship between the pills can be modeled by assessing the direct interaction between medications and diseases derived through prescriptions.
Our Co-graph, $\mathcal{G}_c = \left \langle V, E, W_c \right \rangle$, is a weighted graph whose vertices $V$ represent pill classes, and whose edges' weights $W_c$ reflect the co-occurrence likelihood of the pills. 
As the association between pills do not explicitly present in the prescriptions, we model this relationship utilizing the interaction between medications and diseases using the following criteria.}
\begin{itemize}
    \item [$\bullet$] There is an edge between two pill classes $C_i$ and $C_j$ if and only if they have been prescribed for at least one shared disease.
    \item [$\bullet$] The greater the weight of an edge $E_{ij}$ connecting pill classes $C_i$ and $C_j$, the more likely that these two medications will be prescribed simultaneously.
\end{itemize}
{We first define a so-called \texttt{Diagnose-Pill} impact factor, which reflects how important a pill is to a diagnosis.
Inspired by the Term Frequency (\texttt{tf}) — Inverse Dense Frequency (\texttt{idf}) often used in the Natural Language Processing domain, we define the impact factor of a pill $P_j$ to a diagnosis $D_i$, denoted as $\mathcal{I}(P_j, D_i)$, as follows}
\begin{equation*}
    \small
    \mathcal{I}(P_j, D_i) = \texttt{tf}(D_j, P_i)\times \texttt{idf}(P_i) = \frac{|\mathbb{S}(D_j, P_i)|}{|\mathbb{S}(D_j)|} \times \log \frac{|\mathbb{S}|}{|\mathbb{S}(P_i)|},
\end{equation*}
where $\mathbb{S}$ represents the set of all prescriptions, 
$\mathbb{S}(D_j, P_i)$ depicts the collection of prescriptions containing both $D_j$ and $P_i$, and $\mathbb{S}(D_j)$ illustrates the set of prescriptions containing $D_j$. 
{Intuitively, $\texttt{tf}(D_j, P_i)$ measures how often pill $P_i$ is prescribed for diagnosis $D_j$, thus it reflects the significance of $P_i$ regarding treating $D_j$. However, in practice, some pills are more popular among prescriptions (e.g., Sustenance, Dorogyne, Betaserc, etc.), which may cause negative bias when applying only the $\texttt{tf}$ term. That effect can be mitigated by the term $\texttt{idf}(P_i)$.}

{Once finished formulating the impact factors of the pills and diagnoses, we transform each term $\mathcal{I}(P_j, D_i)$ into a probabilistic view by a simple normalization over all diagnoses as follows}
\begin{equation*}
    \small
    p(P_j, D_i) = \frac{\mathcal{I}(P_j, D_i)}{\sum_{D \in \mathbb{D}}\mathcal{I}(P_j, D)}, 
\end{equation*}
{where $\mathbb{D}$ denotes the set of all diagnoses. Given $p(P_j, D_i)$, we define the weight ${W_c}(P_i, P_j)$  of the edge $E_{ij}$ connecting vertices $P_i$ and $P_j$ as the probability $p(P_i, P_j)$ that $P_i$ and $P_j$ are prescribed for the same diseases. ${W_c}(P_i, P_j)$ can be formulated as follows. }
\begin{equation}
    \small
     {W_c}(P_i, P_j) := p(P_i, P_j) \approx \sum_{D \in \mathbb{D}} p(P_i, D) \times p(P_j, D).
\label{eq:weight_mcg}
\end{equation}


{\subsubsection*{Relative Size Graph Modeling}
\label{subsec:method_size_modeling}
The Size-graph is represented by a directed graph $\mathcal{G}_s = \left \langle V, E, W_s \right \rangle$.
The edge weight $W_s$ is modeled so that the weight of an edge $\overrightarrow{E_{ij}}$ connecting from $P_i$ to $P_j$ is proportional to the size ratio of $P_i$ to $P_j$.
The primary source for constructing the Size-graph is the annotations of the training dataset's bounding boxes. As the camera locations for multiple pictures are different, the exact size of each bounding box cannot be utilized directly.
Therefore, we instead define a so-called \emph{size indicator}, a normalized representation of pill size, which is determined as follows.
\begin{itemize}
    \item \textbf{Step 1:} We begin with an arbitrary pill class by initializing its size indicator to $1$, while those of other pill classes are initialized to $0$.
    \item \textbf{Step 2:} From the current node $P_i$, we traverse through all its 1-hop neighbors $P_j$, and calculate $P_j$'s size indicator $s_j$ as $s_j := s_i \times \frac{|B_j|}{|B_i|}$, where $B_i, B_j$ are the two bounding boxes of $P_i$ and $P_j$ in a particular image in the training set. Step 2 is repeated until all the vertices of $\mathcal{G}_s$ are traversed.
\end{itemize}
Given the size indicators of all vertices, we now define the weight of edge $\overrightarrow{E_{ij}}$ as the ratio of $s_i$ to $s_j$.
\subsection*{Visual Feature Extractor}
\label{sec:method_visual_extractor}
This block is responsible for localizing and extracting the features of Regions of Interest (RoIs).
For this purpose, we adopt components from Faster RCNN~\cite{fasterrcnn}, a conventional two-step object detector architecture. Nevertheless, our proposed framework is compatible with any alternative object detection architecture.
The Visual Feature Extractor consists of three components: a Convolutional Network, a Region Proposal Network, and an RoI Pooling Layer, as depicted in Fig.~\ref{fig:generalflow}. }
{RPN is a fully convolutional network that takes the visual feature vector from the previous module and generates proposals with various scales and aspect ratios. The RoI Pooling layer works simply by splitting each region proposal into a grid of cells and then applying the max pooling operation to each cell in the grid. The combination of the grids' values forms the visual feature vectors of the RoIs.}

{\subsection*{Inter-Pill Relational Feature Extractor}
\label{sec:method_a_module}
To enhance the efficacy of this a priori, we observed that rather than the whole graphs representing the interaction between all pills, we should utilize sub-graphs concentrating on the ones most likely to appear in the image.
Motivated by this observation, we employ the Inter-Pill Relational Feature Extractor, responsible for extracting condensed sub-graphs from generic Co-graph and Size-graph. 
Moreover, previous studies have pointed out that the appearances of pills convey implications about their efficacy or ingredients~\cite{medical3}. In light of this, utilizing pills' visual feature vectors, we develop a visual-based graph that models the implicit relationship between medications indicated by their visual appearance.}

{\noindent \textbf{Condensed Co-graph and Size-graph.} 
Our main idea is to employ a so-called \emph{Pseudo Classifier}, which provides approximate classification results using solely visual features of RoIs. These temporary identification results are then utilized as a filter layer to eliminate redundant information from the original Co-graph and Size-graph, leaving only information about pill classes probable to appear in the input image.
In current implementation, the pseudo classifier is straightforwardly implemented as a fully connected layer. 
Let $N$ be the number of pill classes and $M$ be the number of pill bounding boxes (i.e., RoIs) in the input image. Suppose $P={[p_{ij}]}_{M\times N}$ is the matrix whose row vectors represent the logits produced by the pseudo classifier, and $\mathcal{A}_c={[a^c_{kl}]}_{N\times N}$, $\mathcal{A}_s={[a^s_{kl}]}_{N \times N}$ denote the weighted adjacency matrices of the Co-graph $\mathcal{G}_c$ and Size-graph $\mathcal{G}_s$, respectively. 
The condensed adjacency matrices, denoted as $\Tilde{\mathcal{A}}_c$ and $\Tilde{\mathcal{A}}_s$ are matrices of size $M \times M$, each row depicts the condensed relational information of a pill, i.e., a specific RoI, with others in the input image. $\Tilde{\mathcal{A}_c}$ and $\Tilde{\mathcal{A}_s}$ are obtained by performing a composition of matrix multiplications as follows.
\begin{eqnarray}
    \small
    \label{eq:ac} \Tilde{\mathcal{A}}_c &= \sigma (P) \cdot \mathcal{A}_c \cdot \sigma (P)^T\\
    \label{eq:as} \Tilde{\mathcal{A}}_s &= \sigma (P) \cdot \mathcal{A}_s \cdot \sigma (P)^T, 
\end{eqnarray}
where $\sigma$ denotes the \texttt{Softmax} activation function. 
Intuitively, the item in the $i$-th row and $j$-th column of $\Tilde{\mathcal{A}}_c$ and $\Tilde{\mathcal{A}}_s$ highlights the relationship of the $i$- and $j$-th RoIs. }

{\noindent \textbf{Visual Semantic Graph.}
As mentioned above, the visual semantic graph $\Tilde{\mathcal{G}}_v = \left \langle \Tilde{V}, \Tilde{E}_v, \Tilde{W}_v \right \rangle$ is in charge of capturing the visually semantic correlation among pills in the input image. 
The detailed algorithm to construct this graph is as follows.
All visual feature vectors are first passed through a non-linear function $\mathcal{F}$: $\mathbb{R}^{h} \rightarrow \mathbb{R}^{h'}$ to transform from the original $h$-dimensional space into a $h'$-dimensional latent one, where their relationship can be best presented. The latent output vectors are then directly used for calculating the correlations between RoIs.
Let $R_i, R_j$ are two RoIs in the input image, and $z_i, z_j$ are their feature vectors created by the Visual Feature Extractor block, the weight of the edge connecting $R_i$ and $R_j$ is defined as $ \Tilde{W_v}(R_i, R_j) = z_i \cdot z_j$.}




{\subsection*{Multi-modal Data Fusion}
\label{subsec:method_g_module}

\begin{figure}[!tb]
        \centering
        \includegraphics[width=0.3\columnwidth]{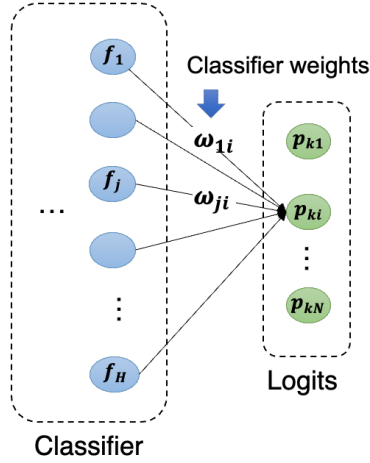}
        \caption{\textbf{Node attribute modeling.} ${\omega}_i={\left [\omega_{1i}, \dots, \omega_{Hi} \right ]}^T$ is the classifier weights corresponding to the $i$-th pill class, capturing the representative features of this class. $\sum_{i=1}^{H}p_{ku} \times {\omega}_i$ is the attribute of the $i$-th RoI. \label{fig:att}}
    \end{figure}

\begin{figure}[!tb]
        \centering
        \includegraphics[width=0.75\columnwidth]{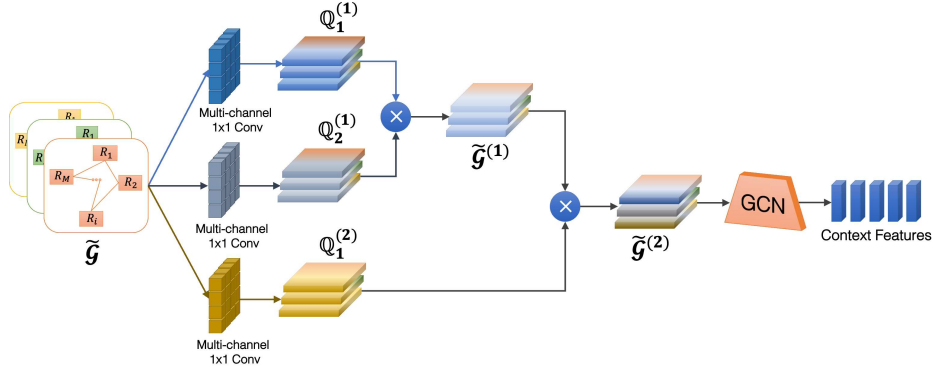}
        \caption{\textbf{Graph Transformer Network (GTN) architecture~\cite{gtn}.} GTN softly selects adjacency matrices (edge types) from the set of adjacency matrices $\mathbb{A}$ of a heterogeneous graph $\Tilde{\mathcal{G}}$ and learns new meta-path graphs represented by $\Tilde{\mathbb{A}}$ via the matrix multiplication of two selected adjacency tensors $\mathbb{Q}_1$ and $\mathbb{Q}_2$. The soft adjacency matrix selection is the weighted sum of candidate adjacency matrices obtained by $C$ channels of $1 \times 1$ convolution with non-negative weights with \texttt{softmax} activation. \label{fig:gtn_architec}}
    \end{figure}

After going through the second and third blocks, we get the visual features of the RoIs and three relational graphs representing the relationships between the RoIs. 
This information is now fed into the Multi-modal Data Fusion to generate the final feature vectors, each of which encapsulates both the intra-Pill visual characteristic of an RoI and the inter-Pill interaction of that RoI with the others. 
The Multi-modal Data Fusion comprises two steps: graph embedding and data concatenation. 
The former obtains the heterogeneous relational graph $\mathcal{G}$ and transforms it into context features in the vector space, while the latter concatenates the context feature vectors with visual features to generate the final enhanced features.
We utilize the Graph Transformer Network (GTN)\cite{gtn} for the graph embedding.
The reason for choosing the GTN is due to its ability to handle heterogeneous input and adaptive graph structures.
Before going into the detail of the GTN, it is crucial to define the node attribute of graph $\Tilde{\mathcal{G}}$.
As each node of $\Tilde{\mathcal{G}}$ represents an RoI, the node attribute should be the most representative characteristic of the ROIs. 
Using the retrieved RoI visual features to depict the relevant ROIs is the most natural solution but is not advantageous due to several factors, including the unreliability in dealing with ambiguous samples or the $intra-variance$ in visual features of one class \cite{reasoning_rcnn}. 
To this end, classifier weights has been introduced as a simple yet effective alternative.
According to ~\cite{weight_3}, the classifier weights connected to the $i$-th neuron in the last layer (which is denoted as ${\omega}_i={\left [\omega_{1i}, ..., \omega_{Hi} \right ]}^T$ in Fig.\ref{fig:att}) corresponds to the $i$-th pill class, encapsulating the representative characteristics of this class.
Let $p_k=\left [p_{k1}, ..., p_{kN}\right ] $ be the logit vector of the $k$-th RoI, where $p_{ki}$ depicts the likelihood for the $k$-th RoI to be classified into the $i$-th class, we define $\sum_{i=1}^{H}p_{ku}\times {\omega}_i$ the attribute of the $k$-th RoI.
Intuitively, this attribute can be considered as a decomposition of the RoI's characteristic in the space of the classes' features.
}
{Figure~\ref{fig:gtn_architec} depicts the GTN's architecture, which consists of two phases. The former can be seen as a meta-path generator that fuses information from multiple input adjacency matrices to generate a composite graph structure. 
This newly generated graph serves as the second stage's input, which comprises a Graph Convolutional Network (GCN) and is responsible for producing a representation for each node.
Specifically, the GTN consists of $l$ Graph Transformer (GT) layer; the $l$-th layer applies the $C$-channel 1D convolution operation on the input graph $\Tilde{\mathcal{G}}$ to obtain a stack of new graph structure $\mathbb{Q}_1^{(l)} \in \mathbb{R}^{M \times M \times C}$ as follows.
\begin{equation}
\small
\mathbb{Q}_1^{(l)} = F(\Tilde{\mathcal{G}}; W_\phi^{(l)}) = \phi(\Tilde{\mathcal{G}};\sigma(W_\phi^{(l)})),
\end{equation}
where $\phi$ indicates the convolution layer, $W_\phi^{(l)} \in \mathbb{R}^{C \times 1 \times 1 \times K}$ represents the parameter of $\phi$, and $K$ implies the number of relations contained in the original graph $\Tilde{\mathcal{G}}$.
The stacked graph $\mathbb{Q}_1^{(l)}$ serve as the first component in creating length $l$ meta-paths, while $\mathbb{Q}_2^{(l)}$ is taken as $\Tilde{\mathcal{G}}^{(l-1)}$, i.e., $\Tilde{\mathcal{G}}^{(l)} = \mathbb{Q}_2^{(l)} \odot \mathbb{Q}_2^{(l)}$. 
%
To balance computational overheads and model performances, with PGPNet, we fix $l = 2$. }

{The resulting graph $\Tilde{\mathcal{G}}^{(2)}$, together with RoIs' representative features $X_{RoI}$, are then utilized as the input for the Graph Convolution Network (GCN) to generate the final node presentations. 
These vectors are directly concatenated with their corresponding RoIs' visual features before getting fed into the Bounding Box Regressor and Classifier to produce the final detection results.} 

\subsection*{PGPNet's Losses}
\label{sec:method_loss}
{This section presents the details about our model's objectives and the corresponding losses to achieve those goals. }

\subsubsection*{Two-step Object Detectors' Losses}
\label{subsec:method_normal_loss}

\hfill\\
{\textbf{The Region Proposal Network's Losses.}
The loss for Region Proposal Network consists of two components: classification loss combined and bounding box regression loss.
Let $p_i, p_i^*$ be the predicted probability of an anchor $i$ being an object and the ground truth label whether anchor $i$ is the object, respectively; $t_i$ and $t_i^*$ depict the differences of four predicted coordinates, and the ground truth coordinates with the coordinates of the anchor boxes, respectively. 
The classification loss $\mathcal{L}_{cls}$ and bounding box regression loss $\mathcal{L}_{\mathrm{box}}$ are defined as follows.
\begin{equation}
\begin{split}
    \small
    \mathcal{L}_{RPN}\left(\left\{p_{i}\right\},\left\{t_{i}\right\}\right) &= \frac{1}{N_{\mathrm{cls}}} \sum_{i} \mathcal{L}_{\mathrm{cls}}\left(p_{i}, p_{i}^{*}\right) \\
    &+ \frac{\lambda}{N_{\mathrm{box}}} \sum_{i} p_{i}^{*} \cdot L_{1}^{\mathrm{smooth}}\left(t_{i}-t_{i}^{*}\right),
\end{split}
\end{equation}
where
\begin{equation}
\small
    L_{1}^{\text {smooth }}(x)= \begin{cases}0.5 x^{2} & \text { if }|x|<1 \\ |x|-0.5 & \text { otherwise }\end{cases}.
\end{equation}
Here the $\mathcal{L}_{cls}$ is a binary classification \emph{log loss}, $N_{cls}$ and $N_{box}$ are two normalization terms, where $N_{cls}$ is set to the mini-batch size, while $N_{box}$ is the number of anchor boxes. 
$\lambda$ is a hyper-parameter, which is responsible for balancing between $\mathcal{L}_{\mathrm{cls}}$ and $\mathcal{L}_{\mathrm{box}}$.}

{\noindent \textbf{Output's Losses.}
The PGPNet's final results consist of the coordinates of the RoIs' bounding boxes and predicted labels for the RoIs. 
We employ two distinct losses to accomplish this objective.
While the loss for a bounding box regressor is equal to that of the RPN network, the classification loss $\mathcal{L}_{cls}^{out}$ is instead the cross entropy loss for the multilabel classification task, which is represented as follows $\mathcal{L}_{cls}^{out} = - \sum^N_{i=0}p^*_i \log(p_i)$.}

{\subsubsection*{Triplet Co-occurrence Enhancement Loss.}
\label{subsec:method_auxi_loss}
In this section, we propose an auxiliary loss named Triplet Co-occurrence Enhancement Loss which leverages the co-occurrence graph to boost the accuracy of the Pseudo Classifier.
The idea behind the auxiliary loss is that it encourages the co-occurrence likelihood of pills that are close together on the co-occurrence graph.
To this end, we construct our auxiliary loss as a contrastive loss that maximizes the co-occurrence probability of positive pairings (i.e., pills joined by edges with the most significant weights in the co-occurrence graphs) while minimizing the co-occurrence probability of negative pairs (i.e., pills that are not connected or connected by edges with smallest weights). In action, for each training mini-batch, PGPNet would treat all the ground truth pills in given images as the set of anchors and build up their corresponding positive as well as negative sets. After that, Triplet Co-occurrence Enhancement Loss would do its job for enhancing the robustness of Pseudo Classifier.
The detail of the auxiliary loss is as follows.}

{Let's denote the $i$-th Region of Interest as $R_i$ with its corresponding label of $l_i$. 
Moreover, let $N^i_{pos}$ and $N^i_{neg}$ be the positive and negative samples of $R_i$, where $N^i_{pos}$ comprises of $k+1$ nearest neighbors and $N^i_{neg}$ consists of $k+1$ furthest neighbors of $R_i$. 
We suppose that the groundtruth labels of $N^i_{pos}$ and $N^i_{neg}$ are  $L_{pos} = \{ l^0_{pos}, l^1_{pos}, \dots, l^{k}_{pos} \}$, and $L_{neg} = \{ l^0_{neg}, l^1_{neg}, \dots, l^{k}_{neg} \}$, respectively. 
The auxiliary loss concerning the $i$-th RoI is defined by 
\begin{equation} \label{eq:aux_i}
\begin{split}
\small
    \mathcal{L}_{aux}^i &= p_i(l_i)p(N_{pos}^i) - (1 -p_i(l_i))p(N_{neg}^i) \\
                        &= p_i(l_i)\sum_{j=0}^{k}[1 - \prod_{m=0}^{M}(1 - p_m(l^j_{pos}))] \\
                        &- (1 - p_i(l_i))\sum_{q=0}^{k}[1 - \prod_{n=0}^{M}(1 - p_n(l^q_{neg}))],
\end{split}
\end{equation}
and those for RoI is $\mathcal{L}_{aux} = \sum_{i=0}^M\mathcal{L}_{aux}^i$.
In Formulas (\ref{eq:aux_i}) $M$ is the total number of RoIs in the image, $p$ is the output after going through \texttt{softmax} activation of logits produced by Pseudo Classifier. 
The objective during the training process is to {maximize} $\mathcal{L}_{aux}$, which in turn maximizes each {positive} term $p_i(l_i)p(N_{pos}^i)$ while minimizing the {negative} opposition $(1 -p_i(l_i))p(N_{neg}^i)$.
}

\section*{Dataset and Experiment Settings}
\label{sec:result}

\begin{figure*}[!t]
\centering
\begin{minipage}{0.6\linewidth}
    \centering \small
    	\setlength\tabcolsep{4pt} 
        \resizebox{\columnwidth}{!}{%
            \begin{tabular}{l|l|l|l}
                \toprule
                & \textbf{NIH} & \textbf{CURE}  & \textbf{VAIPE}            \\ \midrule
                Number of pill images     & 7,000    & 8,973  & 9,426             \\
                Number of pill categories & 1,000    & 196   & 96               \\
                Number of capture devices & 1    & 1   & $>$ 20               \\
                Instance per category     & 7       & 40-50 & $>$ 30 \\
                Illumination conditions   & 1       & 3     & $>$ 50 \\
                Backgrounds   & 1       & 6     & $>$ 50 \\
                Number of prescriptions   & 0       & 0     & 1,527            \\ \bottomrule
            \end{tabular}
        }
        \captionof{table}{\label{tab:dataset_meta} {An overview of existing public datasets for the task of image-based pill detection. To the best of our knowledge, the introduced VAIPE dataset is currently the largest dataset for pill identification, which was collected in real-world settings and came up with prescriptions.}}
\end{minipage}
\hspace{0.1cm}
\begin{minipage}{0.25\linewidth}
    \centering \small
    	\setlength\tabcolsep{4pt} 
        \resizebox{\columnwidth}{!}{%
            \begin{tabular}{ll}
            \toprule
            \multicolumn{2}{c}{\textbf{Training dataset}} \\
            \cmidrule(lr){1-2}
            \multicolumn{1}{c}{Prescriptions} & \multicolumn{1}{c}{Images} \\ \midrule
            1,527 (100\%)             & 7,514 (78\%)    \\ \midrule
             \\
             \\
            \multicolumn{2}{c}{\textbf{Testing dataset}}  \\ 
            \cmidrule(lr){1-2} 
            \multicolumn{1}{c}{Prescriptions} & \multicolumn{1}{c}{Images} \\ \midrule
            0 (0\%) & 1,912 (22\%)     \\ 
            \bottomrule
            \\
            \end{tabular}
        }
        \captionof{table}{Details of training and testing datasets. \label{tab:train_test_split}}
\end{minipage}
\vspace{-10pt}
\end{figure*}
{We conduct extensive experiments to validate the effectiveness of the proposed approach. 
In the following, we first introduce our in-house pill identification dataset, called VAIPE, which will be used to evaluate the proposed approach, and then explain our evaluation metrics and experimental settings. 
To assess the effectiveness of the proposed method, we conducted comparative assessments against a number of established models, including the detection backbones we selected, such as Faster R-CNN~\cite{fasterrcnn} and YOLOv5 \cite{yolov5}, as well as other related frameworks such as SGRN~\cite{sgrn} and the Mask RCNN-based approach described in ~\cite{kwon_pill}. We also perform ablation studies to investigate the efficiency of key components in our framework.
\subsection*{Dataset and Pre-processing}
\textbf{Motivation.}  {To the best of our knowledge, previous studies on the pill identification problem~\cite{tan2021comparison},~\cite{cure1},~\cite{Ling_2020_CVPR},~\cite{cure} only focus on datasets collected in constrained environments. For instance, existing datasets such as {NIH Dataset} \cite{nih_dataset} are constructed under ideal conditions in lighting, backgrounds, and equipment or devices. The CURE dataset~\cite{Ling_2020_CVPR} provides only one pill per image. Hence, these datasets do not reflect the real-world scenarios in which patients take an arbitrary number of drugs, and their environmental conditions (e.g., backgrounds, lighting conditions, mobile devices, etc.) are greatly varied. Additionally, many pills have nearly identical visual appearances. The fact that they appear alone in the images of these datasets will inevitably confuse the detection frameworks. Consequently, none of the existing datasets can be directly applied to the real-world pill detection problem or can only be applied with low reliability. There is no publicly available dataset of these pills images in which the pills follow intakes of actual patients. This limits the development of machine learning algorithms for the detection of pills from images as well as for building real-world medicine inspection applications. To address this challenge, we build and introduce a new, large-scale open dataset of pill images, which we called VAIPE. }

\textbf{Data Descriptor.} The VAIPE is a large-scale and open pill image dataset for visual-based medicine inspection. 
The dataset contains approximately 10,000 pill images that were manually collected in unconstrained environments. In this study, no hypotheses or new interventional procedures were generated. Also, no investigational products or clinical trials were used for patients. In addition, there were no changes in treatment plans for any patients involved. Pill images were retrospectives collected, and all identifiable information of patients was de-identified. Therefore, there was no requirement for ethics approval~\cite{bworld}. 

Pill images are collected in many different contexts (e.g., various backgrounds, lighting conditions, in-hand or out-of-hand, etc.) using smartphones. These images are then manually labeled using the information from the relevant prescriptions. In summary, the number of pills per image is about $5 - 10$, and the total number of pill images collected was $9,426$ pill images with $96$ independent pill labels. To train the proposed deep learning system, the pill images from the VAIPE dataset are resized so that the shortest edges have a size of $800$, with a limit of $1,333$ on the longer edge. The ratios are kept the same as the original images if the max size is reached, then downscale so that the longer edge does not exceed $1,333$.

\textbf{Data Validation.} Patient privacy was controlled and protected. In particular, all images were manually reviewed to ensure that all individually identifiable health information of the patients has been removed to meet the General Data Protection Regulation (GDPR) \cite{gdpr}. Annotations of pill images were also carefully examined. Specifically, all images were manually reviewed case-by-case by a team of 20 human readers to improve the quality of the annotations.

\textbf{Comparison with Existing Datasets.} Table \ref{tab:dataset_meta} provides a summary of the aforementioned datasets (including NIH, CURE, and VAIPE) together with other ones of moderate sizes, meta-data, and other properties. Compared to the two previous datasets, the VAIPE dataset is constructed under a much more flexible procedure that reflects the characteristic real-world data distributions. Hence, the introduced dataset can serve as a reliable data source for training \emph{generic pill detectors}.

\subsection*{Evaluation Metrics}
{We evaluate the proposed method and other related works by the COCO APs metrics~\cite{coco_ap}. This set of metrics is widely accepted and used for evaluating state-of-the-art object detectors. Mean Average Precision (mAP), as its name suggests, is the mean of Average Precision (AP) overall $C$ classes and all the targeted IoU thresholds in the threshold set $T$ calculated by $ mAP = \frac{1}{C |T|} \sum_{i}^{C} \sum_{t \in T} AP_{i,t}$, 
where Average Precision ($AP_{i,t}$) is the area under the Precision-Recall curve, calculated for the class $i$ at a given IoU threshold $t$.} 
\subsection*{Comparison with state-of-the-art methods}
\textbf{Comparison Benchmarks.} {To show the effectiveness of the proposed method, we conducted a comparison with the state-of-the-art object detectors, including our detection backbones: Faster R-CNN~\cite{fasterrcnn}, YOLOv5 \cite{yolov5}, and related works: SGRN~\cite{sgrn}, Mask RCNN-based approach \cite{kwon_pill}. Throughout the literature, the baseline with which PGPNet presently integrates is Faster R-CNN \cite{fasterrcnn}; hence, the original framework is utilized for our comparison. 
We adopt two different CNNs and one Transformer-based module for visual feature extractor, namely ResNet-50-C4, ResNet-50-FPN and Swin Transformer V2 - SwinV2 \cite{swinv2}(Fig.~\ref{fig:generalflow}). Specifically, for two ConvNets, we use a single feature map produced by convolution block C4 of the ResNet-50 model in ResNet-50-C4. In ResNet-50-FPN, we replace C4's feature map with multi-scale feature maps produced by Feature Pyramid Network (FPN)~\cite{fpn}. As for the Swin Transformer module, we are currently utilizing the SwinV2-T configuration \cite{swinv2} to ensure that the number of model parameters is comparable to that of ResNet-50. 
In addition, we also make adaption for PGPNet with YOLOv5 \cite{yolov5} detection backbone. Two configurations of YOLOv5s and YOLOv5n are currently adopted. 
Also, the most relevant frameworks compared with our PGPNet are also put into comparison: a representative approach that utilizes an external knowledge graph for Object Detection task~\cite{sgrn}; a Mask RCNN-based baseline that also proposed to solve the same task of multi-pill detection \cite{kwon_pill}. \\
\noindent For a fair comparison, a fixed set of hyper-parameters is used for PGPNet throughout all experiments.}

\subsection*{Implementation Details}
{We conduct all the experiments using the Pytorch (version 1.10.1) on an Intel Xeon Silver 4210 2.20GHz system with $2$ $\times$ NVIDIA GeForce RTX 3090 GPUs.  We train and test all targeted models on the training and testing sub-datasets provided in Table~\ref{tab:train_test_split}. Specifically, we initialize all the networks with the weights achieved by pre-training them on COCO 2017 dataset~\cite{coco}. We then train the models in $20,000$ iterations with a batch size of $16$. AdamW~\cite{adamw} optimizer is used with the initial learning rate of $0.001$. We also augment the training data by using simple techniques such as random horizontal and vertical flips to prevent overfitting. For our PGPNet implementation, we set the dimensions of node embeddings at $64$. We also design the Graph Transformer Module with only one layer and $10$ channel set.}

\section*{Experimental Results}
This section reports our experimental results. We evaluate the effectiveness of PGPNet in three aspects: robustness, reliability, and explainability. The details are described below. 


\subsection*{Robustness and Reliability of PGPNet}
\label{subsec:eval_backbones}
\subsubsection*{Comparison with Faster R-CNN and YOLOv5}
\begin{table}[!tb]
\centering
\small
\setlength\tabcolsep{3pt} 
\resizebox{0.8\columnwidth}{!}{%
\begin{tabular}{lll|cccccc}
\toprule
\multicolumn{3}{c|}{\textbf{Method}}                                      & \multicolumn{1}{c}{\textbf{mAP}} & \multicolumn{1}{c}{\textbf{AP50}} & \multicolumn{1}{c}{\textbf{AP75}} & \multicolumn{1}{c}{\textbf{APs}} & \multicolumn{1}{c}{\textbf{APm}} & \multicolumn{1}{c}{\textbf{APl}} \\ \midrule
\parbox[t]{2mm}{\multirow{7}{*}{\rotatebox[origin=c]{90}{\textcolor{red}{\textbf{Two-step}}}}} & 
\multicolumn{1}{l|}{\multirow{2}{*}{
    \begin{tabular}[l]{@{}l@{}} Faster R-CNN \\ (ResNet-50-C4)\end{tabular}}}  
    & Vanilla & 62.6               & 87.0                 & 74.4                  & 75.0                     & 58.3                 & 62.9                 \\
\multicolumn{2}{l|}{}                               & PGPNet       & \textbf{68.3 (+9.2\%)}       & \textbf{92.5}        & \textbf{81.7}          & \textbf{80.0}             & \textbf{64.3}         & \textbf{68.7}         \\
\cmidrule(lr){2-9}
& \multicolumn{1}{l|}{\multirow{3}{*}{\begin{tabular}[l]{@{}l@{}} Faster R-CNN \\ (ResNet-50-FPN)\end{tabular}}} & Vanilla & 63.7                & 86.6                & 76.9                   & 71.2                 & 58.1                & 64.6               \\
\multicolumn{2}{l|}{}                               & PGPNet       & \textbf{69.7 (+9.4\%)}       & \textbf{94.4}         & \textbf{83.4}         & \textbf{90.0}    & \textbf{66.4}        & \textbf{70.1}   \\
\multicolumn{2}{l|}{} &   &                &                     &                  &                    &                  &                 \\ 
\cmidrule(lr){2-9}
& \multicolumn{1}{l|}{\multirow{2}{*}{\begin{tabular}[l]{@{}l@{}} \duy{Faster R-CNN} \\ \duy{(SwinV2-T)}\end{tabular}}} & \duy{Vanilla} & \duy{59.7}                & \duy{84.5}                & \duy{72.3}                   & \duy{66.9}                 & \duy{54.0}                & \duy{60.1}               \\
\multicolumn{2}{l|}{}                               & \duy{PGPNet}       & \duy{\textbf{62.6 (+4.8\%)}}       & \duy{\textbf{87.2}}         & \duy{\textbf{75.5}}         & \duy{\textbf{68.6}}    & \duy{\textbf{56.6}}        & \duy{\textbf{62.9}}   \\
\midrule
\parbox[t]{2mm}{\multirow{4}{*}{\rotatebox[origin=c]{90}{{\textbf{One-step}}}}} & \multicolumn{1}{l|}{\multirow{2}{*}{YOLOv5n}}  & Vanilla & 37.9 & 50.8 &45.4    & 87.5    & 49.1    & 38.3  \\
\multicolumn{2}{l|}{}                              & PGPNet  & \textbf{43.0 (+12.0\%)}   & \textbf{58.4}    & \textbf{51.3}    & \textbf{82.5}    & \textbf{52.4}    & \textbf{43.7}    \\ \cmidrule(lr){2-9}
& \multicolumn{1}{l|}{\multirow{2}{*}{YOLOv5s}} & Vanilla & 57.5 & 75.8 & 68.3 & 85.0 & 58.3 & 57.0 \\
\multicolumn{2}{l|}{}                              & PGPNet  & \textbf{63.4 (+10.2\%)} & \textbf{85.9} & \textbf{76.4} & \textbf{89.9} & \textbf{58.3} & \textbf{64.1} \\
\bottomrule
\end{tabular}
}
\caption{Comparison of detection performance of PGPNET with state-of-the-art object detectors (Vanilla)  on VAIPE dataset. 
Best results are highlighted in \textbf{bold} text. \label{tab:eval_backbone}}
\vspace{-15pt}
\end{table}

\begin{figure*}[!t]
\begin{minipage}{0.32\linewidth}
        \centering
        \includegraphics[width=\columnwidth]{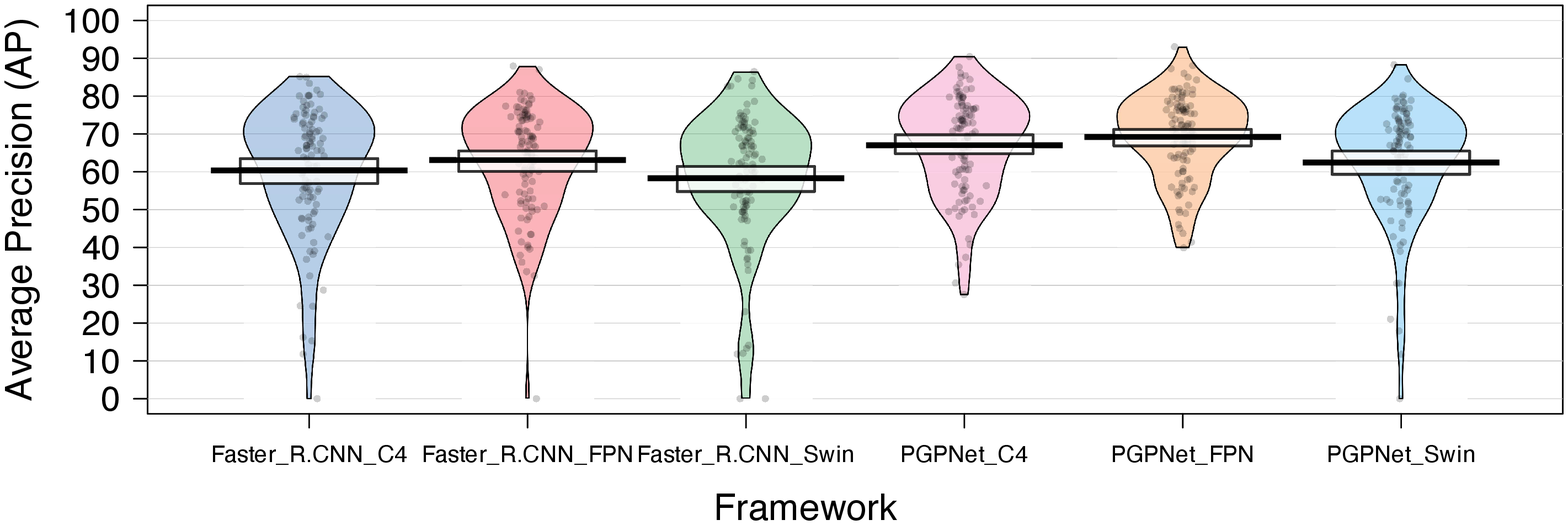}
        \caption{Comparison of the PGPNet performance with the Faster R-CNN baseline over each individual class. \label{fig:eval_backbone_class}}
\end{minipage}
\hspace{0.1cm}
\begin{minipage}{0.66\linewidth}
    \centering
    \subfloat[Faster R-CNN \label{fig:reliable_frcnn}]{%
      \includegraphics[clip,width=0.32\columnwidth]{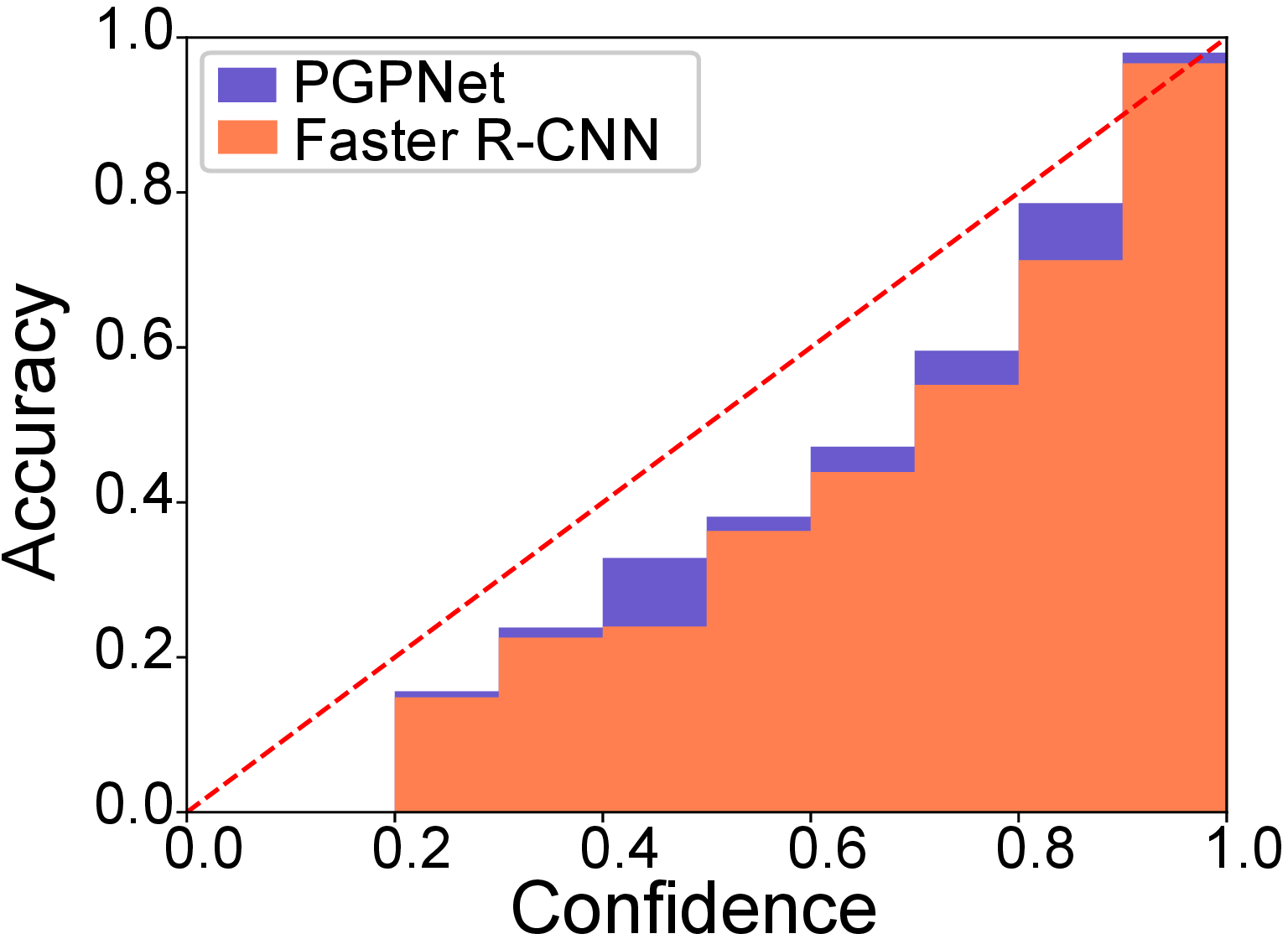}%
    }
    \hfill
    \subfloat[YOLOv5 \label{fig:reliable_yolo}]{%
      \includegraphics[clip,width=0.32\columnwidth]{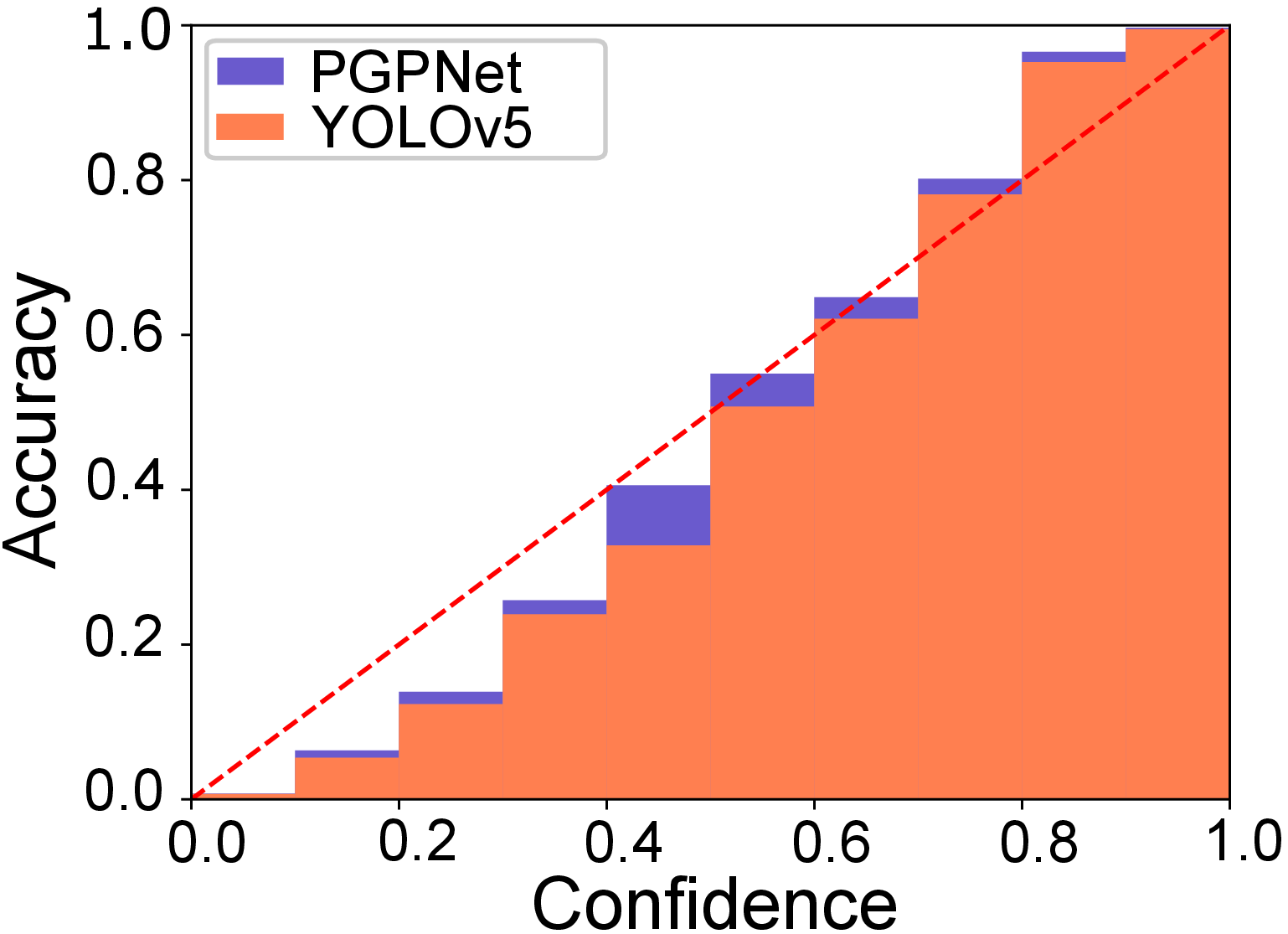}%
    }
    \hfill
    \subfloat[SGRN \label{fig:reliable_sgrn}]{%
      \includegraphics[clip,width=0.32\columnwidth]{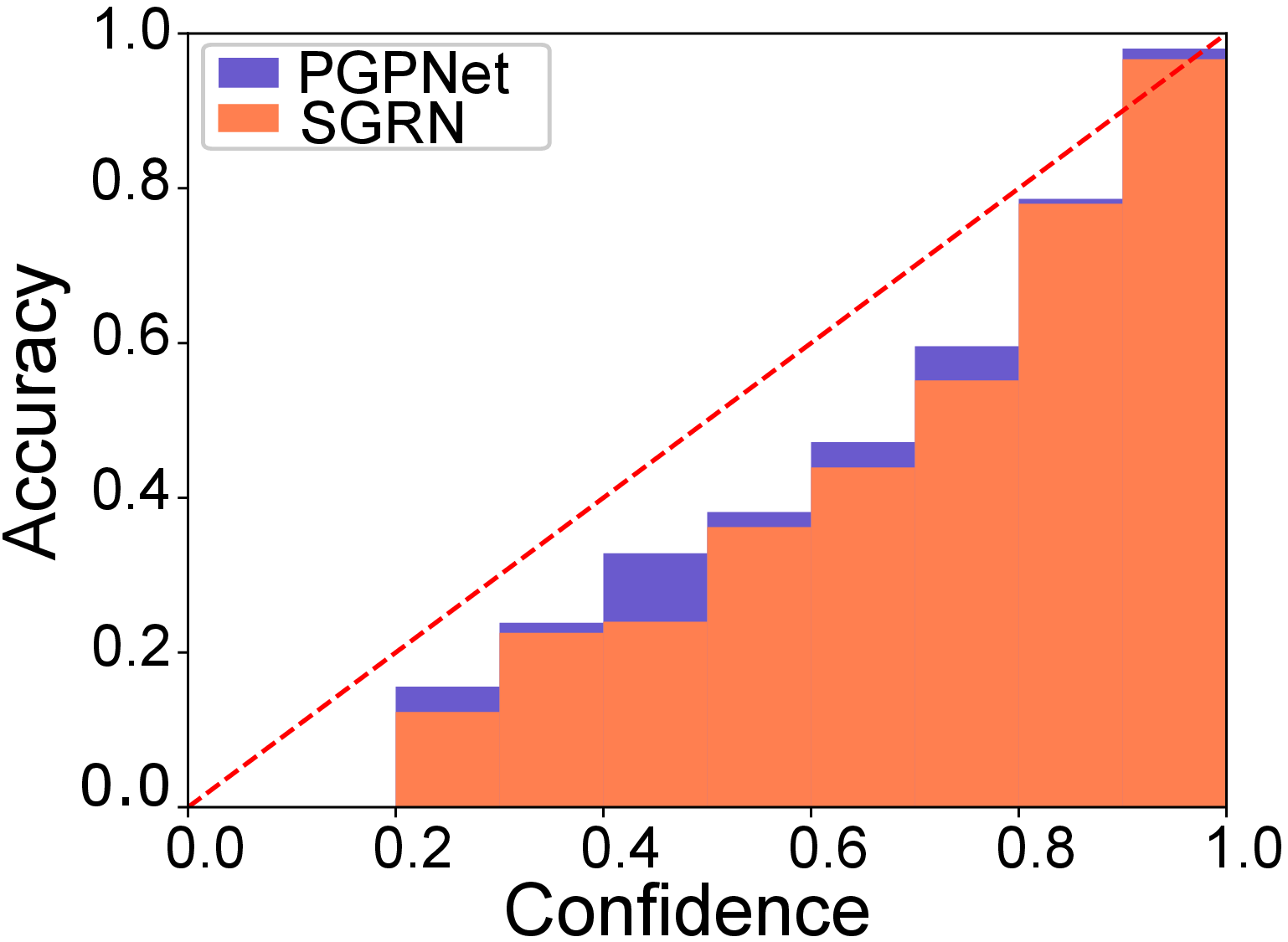}%
    }
    \caption{Reliability investigation for PGPNet and different baseline performances.\label{fig:reliable_all}}
\end{minipage}
\vspace{-0.5cm}
\end{figure*}
\textbf{Detection Performance.}
{Table~\ref{tab:eval_backbone} shows the experimental results of PGPNet and the state-of-the-art object detectors framework (Vanilla), e.g., Faster R-CNN (two-step detector), and YOLOv5~\cite{yolov5} (one-step detector) 
on the VAIPE dataset. As shown, PGPNet obtained better results than Faster R-CNN by large performance gaps for all evaluation metrics. Specifically, when using the ResNet-50-C4 model as the visual feature extractor model, the average precision mAP of Faster R-CNN was $62.6$, while that of PGPNet was $68.3$. The proposed method improves the performance over the baseline Faster R-CNN by $9.2\%$. 
Under strict metrics, e.g., AP75, PGPNet also outperforms Faster R-CNN $8-9\%$.
In addition, we observed similar behavior when using the ResNet-50-FPN model. The proposed PGPNet makes an improvement of $9.4\%$ for the mAP metrics. With a Transformer-based backbone, here a Swin Transformer V2 configuration - SwinV2-T ~\cite{swinv2}, the results are slightly worse compared to those produced by ResNet-based counterparts, for both the vanilla or PGPNet alternatives. However, PGPNet still show its superior when being install with this backbone, as the empirical result for AP metrics is improved by $4.8\%$ compared to the vanilla SwinV2-T Faster R-CNN model.}

{For YOLOv5, PGPNet outperformed Vanilla by a significant margin across all performance metrics in both YOLO instances. Specifically, the average precision AP of the vanilla model with YOLOv5n was $37.9$ while that of PGPNet was $43.0$ ($12\%$ improvement). In the case of a larger alternative, YOLOv5s, a similar conclusion can be drawn, namely that PGPNet improves overall mAP metrics by $5.9$, e.g., $10.2\%$.}


{Figure \ref{fig:eval_backbone_class} visualizes the AP 
for all classes in the dataset when using Faster R-CNN as the backbone. The first three bins denote Faster R-CNN alternatives, and the later three are the corresponding PGPNet configurations. The dots in the figure represent AP values for classes; the vertical line is the indicator for the mean value, while the rectangle bar is the $95\%$ High-Density Interval (HDI) band. Apart from the fact that the mean AP over all classes of PGPNet variances is better than those produced by Faster R-CNN, we found that PGPNet also has more reliable and stable results over all classes.  
Specifically, PGPNet helps to improve the AP of classes that Faster R-CNN frequently confuses (the points with low APs in the blue and pink beans). As a result, the  three beans of Faster R-CNN exhibit a large variance, i.e.,  the AP ranged from $0$ to around $90$. In contrast, the beans of PGPNet performance are more condensed and have shorter tail, i.e., the AP ranged from  $40$ (or $50$) to around $90$.}

\textbf{Pill Classification Accuracy.}
{To further investigate the robustness of the proposed PGPNet, we adopt the visualization techniques presented in~\cite{reliable} to understand the prediction accuracy ( of the pill classification task) better. In this technique, all models' predictions are categorized by their confidence scores into different bins, in which the average accuracy can be calculated. By observing the confidence-accuracy correlation, we can tell whether the models are under or over-confidence with their predictions~\cite{reliable}. Figure~\ref{fig:reliable_all}\subref{fig:reliable_frcnn} visualize those reliability plots of Faster R-CNN and PGPNet. It implies that both models have a propensity toward over-confidence, as the average accuracy of each confidence band is lower than the mean confidence score of that bin. However, that tendency is greatly alleviated in the circumstance of PGPNet, which means that the bins' heights are much closer to the perfect Confidence-Accuracy balance line (the red dashed diagonal line).
Figure~\ref{fig:reliable_all}\subref{fig:reliable_yolo} compares PGPNet's confidence-accuracy correlation and that of YOLOv5. With this backbone, we observed that the proposed PGPNet can produce predictions with a high level of reliability. All the heights of bins are much closer to values suggested by the perfectly-balanced line compared to Vanilla's result.}

\subsubsection*{Comparison with Existing Relavant Frameworks}
{Our work is the first to leverage an external graph in dealing with the Pill Detection challenge; thus, none of the preceding works are genuinely tight-correlated. Indeed, earlier researches only shared some common ground to our approach: (1) About methodology or (2) about research problem.

\noindent For the first group, there are works that utilized external information to solve the Object Detection problem. We adopt one of the most current studies with this direction - \cite{sgrn} to solve our targeted problem and serve as  \duy{a} baseline for PGPNet. 
Spatial-aware Graph Relation Network (SGRN) \cite{sgrn} is a framework that adaptively discovers and incorporates key semantic and spatial relationships for reasoning over each RoI.

\noindent With respect to research problem, as stated earlier, while there are many works which target single-pill detection problem ~\cite{wong2017development, usuyama2020epillid, ling2020few}, only a few directly solve the task of detecting multiple pills per image \cite{Ou2020, kwon_pill}. We attempt to adopt the most recent technique proposed in \cite{kwon_pill} as another baseline to compare with PGPNet. In the original work, the authors purpose is somewhat different from us, since they attempt to develop a framework which is solely trained on single-pill images, since they argued that the multi-pill dataset would scale up exponentially if the number of pills inscrease. This argument is not held in our intuition, and we believe, in reality, since the pills taken together have to be prescribed by pharmacists. We keep the pipeline as the original work, with some adoption for working with our VAIPE dataset: (1) Change Mask R-CNN to Faster R-CNN; (2) The training single-pill dataset is cropped from our VAIPE dataset with bounding box annonations; (3) The automate data labeling process are skipped. Since the original work did not name the proposed pipeline, we called it as \emph{Kwon's Pipeline} for short.
}

\textbf{Detection Performance.} 
\begin{table}[t]
\centering
\setlength\tabcolsep{5pt} 
\begin{tabular}{l|llllll}
\toprule
\textbf{\duy{Model}}           & \textbf{\duy{ mAP}}                                              & \textbf{\duy{ AP50}}                                             & \textbf{\duy{ AP75}}                     & \textbf{\duy{ APs}}                      & \textbf{\duy{ APm}}                                              & \textbf{\duy{ APl}}                      \\ \midrule
\duy{ Faster RCNN}     & \duy{ 63.7}                                             & \duy{ 86.6}                                             & \duy{ 76.9}                     & \duy{ 71.2}                     & \duy{ 58.1}                                             & \duy{ 64.6}                     \\
\duy{ SGRN}            & \duy{ 65.9}                                             & \duy{ 88.8}                                             & \duy{ 79.6}                     & \duy{ 76.3}                     & \duy{ 61.6}                                             & \duy{ 66.3}                     \\
\duy{ Kwon's Pipeline} & \duy{36.2} & \duy{38.5} & \duy{37.2} & \duy{30.3} & \duy{33.1} & \duy{36.0} \\
\duy{ PGPNet}          & \textbf{\duy{ 69.7}}                                             & \textbf{\duy{ 94.4}}                                             & \textbf{\duy{ 83.4}}                     & \textbf{\duy{ 90.0}}                     & \textbf{\duy{ 66.4}}                                             & \textbf{\duy{ 70.1}}                     \\ \bottomrule
\end{tabular}
\caption{Performance comparison of PGPNet with SGRN and Kwon's Pipeline.}\label{tab:eval_related}
\end{table}
Table \ref{tab:eval_related} summarizes the comparison of PGPNet, SGRN and Kwon's Pipeline when 
adopting the visual feature extractor architecture from Faster R-CNN with the Resnet-50-FPN model.
Clearly, SGRN outperforms the baseline Faster R-CNN in terms of overall performance but could not outperforms our proposed method PGPNet. 
Specifically, the mAP metrics achieved by SGRN is $65.9$, and PGPNet achieves the better score with a gap of nearly $4$. Upon other metrics, AP50, AP75, APs, APm, and  APl, PGPNet shows its superior by enhancing the performance from $5.1\%$ (e.g., in AP75 metrics) up to $17.1\%$ (e.g., in APs metrics). This is an expected result because SGRN reveals a major weakness when applying to the challenge of Pill Detection. The spatial relationships between pills in an image are arbitrary and frequently changed. Such noisy and unreliable information leads to the performance of SGRN being unstable and sometimes produce not good enough results. In the case of Kwon's Pipeline, the situation is even worse, since it cannot even beat the vanilla one-step Faster RCNN trained with mutple-pill VAIPE training set. The result of this pipeline is $43.1\%$ and $48.2\%$ worse than vanilla Faster R-CNN and PGPNet respectively. One reason for this deficiency is owing to the quality of its training data. There are many circumstances in which overlap or occlusion occurs, which make the cropped images also contain parts of other pills.

\textbf{Pill Classification Accuracy.} 
Figure \ref{fig:reliable_all}\subref{fig:reliable_sgrn} shows the correlation between the confidence and accuracy of PGPNet in the comparison with those of SGRN. Both the frameworks are based on the Faster R-CNN backbone and achieve similar results e.g., an over-confidence trend in every bin. All the predictions with confidence scores smaller than $0.2$ are totally unreliable (with $0$ accuracy). In addition, PGPNet also shows its superior over SGRN in some bins, in which the over-confidence situation is reduced effectively. We do not plot the Confidence-Accuracy of Kwon's Pipeline owing to space constraint and the obvious performance gap compared to our PGPNet.


\subsubsection*{Ability in Dealing with Hard Samples}
\begin{figure*}[!t]
\begin{minipage}{0.23\linewidth}
    \centering \small
        \includegraphics[width=\columnwidth]{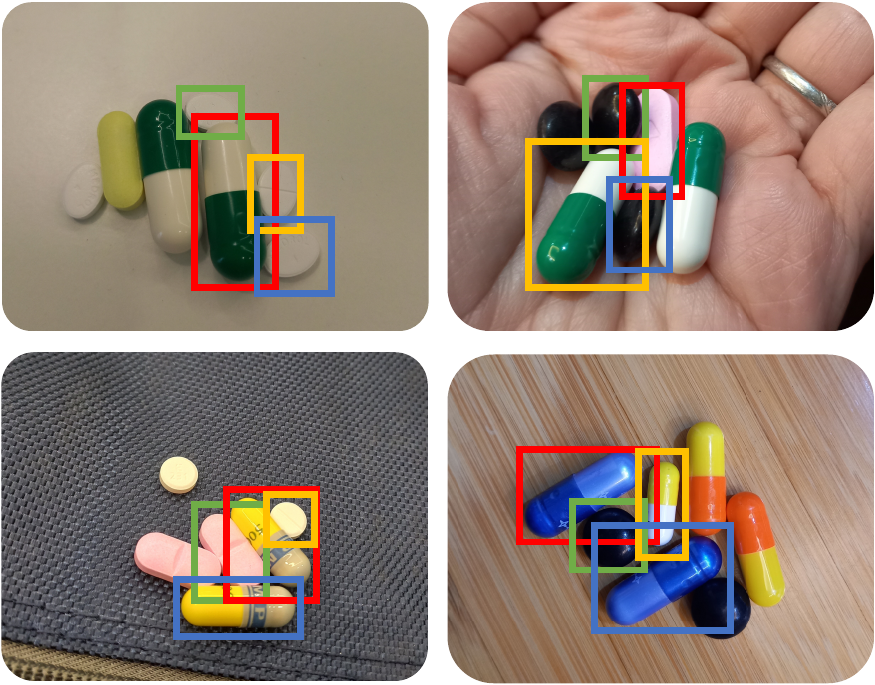}
        \caption{Images with occlusion phenomena in custom occlusion dataset. The rectangles depict examples of tablets with overlapping boundary boxes.
        \label{fig:heavy_occlusion}}
\end{minipage}
\hspace{0.1cm}
\begin{minipage}{0.3\linewidth}
    \centering
    \setlength\tabcolsep{3pt} 
    \resizebox{\columnwidth}{!}{%
        \begin{tabular}{l|cc||cc}
        \toprule
        \multicolumn{1}{c|}{\begin{tabular}[]{@{}c@{}} Test \\dataset \end{tabular}} & 
        \multicolumn{2}{c}{\begin{tabular}[]{@{}c@{}} Custom \\ Occlusion \end{tabular}} & 
        \multicolumn{2}{c}{\begin{tabular}[]{@{}c@{}} Non-\\Occlusion \end{tabular}} \\
        \midrule
        \multicolumn{1}{c|}{Method} & 
        \begin{tabular}[]{@{}c@{}} Faster \\ R-CNN \end{tabular} &  PGPNet & 
        \begin{tabular}[]{@{}c@{}} Faster \\ R-CNN \end{tabular} & PGPNet \\
        \midrule
        mAP     & 59.2 & \textbf{67.5} &   65.6 &   \textbf{71.7} \\
        AP50    & 76.5 & \textbf{81.1} &   87.4 &   \textbf{92.9} \\
        AP75    & 68.9 & \textbf{76.4} &   80.8 &   \textbf{87.0} \\
        APs     & -    & -             &   80.0 &   \textbf{90.0} \\
        APm     & 61.6 & \textbf{68.3} &   56.5 &   \textbf{64.7} \\
        APl     & 60.7 & \textbf{70.1} &   65.1 &   \textbf{70.6} \\
        \bottomrule
        \end{tabular}
    }
    \captionof{table}{Impact of heavy occlusion images on testing performance of PGPNet and Faster R-CNN.\label{tab:occlusion}}
\end{minipage}
\hspace{0.1cm}
\begin{minipage}{0.2\linewidth}
        \centering
        \includegraphics[width=\columnwidth]{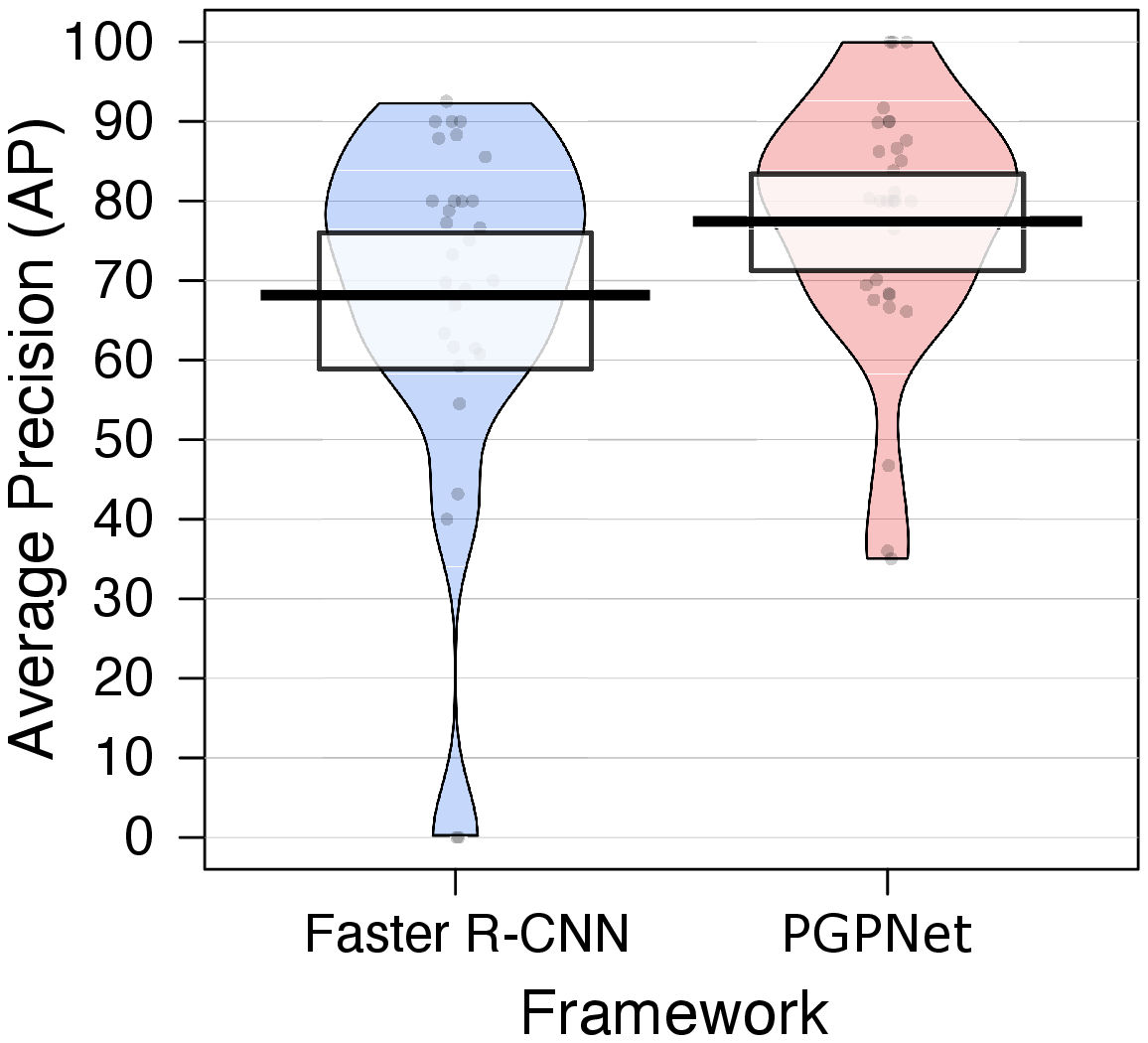}
        \caption{Comparison of PGPNet performance with Faster R-CNN over each individual class in occlusion dataset. \label{fig:occlusion}}
\end{minipage}
\hspace{0.1cm}
\begin{minipage}{0.2\linewidth}
    \centering
        \includegraphics[width=1\columnwidth]{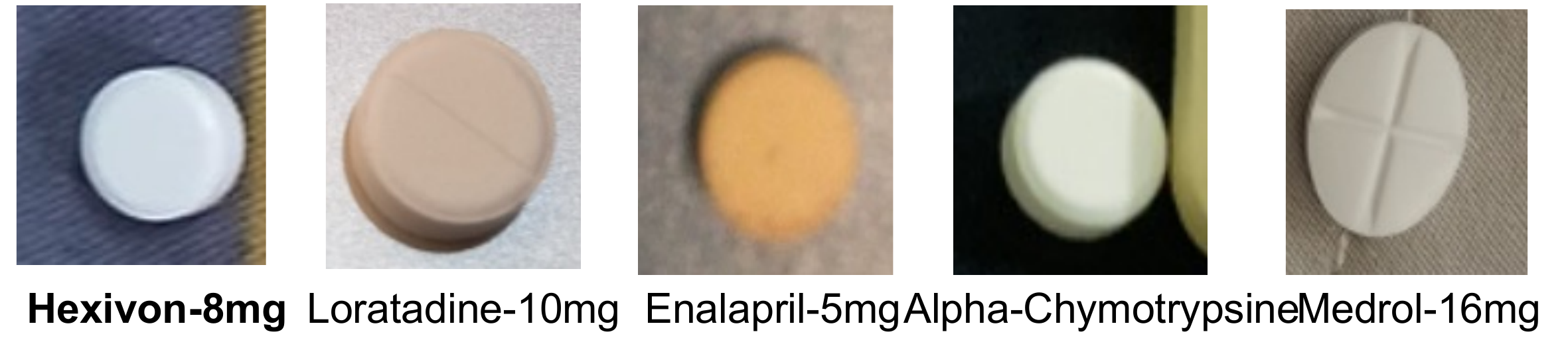}
        \caption{Some sample pills with very identical visual appearance with \emph{Hexinvon-8mg}\label{fig:misclassified}}
\end{minipage}
\vspace{-0.2cm}
\end{figure*}
{In the following, we investigate the ability of PGPNet in dealing with the occlusion phenomenon caused by overlapping pills, which is one of the most critical issues in dealing with multi-pill detection. 
To this end, we create a so-call \emph{custom occlusion sub-dataset} of VAIPE, which contains images with heavy occlusion phenomena, i.e., having at least two RoIs with the IoU beyond 30\% (Fig.\ref{fig:heavy_occlusion}).  We also create 
a custom \emph{custom non-occlusion sub-dataset}
which contains samples that are in the same classes that appear in the \emph{custom occlusion sub-dataset} but with no occlusion. 
The quantitative result is summarized in Table \ref{tab:occlusion}. 
The (-) mark in the table suggests the disregarded or unavailable metrics.
As the numbers suggest, even in cases where heavy occlusion occurs, PGPNet still shows its superior over Faster R-CNN. Specifically, the mAP over all classes in the \emph{custom occlusion sub-dataset} suggests a gap of $8.3\%$ between the two approaches. Interestingly, with the aid of classifier weight as the distinguishing characteristic for each class, PGPNet, even when dealing with occlusion cases still enhances the performance of $1.9\%$ compared to Faster R-CNN handling  the non-occlusion case (e,g, $67.5$ vs. $65.6$, respectively). Figure \ref{fig:occlusion} provides more information about the AP for each class in the \emph{custom occlusion sub-dataset}. PGPNet still outperforms Faster R-CNN in most cases with a large gap, and also produces a more reliable result by introducing a smaller variance over the AP metrics. }

\subsection*{PGPNet's Explainability}

\begin{figure*}[!t]
\begin{minipage}{0.32\linewidth}
        \centering
        \subfloat[Faster R-CNN]{ %
            \includegraphics[width=0.43\columnwidth]{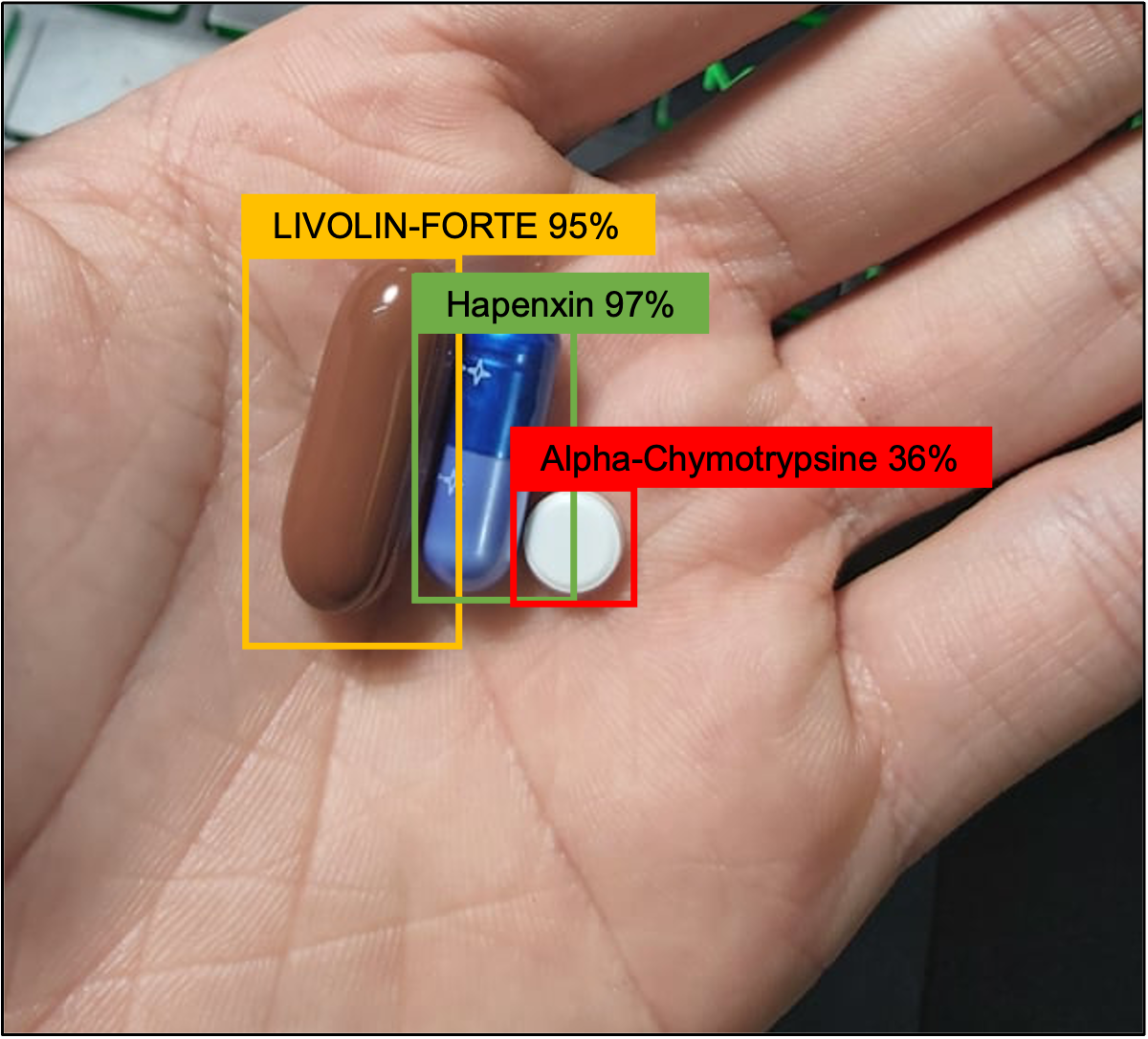}
            \label{fig:eval_rcnn}
        }
        \subfloat[PGPNet]{ %
            \includegraphics[width=0.43\columnwidth]{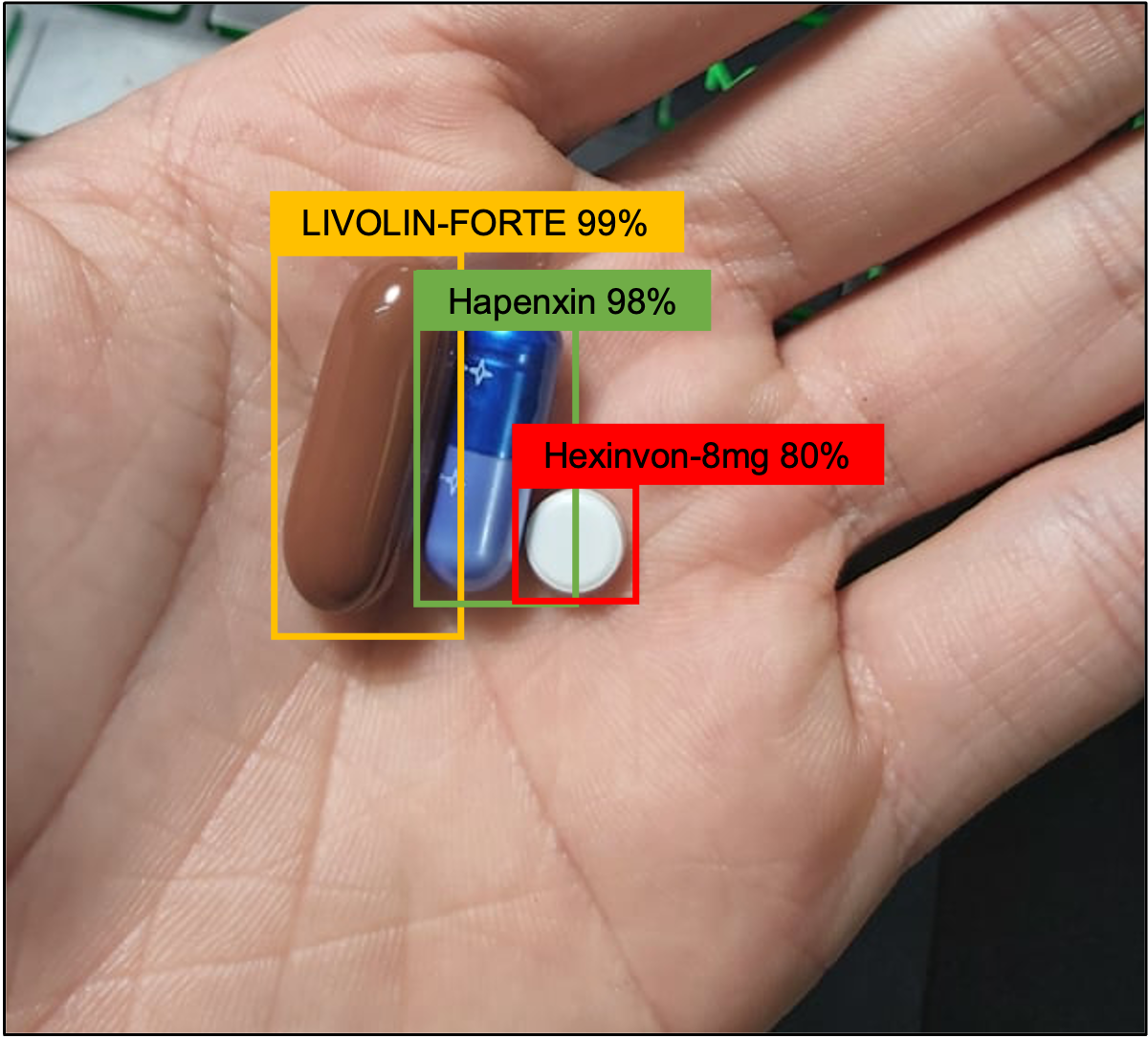}
            \label{fig:eval_kgp}
        }
        \caption[Predictions for a hard sample made by Faster R-CNN and PGPNet given the same image.]{Predictions for a hard sample made by Faster R-CNN and PGPNet given the same image.}
        \label{fig:kg_robust_instance}
        \vspace{0.1cm}
\end{minipage}
\hspace{0.1cm}
\begin{minipage}{0.64\linewidth}
        \centering \small
        \subfloat[Input]{ %
            \includegraphics[width=0.21\columnwidth,valign=c]{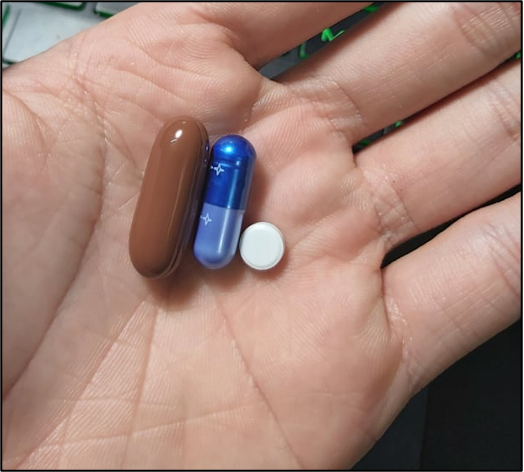}
            \label{fig:original}
        }
        \hspace{-0.15em}%
        \centering
        \subfloat[LIVOLIN-FORTE]{ %
            \includegraphics[width=0.21\columnwidth,valign=c]{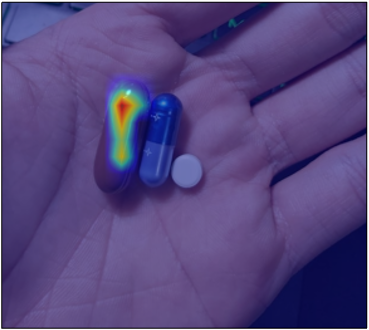}
            \label{fig:exp_1}
        }
        \hspace{-0.15em}%
        \subfloat[Hapenxin]{ %
            \includegraphics[width=0.21\columnwidth,valign=c]{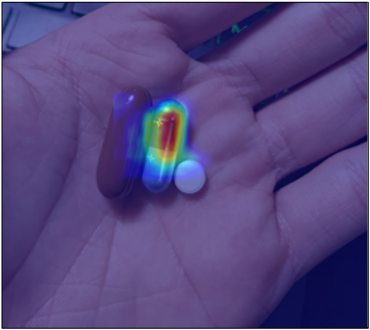}
            \label{fig:exp_2}
        }
        \hspace{-0.15em}%
        \subfloat[Hexinvon-8mg]{ %
            \includegraphics[width=0.21\columnwidth,valign=c]{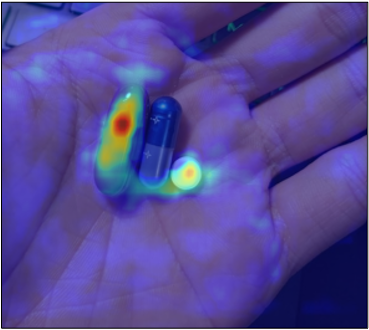}
            \label{fig:exp_3}
        }
        \caption{The saliency maps for each of the groundtruth labels included in the image instance. For simple samples (LIVOLIN-FORTE and Hapenxin), the classifier focuses on the exact location of the tablets to determine their identity. In contrast, for the hard case (Hexivon-8mg), information on both Hexivon and LIVOLIN-FORTE served as evidence. \label{fig:XAI_heatmap} }
\end{minipage}
\vspace{-0.5cm}
\end{figure*}


This section is dedicated to analyzing the results produced by PGPNet through a specific sample. This example demonstrates that the operation of PGPNet is very congruent with our initial motivation and that our designed architecture can materialize this motivation.



\subsubsection*{Experiment settings.} 
In this experiment, we choose a hard sample, namely \emph{Hexinvon-8mg}, with a relatively common appearance, for investigation. Figure \ref{fig:misclassified} visualizes \emph{Hexinvon-8mg} together with other pills in our dataset with almost identical visual appearance (round shape, white tint, etc.). 
As illustrated, these pills are readily confused with \emph{Hexinvon-8mg}.
Indeed, Fig.~\ref{fig:kg_robust_instance} depicts an example in which  \emph{Hexinvon-8mg} is miscategorized as \emph{Alpha-Chymotrypsine} by Faster R-CNN. 
Our PGPNet can, however, successfully distinguish \emph{Hexinvon-8mg} with a high confidence score.
In the following, we applied several Explainable AI techniques to explain the results inferred by our PGPNet. The image of interest consists of three pills: LIVOLIN-FORTE,  Hapenxin, and Hexivon as shown in Fig.~\ref{fig:kg_robust_instance}. 



\subsubsection*{Explanation of the Prediction Results}
We adopt the Excitation Backpropagation technique proposed by Zhang \cite{exb} to construct the saliency maps (Fig.\ref{fig:XAI_heatmap}), which indicate what the classifier has learned to produce the final results.
Firstly, for the easy samples, i.e., LIVOLIN-FORTE and Hapenxin, our model focuses precisely on those pill regions to make the prediction decision. 
In contrast, in the case of the hard sample, i.e., \emph{Hexinvon-8mg}, however, two regions are highlighted: one at the position of \emph{Hexinvon-8mg} and the other at the location of LIVOLIN-FORTE.
It indicates that the classifier solely requires information about LIVOLIN-FORTE and Hapenxin to identify these pills. Nevertheless, for \emph{Hexinvon-8mg}, the classifier must additionally incorporate information about its neighbor, i.e., LIVOLIN-FORTE.
This hypothesis is also supported by the Probabilistic score matrix shown in Fig.~\ref{fig:pseudo_scores}.
The probabilistic score matrix represents the prediction results generated by our Pseudo Classifier, which relies mainly on the pill's visual characteristics.
As demonstrated, Pseudo Classifier can accurately detect the proper labels of two simple samples, with their prediction scores approaching $1$,  and boost up their neighbors' probabilities (label ID $7$, $17$, etc.). However, with the case of \emph{Hexinvon-8mg}, the probability scores are relatively low, with all RoIs being investigated achieving scores of only about $0.3$.

Now, we utilize another explainable AI technique named GNNExplainer \cite{gnn_explainer} to investigate further the reason for identifying the hard sample, \emph{Hexinvon-8mg}. GNNExplainer is a model-agnostic architecture that can provide interpretable explanations for predictions of graph-based models. Specifically, GNNExplainer may identify a subgraph and a subset of node features that have a significant role in the prediction outcomes.
In our experiment, we treat our Graph Transformer Network as a module that produces regression output, i.e., the context vectors corresponding to all RoIs. For a more comprehensible result, we set the number of RoIs selected from the RPN module to ten, consisting of the five RoIs with the greatest \emph{objectness} scores and the other five with the lowest score. 
We utilize GNNExplainer to identify the sub-graph that contributes the most in recognizing \emph{Hexinvon-8mg}. 
The results are demonstrated in Fig.~\ref{fig:gnn_saliency}. 
In this figure, the white box depicts the RoI of \emph{Hexinvon-8mg}, the two orange boxes and blue boxes represent the RoIs of LIVOLIN-FORTE, and Hapenxin, respectively, while the five gray boxes indicate the RoIs of noise.
The black edges represent the vital connections, whose weights are proportionate to the width of the edges.
First, there are almost no edges between the nodes representing \emph{Hexinvon-8mg} and those of the noise RoIs.
It implies that the noise RoIs do not cue the prediction of \emph{Hexinvon-8mg}.
In contrast, there are bolded linkages between the RoIs of LIVOLIN-FORTE, Hapenxin, and \emph{Hexinvon-8mg}. 
These findings, along with the saliency map (Fig.\ref{fig:XAI_heatmap}), interpret that PGPNet has learned both the visual characteristic of the pill itself and the relationship between that pill and the others to make the final decision.

\begin{figure*}[!t]
\begin{minipage}{0.4\linewidth}
        \centering
        \includegraphics[width=0.9\columnwidth]{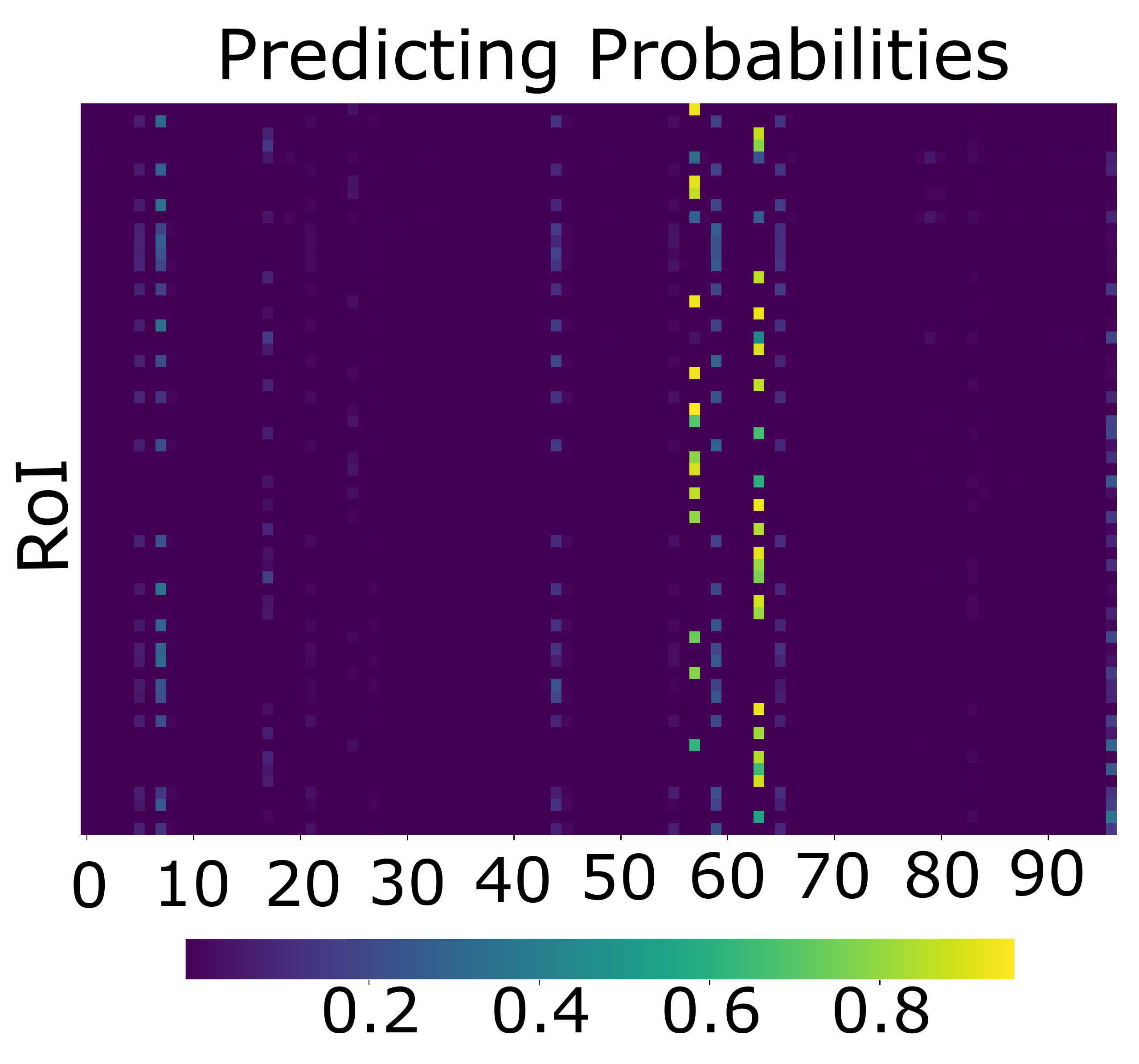}
        \caption{Probabilistic scores produced by PGPNet's Pseudo Classifier. \label{fig:pseudo_scores}}
\end{minipage}
\hspace{0.1cm}
\begin{minipage}{0.55\linewidth}
        \centering \small
        \subfloat[Bounding boxes of the RoIs.]{ %
            \includegraphics[width=0.4\columnwidth,valign=c]{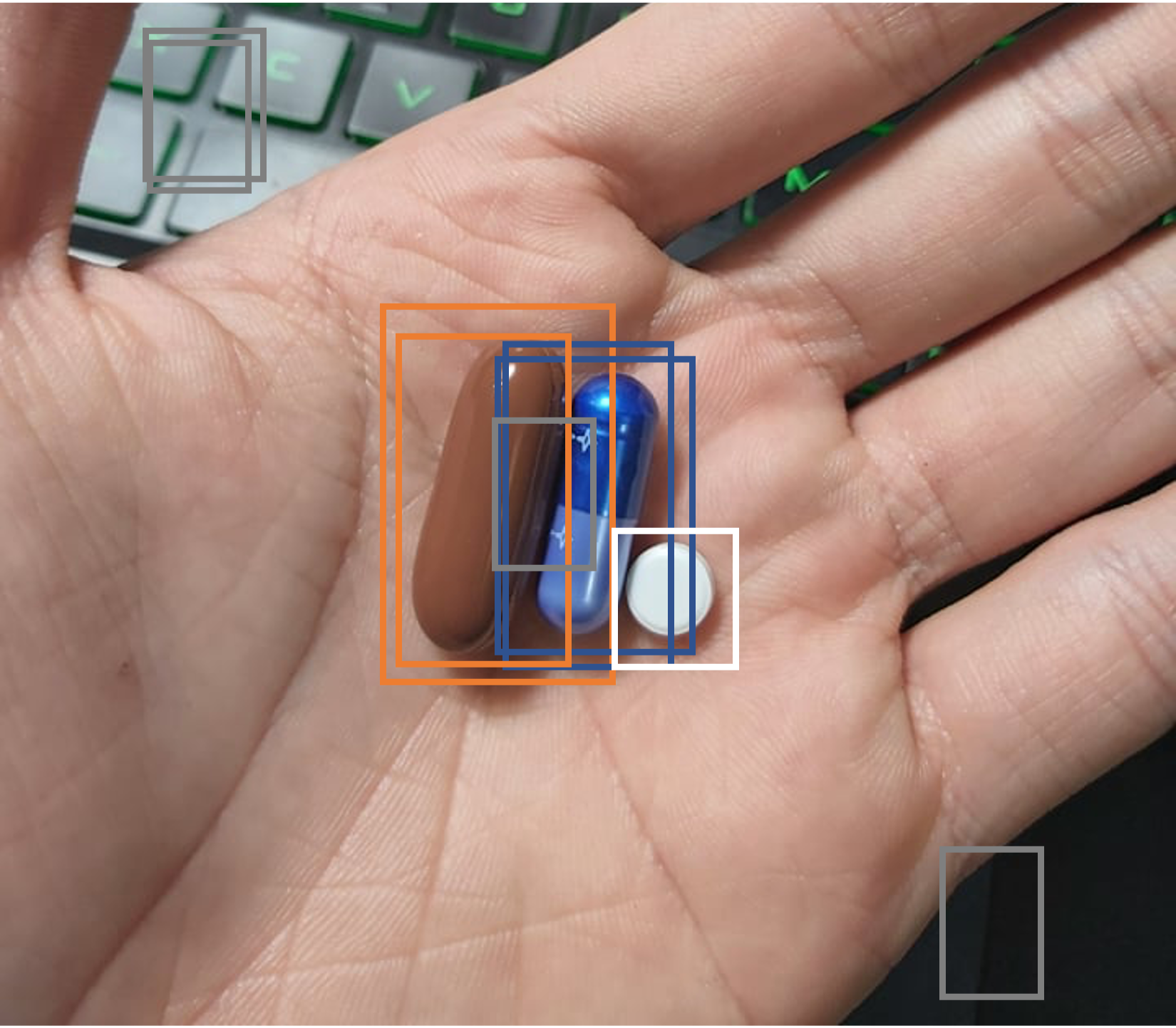}
            \label{fig:rpn_choose10}
        }
        \hfill%
        \centering
        \subfloat[Sub-graph identified by GNNExplainer.]{ %
            \includegraphics[width=0.5\columnwidth,valign=c]{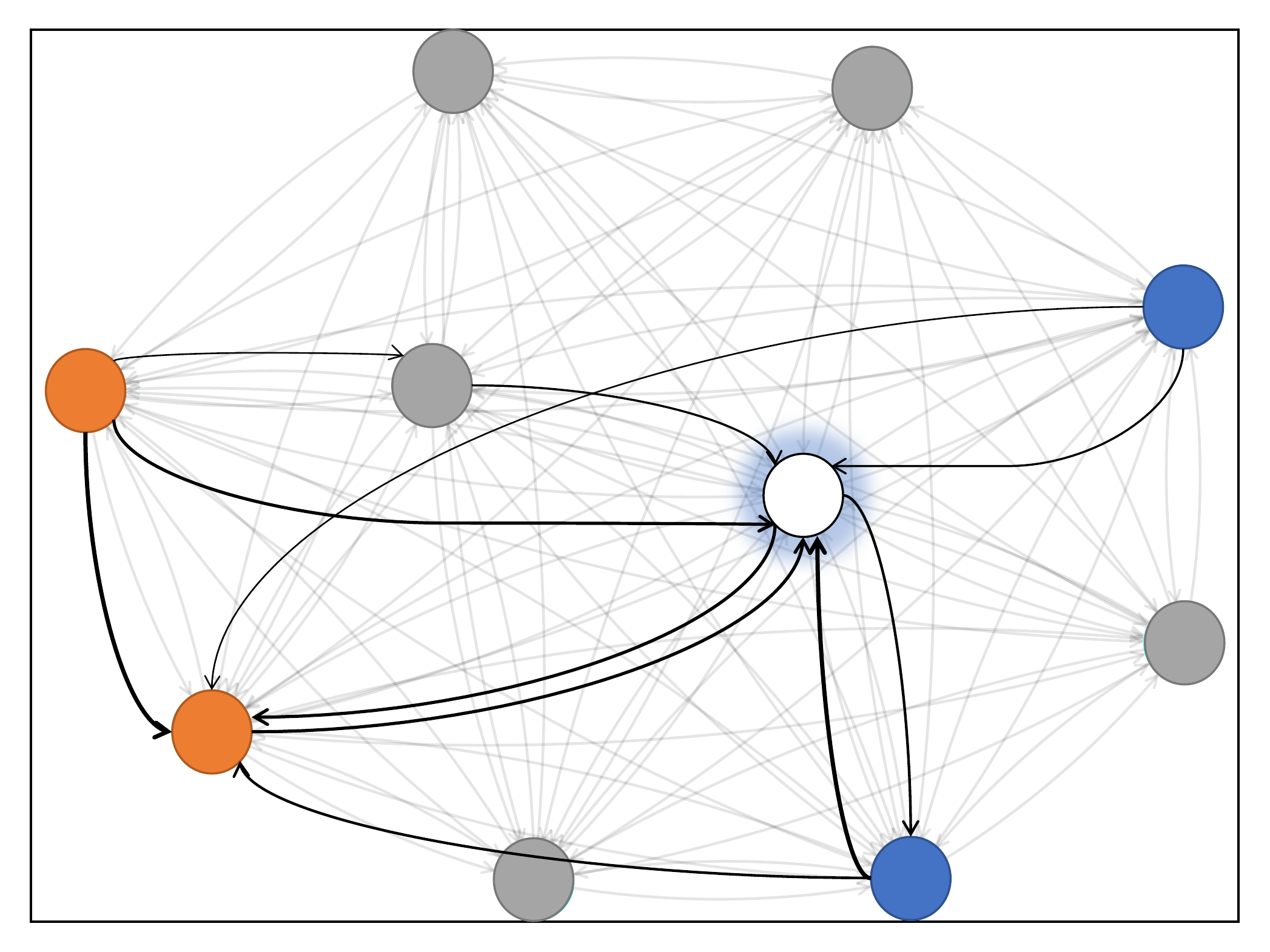}
            \label{fig:gnn_saliency}
        }
         \caption{Interpretation of the prediction result for \emph{Hexinvon-8mg} using GNNExplainer.  \textbf{(\subref{fig:gnn_saliency})} indicate the RoIs in \textbf{(\subref{fig:rpn_choose10})} most influential to the prediction of \emph{Hexinvon-8mg}. \label{fig:gnn_explain}} 
    
        
   
\end{minipage}
\vspace{-0.3cm}
\end{figure*}

\begin{figure}[tb]
    \begin{subfigure}[t]{0.48\columnwidth}
            \includegraphics[width=0.9\textwidth]{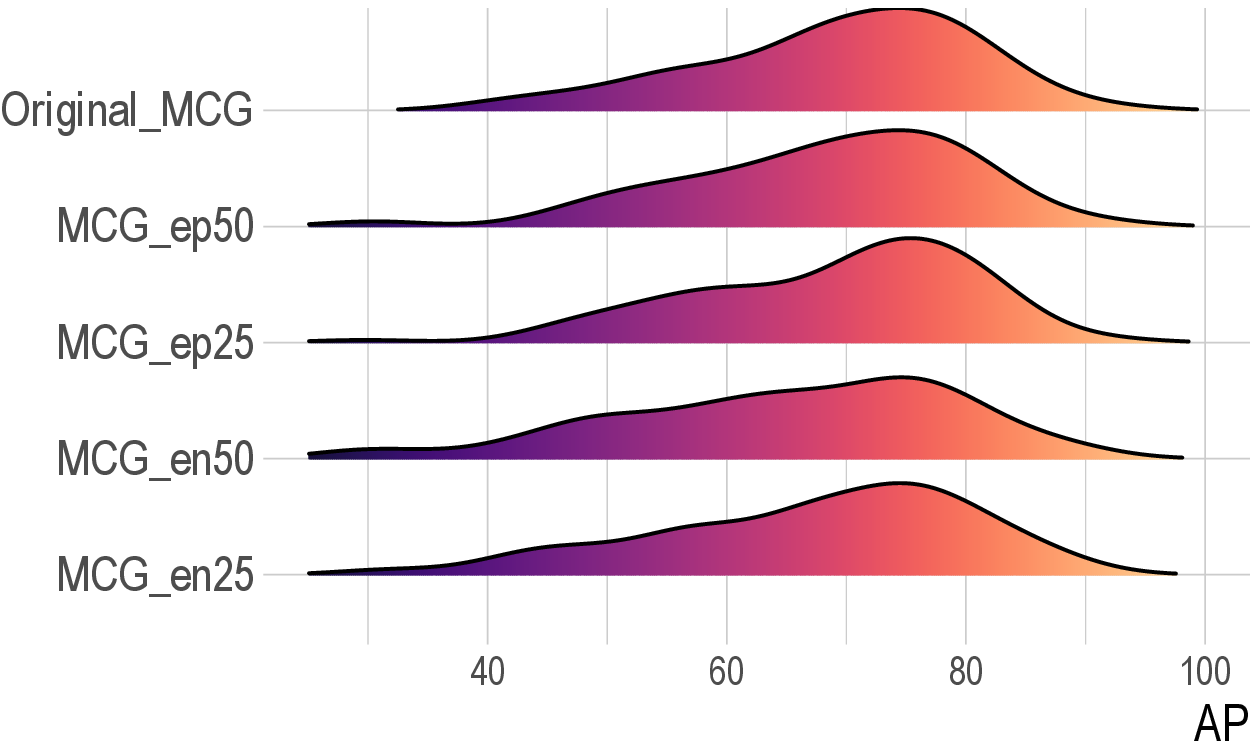}
            \caption{Distributions of Average Precision recorded over all classes produced by PGPNet with different MCG versions. \label{fig:edge_modify}}
    \end{subfigure}
    \hfill%
    \begin{subfigure}[t]{0.48\columnwidth}
        \includegraphics[width=0.9\columnwidth]{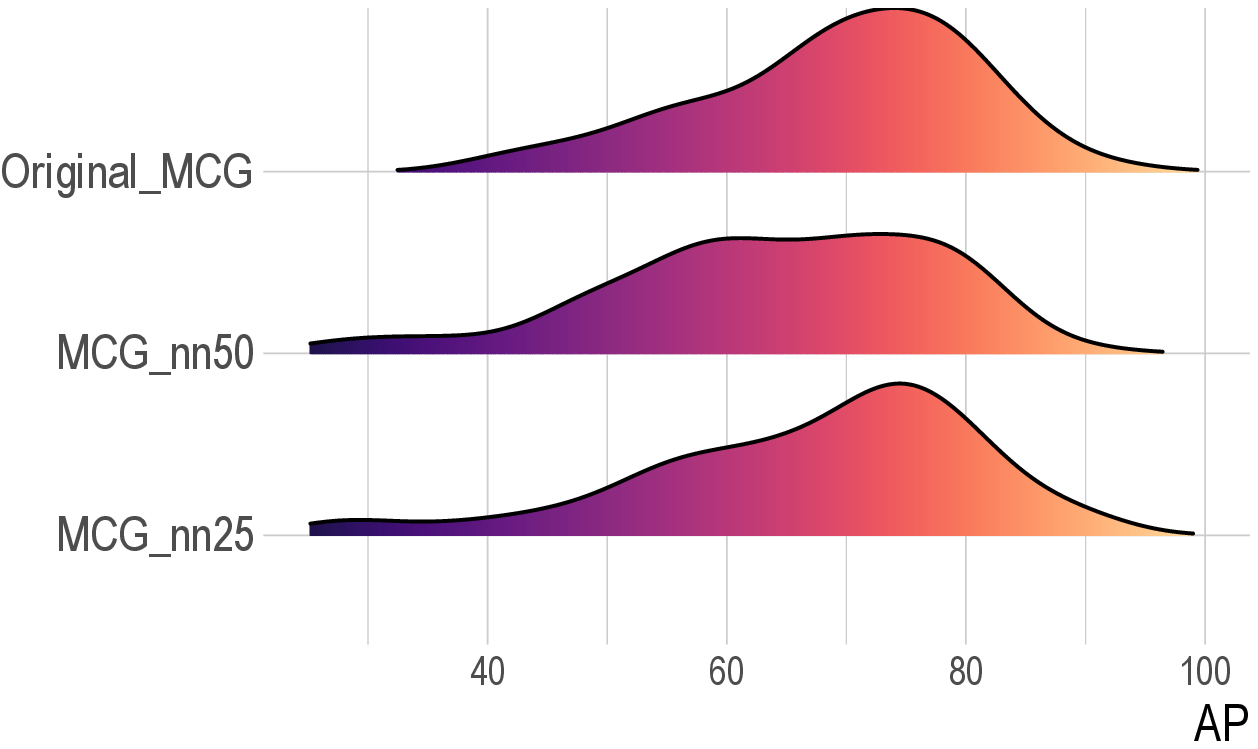}
        \caption{Distributions of Average Precision recorded over the classes in $N_A$ set produced by PGPNet with different MCG versions. \label{fig:node_modify}}
    \end{subfigure}
    \caption{Empirical result of node set and edge set modification.}
\end{figure}

\subsection*{Ablation Studies}
In this section, we perform extensive ablation studies to investigate the impacts of the main techniques proposed in our PGPNet and to investigate how each component in the proposed method helps to improve learning performance. Specifically, we alter the Co-occurrence Graph and observe how it affects the detection results in Section \nameref{sec:ablation_mcg}. We then assess the effects of using the relational graphs, the Graph transformer network, and the proposed auxiliary loss in Sections \nameref{sec:ablation_relational graphs}, \nameref{sec:ablation_gtn}, 
respectively.

\subsubsection*{Effect of Co-occurrence Graph's Quality}
\label{sec:ablation_mcg}
In this section, we perform two experiments to observe how the performance is changed when the nodes set and edges set of MCG are modified respectively.

\textbf{Edge Set Modification.}
We first observe the behavior of our PGPNet when adding noise edges and removing actual edges. We set up four scenarios which are the combinations of removing $25\%$ and $50\%$ of the edges in the set $E_1$, and adding a number of synthesized edges corresponding to $25\%$ and $50\%$ of the cardinality of $E_1$.

Figure \ref{fig:edge_modify} illustrates the performances of PGPNet with all Medical Co-occurrence Graph variances when being put into comparison with the original one. The performance here is denoted by the general metrics AP. As indicated by AP density, PGPNet with original MCG generates a more concentrated density with a smaller variance and a higher mean than other variances. In addition, when $50\%$ of edges are eliminated, the performance is clearly inferior to when $25\%$ of edges are eliminated. The figure concludes with the intriguing observation that eliminating edges at random would result in a greater performance decrease than adding noisy edges. This is because, even with the addition of noisy edges, PGPNet could still filter out unnecessary information through the training process. When excluding edges, the situation is different because the framework cannot learn the external knowledge contained in the eliminated edges.

\begin{figure*}[!t]
\begin{minipage}{0.32\linewidth}
        \includegraphics[width=0.9\columnwidth]{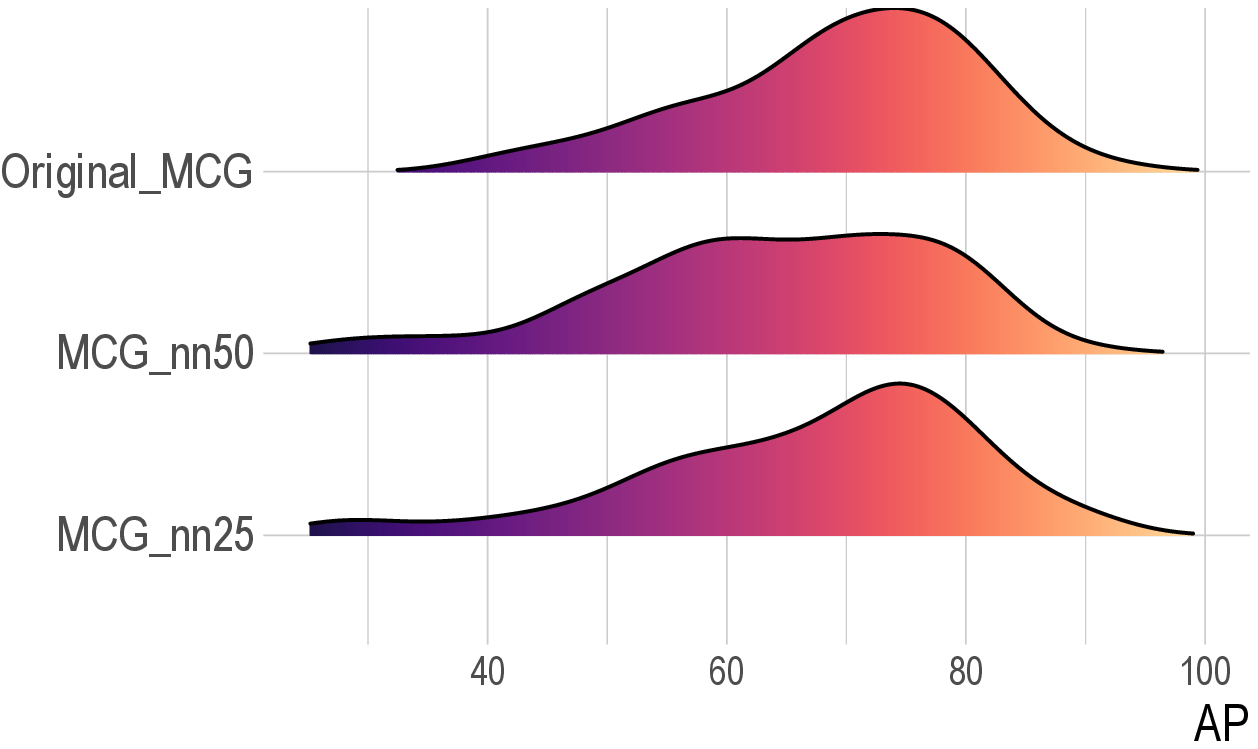}
        \caption{Distributions of Average Precision recorded over the classes in $N_A$ set produced by PGPNet with different MCG versions. \label{fig:node_modify}}
\end{minipage}
\hspace{0.1cm}
\begin{minipage}{0.68\linewidth}
    \centering
    \small
    \resizebox{\columnwidth}{!}{%
    \begin{tabular}{l|ccccc|l|l|l|l|l|l}
    \toprule
    \multicolumn{1}{l|}{} & \multicolumn{5}{c|}{\textbf{Component}}  & \multicolumn{6}{c}{\textbf{Performance}}
    \\
    & $\mathcal{G}_c$ & $\mathcal{G}_s$ & $\mathcal{G}_v$ & GTN & $\mathcal{L}_{aux}$ &   mAP & AP50 & AP75 & APs & APm & APl
    \\ \midrule
    Faster R-CNN & $\times$  & $\times$   & $\times$   & $\times$    & $\times$    &
      \begin{tabular}[c]{@{}r@{}}63.7 (-8.6)\end{tabular} &
      \begin{tabular}[c]{@{}r@{}}86.7 (-8.4)\end{tabular} &
      \begin{tabular}[c]{@{}r@{}}76.9 (-7.9)\end{tabular} &
      \begin{tabular}[c]{@{}r@{}}71.3 (-20.8)\end{tabular} &
      \begin{tabular}[c]{@{}r@{}}58.1 (-10.6)\end{tabular} &
      \begin{tabular}[c]{@{}r@{}}64.6 (-7.8)\end{tabular}  \\
    PGPNet-v1    & \checkmark  & $\times$   & $\times$   & $\times$    & $\times$     &
      \begin{tabular}[c]{@{}r@{}}65.9 (-5.5)\end{tabular} &
      \begin{tabular}[c]{@{}r@{}}91.9 (-2.9)\end{tabular} &
      \begin{tabular}[c]{@{}r@{}}79.6 (-4.6)\end{tabular} &
      \begin{tabular}[c]{@{}r@{}}72.5 (-19.4)\end{tabular} &
      \begin{tabular}[c]{@{}r@{}}62.3 (-4.3)\end{tabular} &
      \begin{tabular}[c]{@{}r@{}}66.0 (-5.8)\end{tabular}  \\
    PGPNet-v2    & \checkmark  & $\times$  & \checkmark  & \checkmark   & \checkmark   &
      \begin{tabular}[c]{@{}r@{}}66.9 (-3.9)\end{tabular} &
      \begin{tabular}[c]{@{}r@{}}92.1 (-2.7)\end{tabular} &
      \begin{tabular}[c]{@{}r@{}}81.1 (-2.8)\end{tabular} &
      \begin{tabular}[c]{@{}r@{}}80.0 (-11.1)\end{tabular} &
      \begin{tabular}[c]{@{}r@{}}61.3 (-5.9)\end{tabular} &
      \begin{tabular}[c]{@{}r@{}}67.6 (-3.6)\end{tabular} \\
    PGPNet-v3    & \checkmark  & \checkmark  & $\times$  & \checkmark   & \checkmark   &
      \begin{tabular}[c]{@{}r@{}}67.8 (-2.8)\end{tabular} &
      \begin{tabular}[c]{@{}r@{}}92.9 (-1.9)\end{tabular} &
      \begin{tabular}[c]{@{}r@{}}82.3 (-1.5)\end{tabular} &
      \begin{tabular}[c]{@{}r@{}}82.5 (-8.3)\end{tabular} &
      \begin{tabular}[c]{@{}r@{}}62.7 (-3.5)\end{tabular} &
      \begin{tabular}[c]{@{}r@{}}68.1 (-2.8)\end{tabular} \\
    PGPNet-v4    & \checkmark  & \checkmark  & \checkmark  & $\times$   & \checkmark  &
      \begin{tabular}[c]{@{}r@{}}68.4 (-1.9)\end{tabular} &
      \begin{tabular}[c]{@{}r@{}}92.6 (-2.2)\end{tabular} &
      \begin{tabular}[c]{@{}r@{}}81.7 (-2.1)\end{tabular} &
      \begin{tabular}[c]{@{}r@{}}80.0 (-11.1)\end{tabular} &
      \begin{tabular}[c]{@{}r@{}}64.4 (-1.0)\end{tabular} &
      \begin{tabular}[c]{@{}r@{}}68.8 (-1.8)\end{tabular}  \\
    PGPNet-v5    & \checkmark  & \checkmark & \checkmark  & \checkmark   & $\times$  &
      \begin{tabular}[c]{@{}r@{}}67.2 (-3.6)\end{tabular} &
      \begin{tabular}[c]{@{}r@{}}91.3 (-3.5)\end{tabular} &
      \begin{tabular}[c]{@{}r@{}}80.9 (-3.1)\end{tabular} &
      90.0 (+0.0) &
      \begin{tabular}[c]{@{}r@{}}62.2 (-4.3)\end{tabular} &
      \begin{tabular}[c]{@{}r@{}}67.3 (-3.9)\end{tabular}  \\
    PGPNet    & \checkmark  & \checkmark  & \checkmark  & \checkmark   & \checkmark   &
      69.7 &
      94.7 &
      83.5 &
      90.0 &
      65.0 &
      70.0 \\ \bottomrule
    \end{tabular}
    }
    \captionof{table}{Performance of PGPNet with the diferent combination of its components, i.e., when removing (marked as $\times$) / keeping (marked as \checkmark) the relational graph, GTN and auxiliary loss. Numbers inside the (.) represent the gap in percentage compared to the full version of PGPNet.\label{tab:ablation_result}}
\end{minipage}
\vspace{-0.5cm}
\end{figure*}

\textbf{Node Set Modification.}
To observe PGPNet's performance when the Medical Co-occurrence Graph lacks information on some specific nodes - classes, we design two different scenarios. In the first one, $25\%$ nodes are removed in the original graph, this set is denoted as $N_A$. For the latter, $50\%$ of nodes are eliminated, and the corresponding set $N_B$ is ensured to be a superset of $N_A$. The performances of PGPNet in two circumstances are compared with itself when having the full MCG, considering only the classes appeared in the set $N_A$.

Figure \ref{fig:node_modify} depicts the outcome of this experiment. The AP across all $N_A$ classes is used to evaluate performance here. As indicated by the graph, node removals also result in a significant decrease in model performance. More interestingly, the more nodes being eliminated, the greater drop is captured. Specifically, the AP density in case MCG contains only $50\%$ of remaining nodes has a great variance, with the mean value only around $60\%$. 

In the following, we study the effectiveness of the relational graphs, Graph Transformer Network (GTN) block, and auxiliary loss. 
The detailed configurations are presented in Table \ref{tab:ablation_result}. The $+$ sign indicates the presence of a component in a specific version, while $-$ denotes the opposite.  

\subsubsection*{Effects of the Relational Graphs}
\label{sec:ablation_relational graphs}
In this section, we study the effectiveness of the Size-graph and visual-based graph. 
To this end, we implement two simplified versions of PGPNet, namely PGPNet-v2 and  PGPNet-v3, in which we remove the Size-graph and visual-based graph, respectively. 
As shown in Table \ref{tab:ablation_result}, eliminating the Size-graph causes a decrease in performance from 3.9\% to 11.1\%, while omitting the visual-based graph reduces the accuracy from 2.8\% to 8.3\%.
An interesting finding is that the deterioration gap when removing the size graph is more significant than those when eliminating the visual-based graph in terms of all evaluation metrics. 
These findings imply the effectiveness of the Size-graph over the visual-based graph.
Moreover, it can be observed that mAP is the most impacted when the relational graphs are removed, followed by AP50, when comparing mAP, AP50, and AP75.
This can be explained as follows.
In AP75, we measure the precision of RoIs with the IoU beyond 75\%, which presumably has a high degree of confidence regarding the objective.
In contrast, when we reduce the IoU threshold, such as AP50 and mAP, the overlap area of the objective drops, resulting in a model with a significant degree of uncertainty.
In this case, integrating relational graphs provides additional data that reduces uncertainty, thereby boosting detection accuracy.
\subsubsection*{Effects of the Multi-modal Data Fusion Block and Auxiliary Loss}
\label{sec:ablation_gtn}
To investigate the effectiveness of the GTN, we implement PGPNet-v4, omitting the GTN block and relying solely on the GCN to learn the node representation. 
Results in Table~\ref{tab:ablation_result} reveal that GTN enhances the model's accuracy from $1.0\%$ to $11.1\%$. 
Comparing mAP, AP50, and AP75, AP50, and AP75 are slightly more influenced by GTN than mAP, but the gaps are trivial.
We employ PGPNet-v5, which eliminates the proposed auxiliary loss and compare its performance with the original PGPNet. 
As illustrated in Table \ref{tab:ablation_result}, adopting our auxiliary loss may result in a 3 to 4 percent performance gain for most evaluation metrics. 
In the final ablation study, we implement PGPNet-v1, which retains only the co-occurrence graph and removes all the other components.
As depicted in Table \ref{tab:ablation_result}, the detection accuracy degrades significantly, with a gap ranging from $2.9\%$ to $19.4\%$. However, even with this version, PGPNet is still superior to Faster RCNN, with a performance margin of up to $7.1\%$.

In conclusion, the PGPNet version with all components exhibits its superiority in all evaluation metrics. In addition, all versions of PGPNet are superior to the Faster R-CNN backbone, demonstrating the contribution of each component to the overall performance of PGPNet.

\begin{figure*}[!t]
\centering
\begin{minipage}{0.55\linewidth}
    \captionof{table}{\label{tab:dataset_meta} {An overview of existing public datasets for the task of image-based pill detection. To the best of our knowledge, the introduced VAIPE dataset is currently the largest dataset for pill identification, which was collected in real-world settings and came up with prescriptions.}}
    \centering \small
    	\setlength\tabcolsep{4pt} 
        \resizebox{\columnwidth}{!}{%
            \begin{tabular}{l|l|l|l}
                \toprule
                & \textbf{NIH} & \textbf{CURE}  & \textbf{VAIPE}            \\ \midrule
                Number of pill images     & 7,000    & 8,973  & 9,426             \\
                Number of pill categories & 1,000    & 196   & 96               \\
                Number of capture devices & 1    & 1   & $>$ 20               \\
                Instance per category     & 7       & 40-50 & $>$ 30 \\
                Illumination conditions   & 1       & 3     & $>$ 50 \\
                Backgrounds   & 1       & 6     & $>$ 50 \\
                Number of prescriptions   & 0       & 0     & 1,527            \\ \bottomrule
            \end{tabular}
        }
\end{minipage}
\hspace{0.1cm}
\begin{minipage}{0.3\linewidth}
    \captionof{table}{Details of training and testing datasets. \label{tab:train_test_split}}
    \centering \small
    	\setlength\tabcolsep{4pt} 
        \resizebox{\columnwidth}{!}{%
            \begin{tabular}{ll}
            \toprule
            \multicolumn{2}{c}{\textbf{Training dataset}} \\
            \cmidrule(lr){1-2}
            \multicolumn{1}{c}{Prescriptions} & \multicolumn{1}{c}{Images} \\ \midrule
            1,527 (100\%)             & 7,514 (78\%)    \\ \midrule
             \\
             \\
            \multicolumn{2}{c}{\textbf{Testing dataset}}  \\ 
            \cmidrule(lr){1-2} 
            \multicolumn{1}{c}{Prescriptions} & \multicolumn{1}{c}{Images} \\ \midrule
            0 (0\%) & 1,912 (22\%)     \\ 
            \bottomrule
            \\
            \end{tabular}
        }
\end{minipage}
\vspace{-10pt}
\end{figure*}
We conduct extensive experiments to validate the effectiveness of the proposed approach. 
In the following, we first introduce our in-house pill identification dataset, called VAIPE, which will be used to evaluate the proposed approach, and then explain our evaluation metrics and experimental settings. 
To assess the effectiveness of the proposed method, we conducted comparative assessments against a number of established models, including the detection backbones we selected, such as Faster R-CNN~\cite{fasterrcnn} and YOLOv5 \cite{yolov5}, as well as other related frameworks such as SGRN~\cite{sgrn} and the Mask RCNN-based approach described in ~\cite{kwon_pill}. We also perform ablation studies to investigate the efficiency of key components in our framework.
\subsection*{Dataset and Pre-processing}

\textbf{Motivation.}  {To the best of our knowledge, previous studies on the pill identification problem~\cite{tan2021comparison},~\cite{cure1},~\cite{Ling_2020_CVPR},~\cite{cure} only focus on datasets collected in constrained environments. For instance, existing datasets such as {NIH Dataset} \cite{nih_dataset} are constructed under ideal conditions in lighting, backgrounds, and equipment or devices. The CURE dataset~\cite{Ling_2020_CVPR} provides only one pill per image. Hence, these datasets do not reflect the real-world scenarios in which patients take an arbitrary number of drugs, and their environmental conditions (e.g., backgrounds, lighting conditions, mobile devices, etc.) are greatly varied. Additionally, many pills have nearly identical visual appearances. The fact that they appear alone in the images of these datasets will inevitably confuse the detection frameworks. Consequently, none of the existing datasets can be directly applied to the real-world pill detection problem or can only be applied with low reliability. There is no publicly available dataset of these pills images in which the pills follow intakes of actual patients. This limits the development of machine learning algorithms for the detection of pills from images as well as for building real-world medicine inspection applications. To address this challenge, we build and introduce a new, large-scale open dataset of pill images, which we called VAIPE. }

\textbf{Data Descriptor.} The VAIPE is a large-scale and open pill image dataset for visual-based medicine inspection. 
The dataset contains approximately 10,000 pill images that were manually collected in unconstrained environments. In this study, no hypotheses or new interventional procedures were generated. Also, no investigational products or clinical trials were used for patients. In addition, there were no changes in treatment plans for any patients involved. Pill images were retrospectives collected, and all identifiable information of patients was de-identified. Therefore, there was no requirement for ethics approval~\cite{bworld}. 

Pill images are collected in many different contexts (e.g., various backgrounds, lighting conditions, in-hand or out-of-hand, etc.) using smartphones. These images are then manually labeled using the information from the relevant prescriptions. In summary, the number of pills per image is about $5 - 10$, and the total number of pill images collected was $9,426$ pill images with $96$ independent pill labels. To train the proposed deep learning system, the pill images from the VAIPE dataset are resized so that the shortest edges have a size of $800$, with a limit of $1,333$ on the longer edge. The ratios are kept the same as the original images if the max size is reached, then downscale so that the longer edge does not exceed $1,333$.

\textbf{Data Validation.} Patient privacy was controlled and protected. In particular, all images were manually reviewed to ensure that all individually identifiable health information of the patients has been removed to meet the General Data Protection Regulation (GDPR) \cite{gdpr}. Annotations of pill images were also carefully examined. Specifically, all images were manually reviewed case-by-case by a team of 20 human readers to improve the quality of the annotations.

\textbf{Comparison with Existing Datasets.} Table \ref{tab:dataset_meta} provides a summary of the aforementioned datasets (including NIH, CURE, and VAIPE) together with other ones of moderate sizes, meta-data, and other properties. Compared to the two previous datasets, the VAIPE dataset is constructed under a much more flexible procedure that reflects the characteristic real-world data distributions. Hence, the introduced dataset can serve as a reliable data source for training \emph{generic pill detectors}.

\subsection*{Evaluation Metrics}
{We evaluate the proposed method and other related works by the COCO APs metrics~\cite{coco_ap}. This set of metrics is widely accepted and used for evaluating state-of-the-art object detectors. Mean Average Precision (mAP), as its name suggests, is the mean of Average Precision (AP) overall $C$ classes and all the targeted IoU thresholds in the threshold set $T$ calculated by $ mAP = \frac{1}{C |T|} \sum_{i}^{C} \sum_{t \in T} AP_{i,t}$, 
where Average Precision ($AP_{i,t}$) is the area under the Precision-Recall curve, calculated for the class $i$ at a given IoU threshold $t$.} 
\subsection*{Comparison with state-of-the-art methods}
\textbf{Comparison Benchmarks.} To show the effectiveness of the proposed method, we conducted a comparison with the state-of-the-art object detectors, including our detection backbones: Faster R-CNN~\cite{fasterrcnn}, YOLOv5 \cite{yolov5}, and related works: SGRN~\cite{sgrn}, Mask RCNN-based approach \cite{kwon_pill}. Throughout the literature, the baseline with which PGPNet presently integrates is Faster R-CNN \cite{fasterrcnn}; hence, the original framework is utilized for our comparison. 
We adopt two different CNNs and one Transformer-based module for visual feature extractor, namely ResNet-50-C4, ResNet-50-FPN and Swin Transformer V2 - SwinV2 \cite{swinv2} (Fig.~\ref{fig:generalflow}). Specifically, for two ConvNets, we use a single feature map produced by convolution block C4 of the ResNet-50 model in ResNet-50-C4. In ResNet-50-FPN, we replace C4's feature map with multi-scale feature maps produced by Feature Pyramid Network (FPN)~\cite{fpn}. As for the Swin Transformer module, we are currently utilizing the SwinV2-T configuration \cite{swinv2} to ensure that the number of model parameters is comparable to that of ResNet-50.
In addition, we also make adaption for PGPNet with YOLOv5 \cite{yolov5} detection backbone. Two configurations of YOLOv5s and YOLOv5n are currently adopted. 
Also, the most relevant frameworks compared with our PGPNet are also put into comparison: a representative approach that utilizes an external knowledge graph for Object Detection task~\cite{sgrn}; a Mask RCNN-based baseline that also proposed to solve the same task of multi-pill detection \cite{kwon_pill}. \\
\noindent For a fair comparison, a fixed set of hyper-parameters is used for PGPNet throughout all experiments.}

\subsection*{Implementation Details}
{We conduct all the experiments using the Pytorch (version 1.10.1) on an Intel Xeon Silver 4210 2.20GHz system with $2$ $\times$ NVIDIA GeForce RTX 3090 GPUs.  We train and test all targeted models on the training and testing sub-datasets provided in Table~\ref{tab:train_test_split}. Specifically, we initialize all the networks with the weights achieved by pre-training them on COCO 2017 dataset~\cite{coco}. We then train the models in $20,000$ iterations with a batch size of $16$. AdamW~\cite{adamw} optimizer is used with the initial learning rate of $0.001$. We also augment the training data by using simple techniques such as random horizontal and vertical flips to prevent overfitting. For our PGPNet implementation, we set the dimensions of node embeddings at $64$. We also design the Graph Transformer Module with only one layer and $10$ channel set.}

\section*{Experimental Results}
This section reports our experimental results. We evaluate the effectiveness of PGPNet in three aspects: robustness, reliability, and explainability. The details are described below. 


\subsection*{Robustness and Reliability of PGPNet}
\label{subsec:eval_backbones}
\subsubsection*{Comparison with Faster R-CNN and YOLOv5}
\begin{table}[!tb]
Comparing PGPNet's detection performance with state-of-the-art vanilla object detectors on VAIPE dataset.
Best results are highlighted in \textbf{bold} text. \label{tab:eval_backbone}
\centering
\small
\setlength\tabcolsep{3pt} 
\resizebox{0.8\columnwidth}{!}{%
\begin{tabular}{lll|cccccc}
\toprule
\multicolumn{3}{c|}{\textbf{Method}}                                      & \multicolumn{1}{c}{\textbf{mAP}} & \multicolumn{1}{c}{\textbf{AP50}} & \multicolumn{1}{c}{\textbf{AP75}} & \multicolumn{1}{c}{\textbf{APs}} & \multicolumn{1}{c}{\textbf{APm}} & \multicolumn{1}{c}{\textbf{APl}} \\ \midrule
\parbox[t]{2mm}{\multirow{7}{*}{\rotatebox[origin=c]{90}{\textcolor{red}{\textbf{Two-step}}}}} & 
\multicolumn{1}{l|}{\multirow{2}{*}{
    \begin{tabular}[l]{@{}l@{}} Faster R-CNN \\ (ResNet-50-C4)\end{tabular}}}  
    & Vanilla & 62.6               & 87.0                 & 74.4                  & 75.0                     & 58.3                 & 62.9                 \\
\multicolumn{2}{l|}{}                               & PGPNet       & \textbf{68.3 (+9.2\%)}       & \textbf{92.5}        & \textbf{81.7}          & \textbf{80.0}             & \textbf{64.3}         & \textbf{68.7}         \\
\cmidrule(lr){2-9}
& \multicolumn{1}{l|}{\multirow{3}{*}{\begin{tabular}[l]{@{}l@{}} Faster R-CNN \\ (ResNet-50-FPN)\end{tabular}}} & Vanilla & 63.7                & 86.6                & 76.9                   & 71.2                 & 58.1                & 64.6               \\
\multicolumn{2}{l|}{}                               & PGPNet       & \textbf{69.7 (+9.4\%)}       & \textbf{94.4}         & \textbf{83.4}         & \textbf{90.0}    & \textbf{66.4}        & \textbf{70.1}   \\
\multicolumn{2}{l|}{} &    &                 &                    &                   &                    &                   &                   \\ 
\cmidrule(lr){2-9}
& \multicolumn{1}{l|}{\multirow{2}{*}{\begin{tabular}[l]{@{}l@{}} \duy{Faster R-CNN} \\ \duy{(SwinV2-T)}\end{tabular}}} & \duy{Vanilla} & \duy{59.7}                & \duy{84.5}                & \duy{72.3}                   & \duy{66.9}                 & \duy{54.0}                & \duy{60.1}               \\
\multicolumn{2}{l|}{}                               & \duy{PGPNet}       & \duy{\textbf{62.6 (+4.8\%)}}       & \duy{\textbf{87.2}}         & \duy{\textbf{75.5}}         & \duy{\textbf{68.6}}    & \duy{\textbf{56.6}}        & \duy{\textbf{62.9}}   \\
\midrule
\parbox[t]{2mm}{\multirow{4}{*}{\rotatebox[origin=c]{90}{{\textbf{One-step}}}}} & \multicolumn{1}{l|}{\multirow{2}{*}{YOLOv5n}}  & Vanilla & 37.9 & 50.8 &45.4    & 87.5    & 49.1    & 38.3  \\
\multicolumn{2}{l|}{}                              & PGPNet  & \textbf{43.0 (+12.0\%)}   & \textbf{58.4}    & \textbf{51.3}    & \textbf{82.5}    & \textbf{52.4}    & \textbf{43.7}    \\ \cmidrule(lr){2-9}
& \multicolumn{1}{l|}{\multirow{2}{*}{YOLOv5s}} & Vanilla & 57.5 & 75.8 & 68.3 & 85.0 & 58.3 & 57.0 \\
\multicolumn{2}{l|}{}                              & PGPNet  & \textbf{63.4 (+10.2\%)} & \textbf{85.9} & \textbf{76.4} & \textbf{89.9} & \textbf{58.3} & \textbf{64.1} \\
\bottomrule
\end{tabular}
}
\vspace{-5pt}
\end{table}

 \begin{figure}[!tb]
        \centering
        \includegraphics[width=0.85\columnwidth]{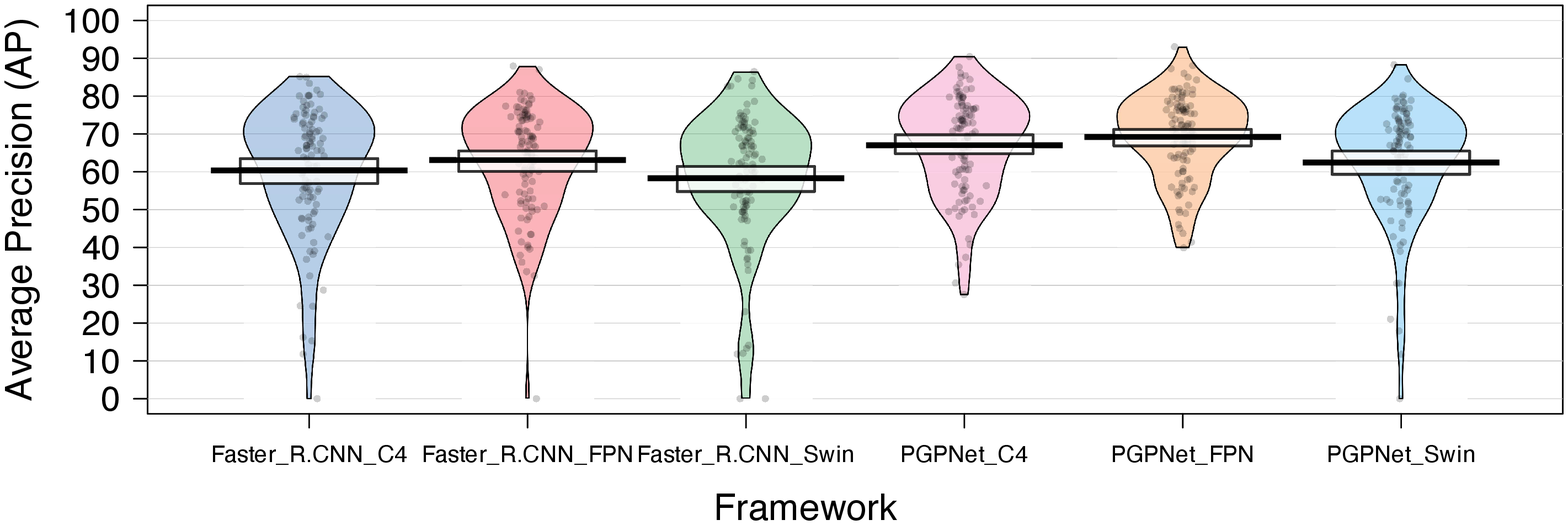}
        \caption{Comparison of the PGPNet performance with the Faster R-CNN baseline over each individual class. \label{fig:eval_backbone_class}}
    \end{figure}

\begin{figure}[tbh]
    \captionsetup[subfigure]{justification=centering}
    \centering
    \subfloat[Faster R-CNN \label{fig:reliable_frcnn}]{%
      \includegraphics[clip,width=0.31\columnwidth]{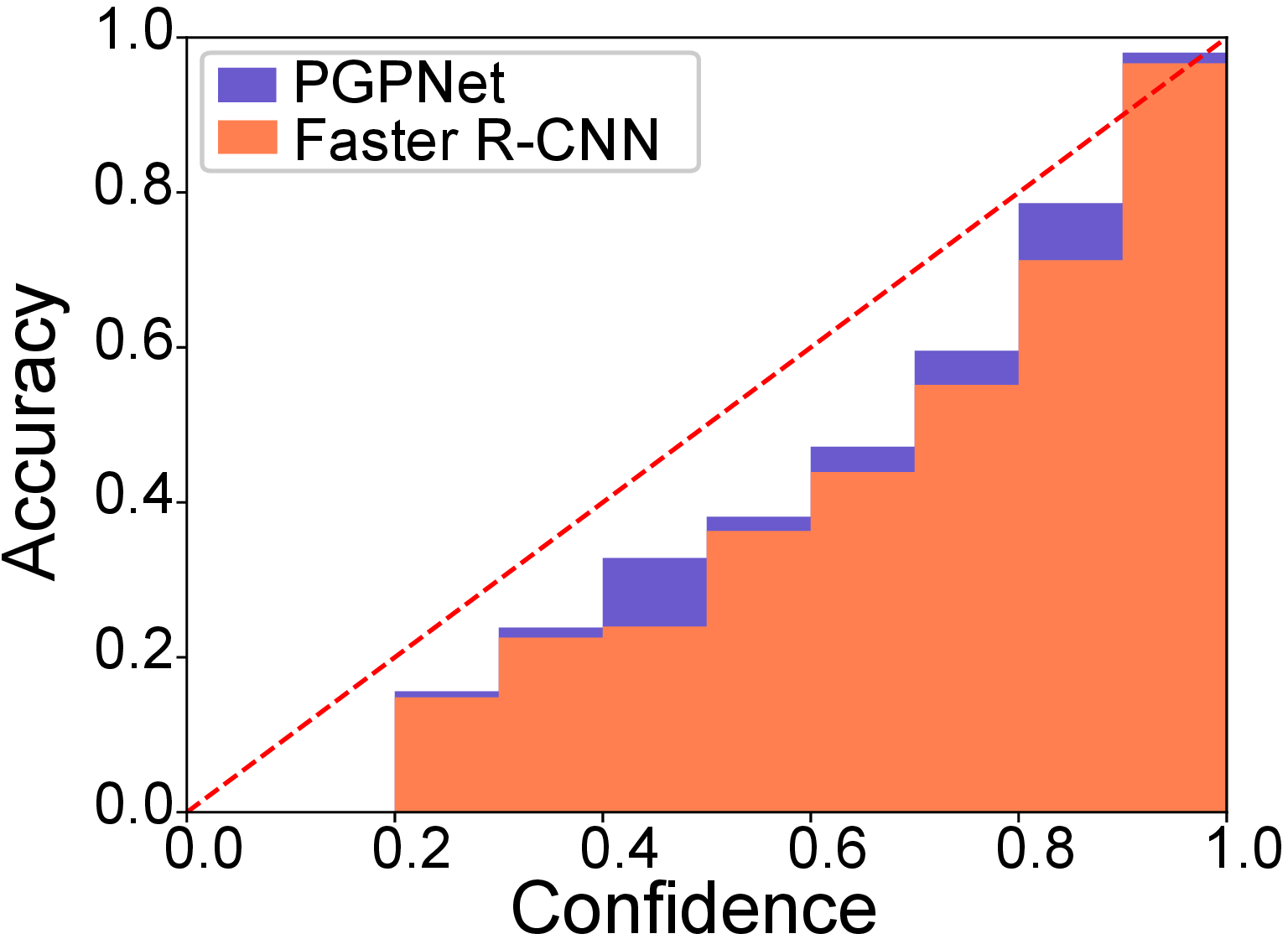}%
    }%
    \hfill%
    \subfloat[YOLOv5 \label{fig:reliable_yolo}]{%
      \includegraphics[clip,width=0.31\columnwidth]{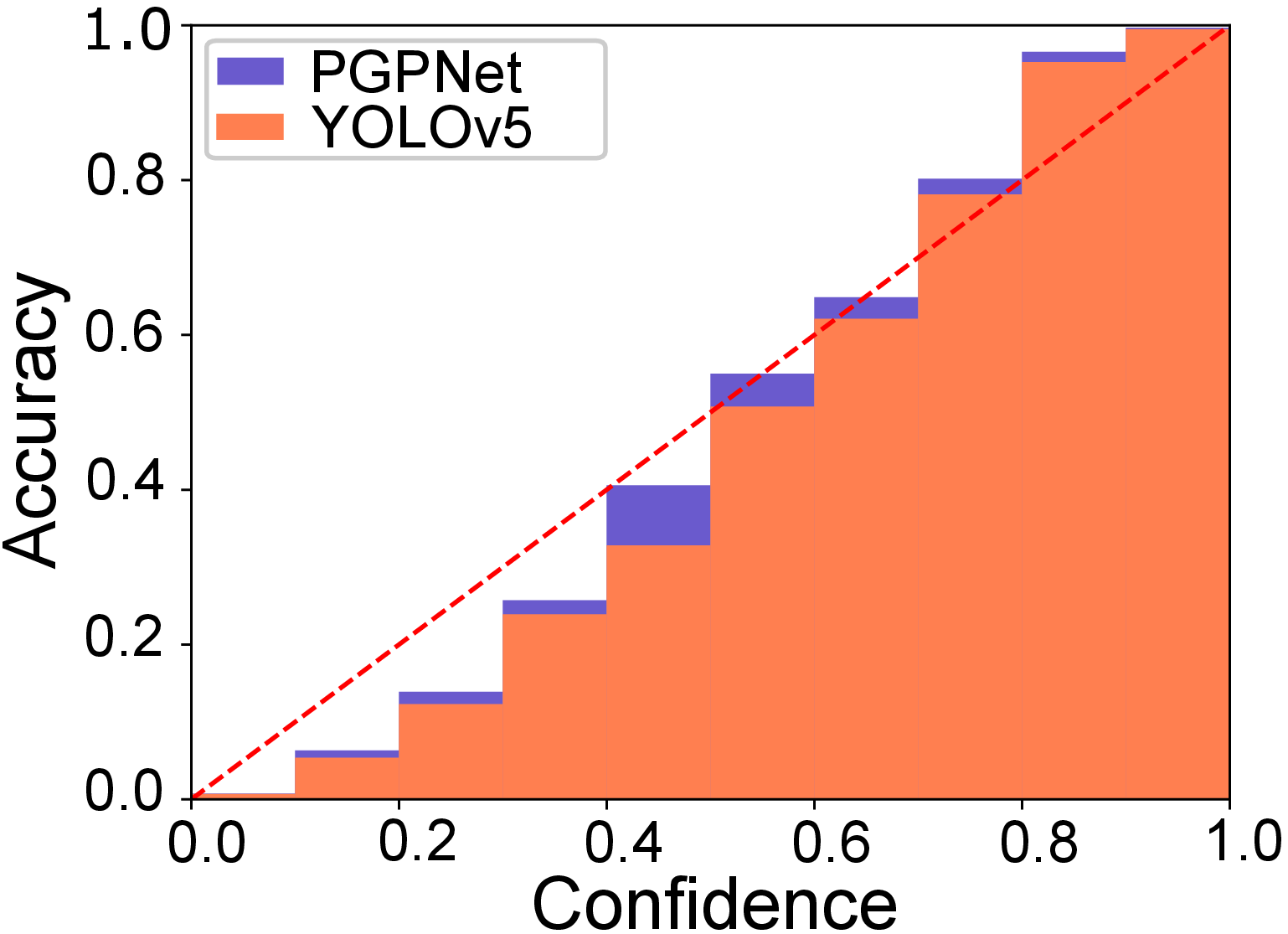}%
    }%
    \hfill%
    \subfloat[SGRN \label{fig:reliable_sgrn}]{%
      \includegraphics[clip,width=0.31\columnwidth]{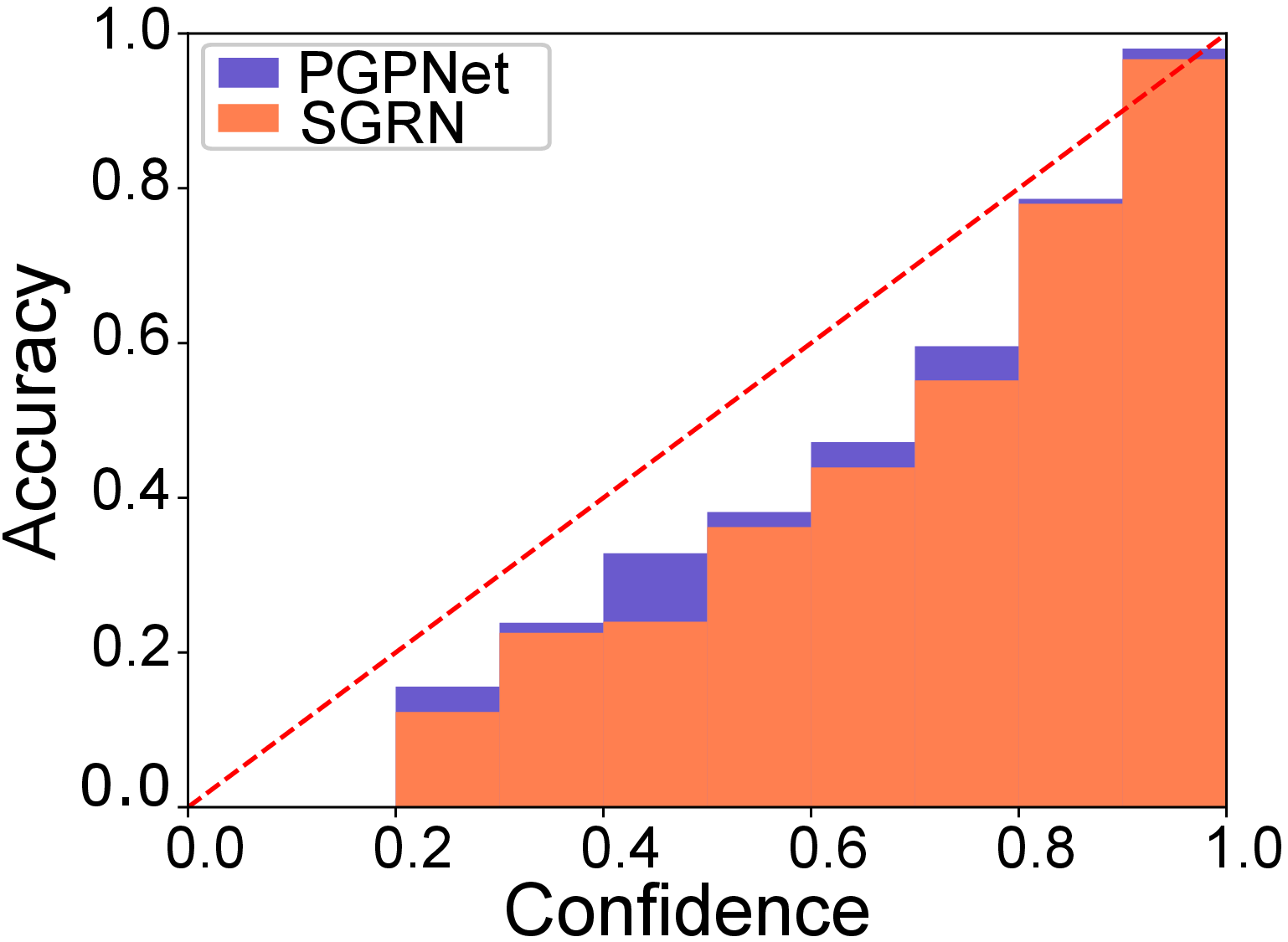}%
    }%
    \caption{Reliability investigation for PGPNet and different baseline performances.\label{fig:reliable_all}}
\end{figure}
    
\textbf{Detection Performance.}
Table~\ref{tab:eval_backbone} shows the experimental results of PGPNet and the state-of-the-art object detectors framework (Vanilla), e.g., Faster R-CNN (two-step detector), and YOLOv5~\cite{yolov5} (one-step detector) 
on the VAIPE dataset. As shown, PGPNet obtained better results than Faster R-CNN by large performance gaps for all evaluation metrics. Specifically, when using the ResNet-50-C4 model as the visual feature extractor model, the average precision mAP of Faster R-CNN was $62.6$, while that of PGPNet was $68.3$. The proposed method improves the performance over the baseline Faster R-CNN by $9.2\%$. 
Under strict metrics, e.g., AP75, PGPNet also outperforms Faster R-CNN $8-9\%$.
In addition, we observed similar behavior when using the ResNet-50-FPN model. The proposed PGPNet makes an improvement of $9.4\%$ for the mAP metrics. With a Transformer-based backbone, here a Swin Transformer V2 configuration - SwinV2-T ~\cite{swinv2}, the results are slightly worse compared to those produced by ResNet-based counterparts, for both the vanilla or PGPNet alternatives. However, PGPNet still show its superior when being install with this backbone, as the empirical result for AP metrics is improved by $4.8\%$ compared to the vanilla SwinV2-T Faster R-CNN model.

{For YOLOv5, PGPNet outperformed Vanilla by a significant margin across all performance metrics in both YOLO instances. Specifically, the average precision AP of the vanilla model with YOLOv5n was $37.9$ while that of PGPNet was $43.0$ ($12\%$ improvement). In the case of a larger alternative, YOLOv5s, a similar conclusion can be drawn, namely that PGPNet improves overall mAP metrics by $5.9$, e.g., $10.2\%$.}


Figure \ref{fig:eval_backbone_class} visualizes the AP 
for all classes in the dataset when using Faster R-CNN as the backbone. The first three bins denote Faster R-CNN alternatives, and the later three are the corresponding PGPNet configurations. The dots in the figure represent AP values for classes; the vertical line is the indicator for the mean value, while the rectangle bar is the $95\%$ High-Density Interval (HDI) band. Apart from the fact that the mean AP over all classes of PGPNet variances is better than those produced by Faster R-CNN, we found that PGPNet also has more reliable and stable results over all classes.  
Specifically, PGPNet helps to improve the AP of classes that Faster R-CNN frequently confuses (the points with low APs in the blue and pink beans). As a result, the three beans of Faster R-CNN exhibit a large variance, i.e.,  the AP ranged from $0$ to around $90$. In contrast, the beans of PGPNet performance are more condensed and have shorter tail, i.e., the AP ranged from  $40$ (or $50$) to around $90$.

\textbf{Pill Classification Accuracy.}
{To further investigate the robustness of the proposed PGPNet, we adopt the visualization techniques presented in~\cite{reliable} to understand the prediction accuracy ( of the pill classification task) better. In this technique, all models' predictions are categorized by their confidence scores into different bins, in which the average accuracy can be calculated. By observing the confidence-accuracy correlation, we can tell whether the models are under or over-confidence with their predictions~\cite{reliable}. Figure~\ref{fig:reliable_all}\subref{fig:reliable_frcnn} visualize those reliability plots of Faster R-CNN and PGPNet. It implies that both models have a propensity toward over-confidence, as the average accuracy of each confidence band is lower than the mean confidence score of that bin. However, that tendency is greatly alleviated in the circumstance of PGPNet, which means that the bins' heights are much closer to the perfect Confidence-Accuracy balance line (the red dashed diagonal line).
Figure~\ref{fig:reliable_all}\subref{fig:reliable_yolo} compares PGPNet's confidence-accuracy correlation and that of YOLOv5. With this backbone, we observed that the proposed PGPNet can produce predictions with a high level of reliability. All the heights of bins are much closer to values suggested by the perfectly-balanced line compared to Vanilla's result.}

\subsubsection*{Comparison with Existing  Relavant Frameworks}
Our work is the first to leverage an external graph in dealing with the Pill Detection challenge; thus, none of the preceding works are genuinely tight-correlated. Indeed, earlier researches only shared some common ground to our approach: (1) About methodology or (2) about research problem.

\noindent For the first group, there are works that utilized external information to solve the Object Detection problem. We adopt one of the most current studies with this direction - \cite{sgrn} to solve our targeted problem and serve as a baseline for PGPNet. 
Spatial-aware Graph Relation Network (SGRN) \cite{sgrn} is a framework that adaptively discovers and incorporates key semantic and spatial relationships for reasoning over each RoI.

\noindent With respect to research problem, as stated earlier, while there are many works which target single-pill detection problem ~\cite{wong2017development, usuyama2020epillid, ling2020few}, only a few directly solve the task of detecting multiple pills per image \cite{Ou2020, kwon_pill}. We attempt to adopt the most recent technique proposed in \cite{kwon_pill} as another baseline to compare with PGPNet. In the original work, the authors purpose is somewhat different from us, since they attempt to develop a framework which is solely trained on single-pill images, since they argued that the multi-pill dataset would scale up exponentially if the number of pills inscrease. This argument is not held in our intuition, and we believe, in reality, since the pills taken together have to be prescribed by pharmacists. We keep the pipeline as the original work, with some adoption for working with our VAIPE dataset: (1) Change Mask R-CNN to Faster R-CNN; (2) The training single-pill dataset is cropped from our VAIPE dataset with bounding box annonations; (3) The automate data labeling process are skipped. Since the original work did not name the proposed pipeline, we called it as \emph{Kwon's Pipeline} for short.

\textbf{Detection Performance.} 
\begin{table}[t]
\caption{Performance comparison of PGPNet with SGRN and Kwon's Pipeline.\label{tab:eval_related}}
\centering
\setlength\tabcolsep{5pt} 
\begin{tabular}{l|llllll}
\toprule
\textbf{\duy{Model}}           & \textbf{\duy{ mAP}}                                              & \textbf{\duy{ AP50}}                                             & \textbf{\duy{ AP75}}                     & \textbf{\duy{ APs}}                      & \textbf{\duy{ APm}}                                              & \textbf{\duy{ APl}}                      \\ \midrule
\duy{ Faster RCNN}     & \duy{ 63.7}                                             & \duy{ 86.6}                                             & \duy{ 76.9}                     & \duy{ 71.2}                     & \duy{ 58.1}                                             & \duy{ 64.6}                     \\
\duy{ SGRN}            & \duy{ 65.9}                                             & \duy{ 88.8}                                             & \duy{ 79.6}                     & \duy{ 76.3}                     & \duy{ 61.6}                                             & \duy{ 66.3}                     \\
\duy{ Kwon's Pipeline} & \duy{36.2} & \duy{38.5} & \duy{37.2} & \duy{30.3} & \duy{33.1} & \duy{36.0} \\
\duy{ PGPNet}          & \textbf{\duy{ 69.7}}                                             & \textbf{\duy{ 94.4}}                                             & \textbf{\duy{ 83.4}}                     & \textbf{\duy{ 90.0}}                     & \textbf{\duy{ 66.4}}                                             & \textbf{\duy{ 70.1}}                     \\ \bottomrule
\end{tabular}
\end{table}
Table \ref{tab:eval_related} summarizes the comparison of PGPNet, SGRN and Kwon's Pipeline when 
adopting the visual feature extractor architecture from Faster R-CNN with the Resnet-50-FPN model.
Clearly, SGRN outperforms the baseline Faster R-CNN in terms of overall performance but could not outperforms our proposed method PGPNet. 
Specifically, the mAP metrics achieved by SGRN is $65.9$, and PGPNet achieves the better score with a gap of nearly $4$. Upon other metrics, AP50, AP75, APs, APm, and  APl, PGPNet shows its superior by enhancing the performance from $5.1\%$ (e.g., in AP75 metrics) up to $17.1\%$ (e.g., in APs metrics). This is an expected result because SGRN reveals a major weakness when applying to the challenge of Pill Detection. The spatial relationships between pills in an image are arbitrary and frequently changed. Such noisy and unreliable information leads to the performance of SGRN being unstable and sometimes produce not good enough results. In the case of Kwon's Pipeline, the situation is even worse, since it cannot even beat the vanilla one-step Faster RCNN trained with mutple-pill VAIPE training set. The result of this pipeline is $43.1\%$ and $48.2\%$ worse than vanilla Faster R-CNN and PGPNet respectively. One reason for this deficiency is owing to the quality of its training data. There are many circumstances in which overlap or occlusion occurs, which make the cropped images also contain parts of other pills.

\textbf{Pill Classification Accuracy.} 
Figure \ref{fig:reliable_all}\subref{fig:reliable_sgrn} shows the correlation between the confidence and accuracy of PGPNet in the comparison with those of SGRN. Both the frameworks are based on the Faster R-CNN backbone and achieve similar results e.g., an over-confidence trend in every bin. All the predictions with confidence scores smaller than $0.2$ are totally unreliable (with $0$ accuracy). In addition, PGPNet also shows its superior over SGRN in some bins, in which the over-confidence situation is reduced effectively. We do not plot the Confidence-Accuracy of Kwon's Pipeline owing to space constraint and the obvious performance gap compared to our PGPNet.


\subsubsection*{Ability in Dealing with Hard Samples}
\begin{figure*}[!t]
\begin{minipage}{0.25\linewidth}
    \centering \small
        \includegraphics[width=\columnwidth]{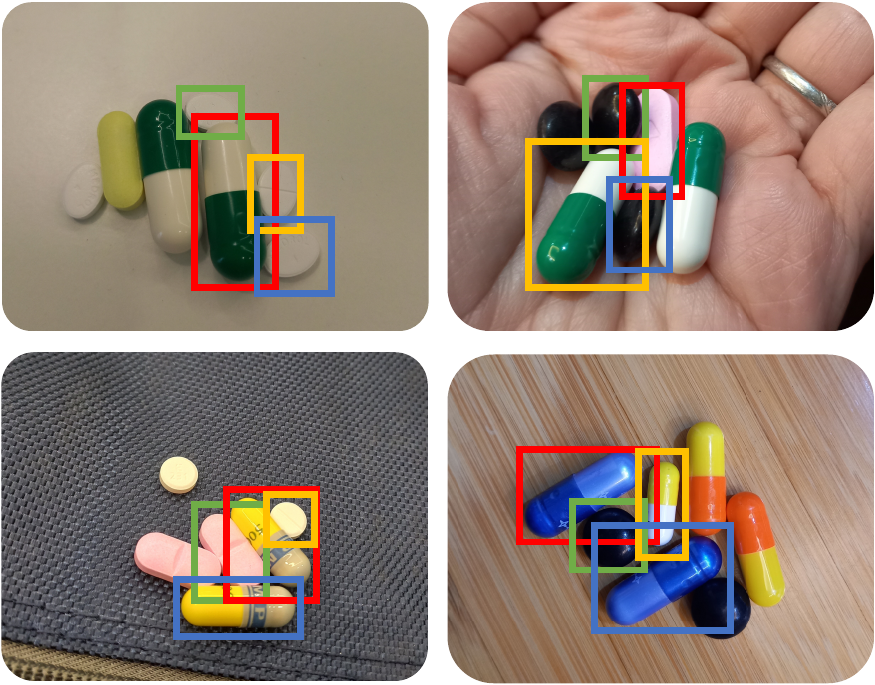}
        \caption{Images with occlusion phenomena in custom occlusion dataset. The rectangles depict examples of tablets with overlapping boundary boxes.
        \label{fig:heavy_occlusion}}
\end{minipage}
\hspace{0.1cm}
\begin{minipage}{0.45\linewidth}
    \captionof{table}{Impact of heavy occlusion images on testing performance of PGPNet and Faster R-CNN.\label{tab:occlusion}}
    \centering
    \setlength\tabcolsep{3pt} 
    \resizebox{\columnwidth}{!}{%
        \begin{tabular}{l|cc||cc}
        \toprule
        \multicolumn{1}{c|}{\begin{tabular}[]{@{}c@{}} Test \\dataset \end{tabular}} & 
        \multicolumn{2}{c}{\begin{tabular}[]{@{}c@{}} Custom \\ Occlusion \end{tabular}} & 
        \multicolumn{2}{c}{\begin{tabular}[]{@{}c@{}} Non-\\Occlusion \end{tabular}} \\
        \midrule
        \multicolumn{1}{c|}{Method} & 
        \begin{tabular}[]{@{}c@{}} Faster \\ R-CNN \end{tabular} &  PGPNet & 
        \begin{tabular}[]{@{}c@{}} Faster \\ R-CNN \end{tabular} & PGPNet \\
        \midrule
        mAP     & 59.2 & \textbf{67.5} &   65.6 &   \textbf{71.7} \\
        AP50    & 76.5 & \textbf{81.1} &   87.4 &   \textbf{92.9} \\
        AP75    & 68.9 & \textbf{76.4} &   80.8 &   \textbf{87.0} \\
        APs     & -    & -             &   80.0 &   \textbf{90.0} \\
        APm     & 61.6 & \textbf{68.3} &   56.5 &   \textbf{64.7} \\
        APl     & 60.7 & \textbf{70.1} &   65.1 &   \textbf{70.6} \\
        \bottomrule
        \end{tabular}
    }
\end{minipage}
\hspace{0.1cm}
\begin{minipage}{0.25\linewidth}
        \centering
        \includegraphics[width=\columnwidth]{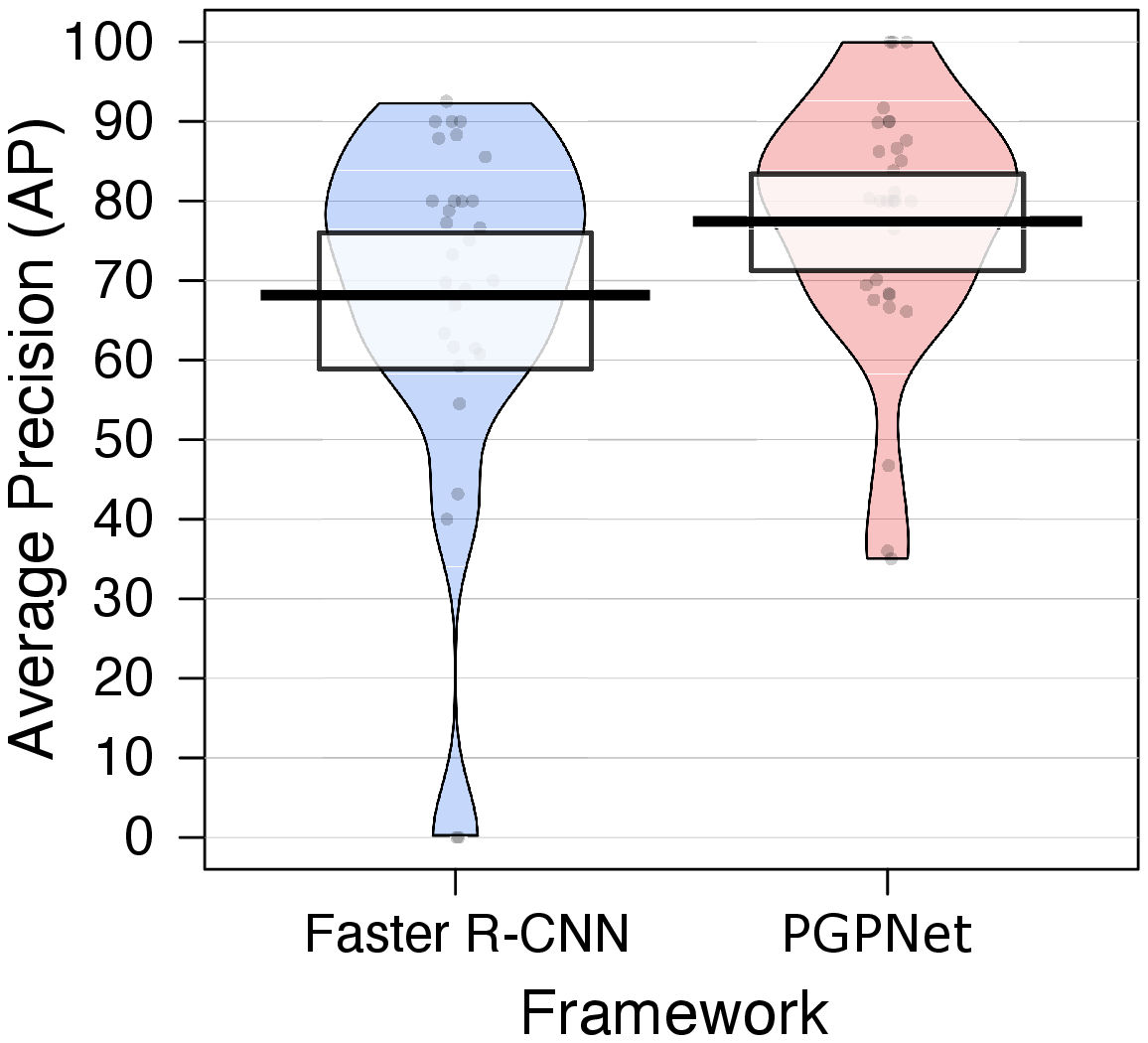}
        \caption{Comparison of PGPNet performance with Faster R-CNN over each individual class in occlusion dataset. \label{fig:occlusion}}
\end{minipage}
\vspace{-0.2cm}
\end{figure*}

\begin{figure}
    \centering
        \includegraphics[width=0.75\columnwidth]{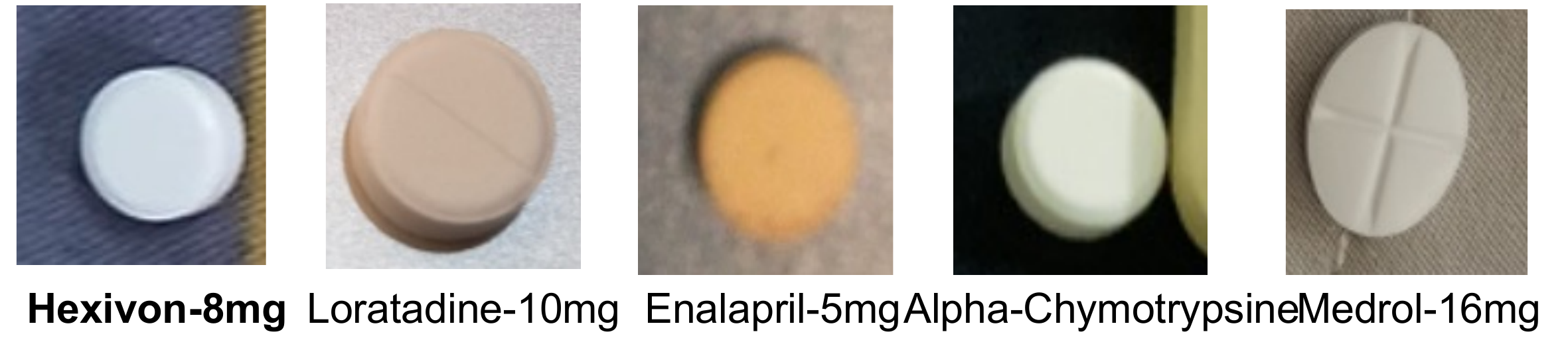}
        \caption{Some sample pills with very identical visual appearance with \emph{Hexinvon-8mg}\label{fig:misclassified}}
\end{figure}

{In the following, we investigate the ability of PGPNet in dealing with the occlusion phenomenon caused by overlapping pills, which is one of the most critical issues in dealing with multi-pill detection. 
To this end, we create a so-call \emph{custom occlusion sub-dataset} of VAIPE, which contains images with heavy occlusion phenomena, i.e., having at least two RoIs with the IoU beyond 30\% (Fig.\ref{fig:heavy_occlusion}).  We also create 
a custom \emph{custom non-occlusion sub-dataset}
which contains samples that are in the same classes that appear in the \emph{custom occlusion sub-dataset} but with no occlusion. 
The quantitative result is summarized in Table \ref{tab:occlusion}. 
The (-) mark in the table suggests the disregarded or unavailable metrics.
As the numbers suggest, even in cases where heavy occlusion occurs, PGPNet still shows its superior over Faster R-CNN. Specifically, the mAP over all classes in the \emph{custom occlusion sub-dataset} suggests a gap of $8.3\%$ between the two approaches. Interestingly, with the aid of classifier weight as the distinguishing characteristic for each class, PGPNet, even when dealing with occlusion cases still enhances the performance of $1.9\%$ compared to Faster R-CNN handling  the non-occlusion case (e,g, $67.5$ vs. $65.6$, respectively). Figure \ref{fig:occlusion} provides more information about the AP for each class in the \emph{custom occlusion sub-dataset}. PGPNet still outperforms Faster R-CNN in most cases with a large gap, and also produces a more reliable result by introducing a smaller variance over the AP metrics. }

\subsection*{PGPNet's Explainability}

\begin{figure*}[!t]
\begin{minipage}{0.32\linewidth}
        \centering
        \subfloat[Faster R-CNN]{ %
            \includegraphics[width=0.43\columnwidth]{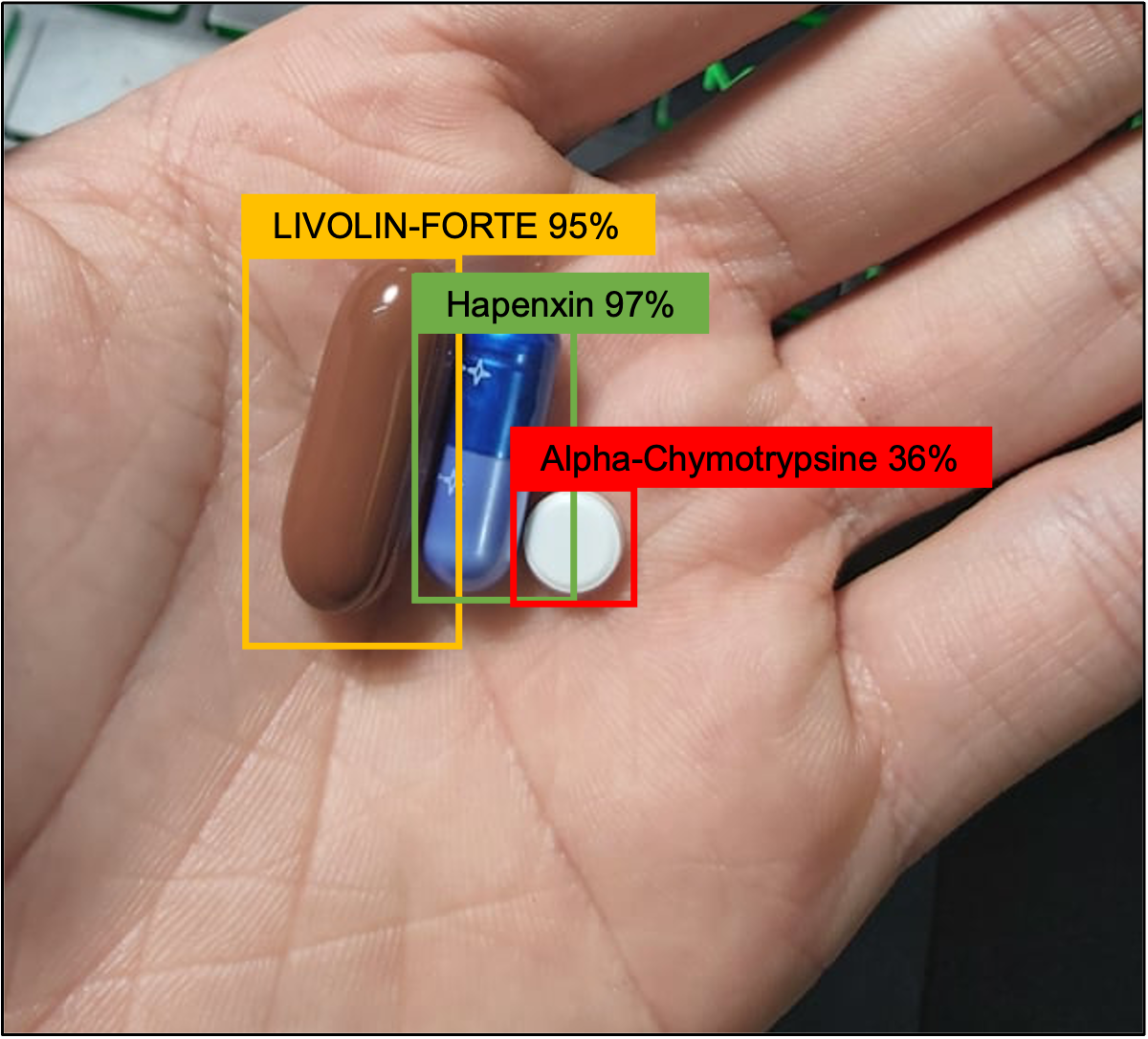}
            \label{fig:eval_rcnn}
        }
        \subfloat[PGPNet]{ %
            \includegraphics[width=0.43\columnwidth]{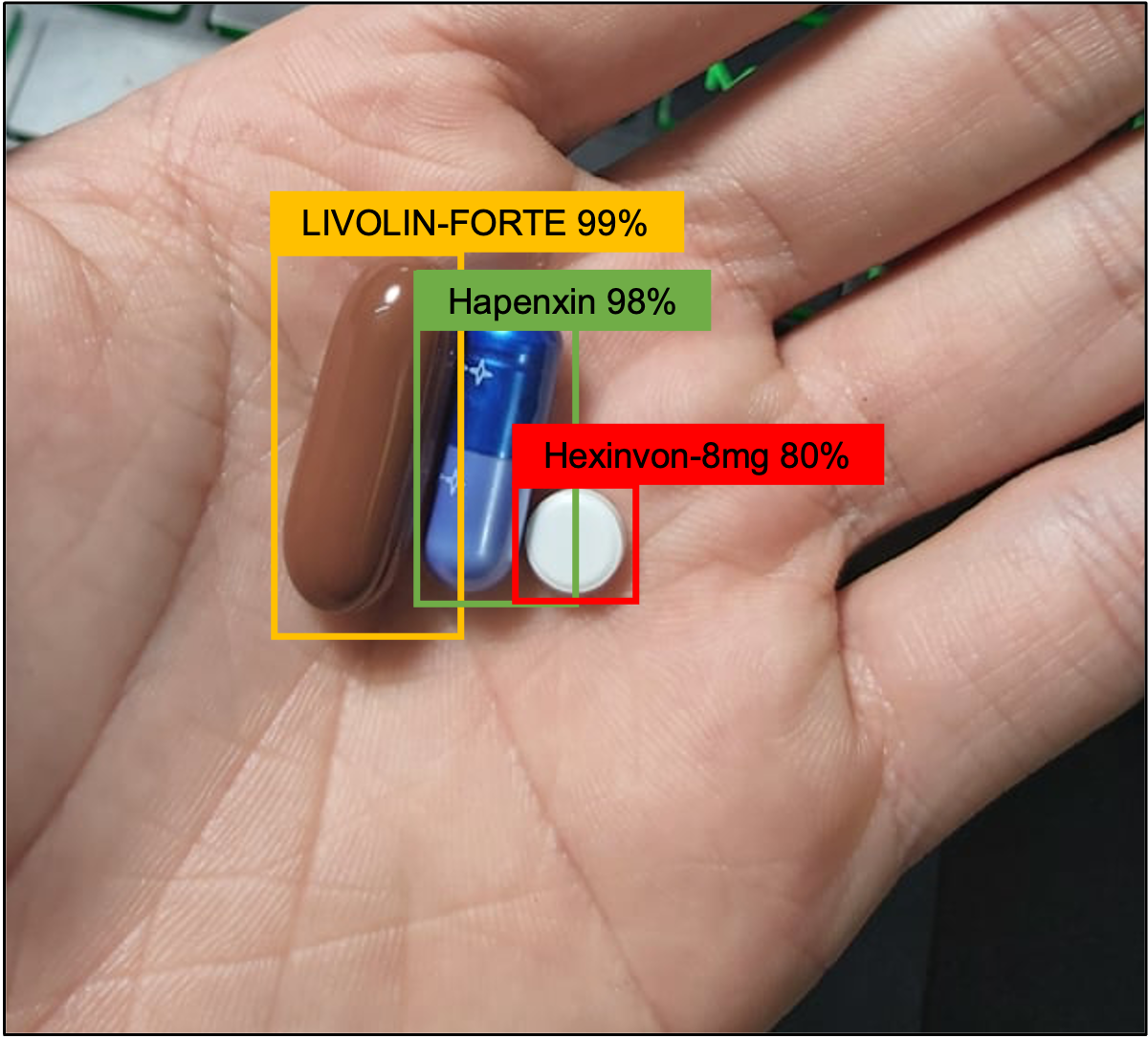}
            \label{fig:eval_kgp}
        }
        \caption[Predictions for a hard sample made by Faster R-CNN and PGPNet given the same image.]{Predictions for a hard sample made by Faster R-CNN and PGPNet given the same image.}
        \label{fig:kg_robust_instance}
        \vspace{0.1cm}
\end{minipage}
\hspace{0.1cm}
\begin{minipage}{0.64\linewidth}
        \centering \small
        \subfloat[Input]{ %
            \includegraphics[width=0.21\columnwidth,valign=c]{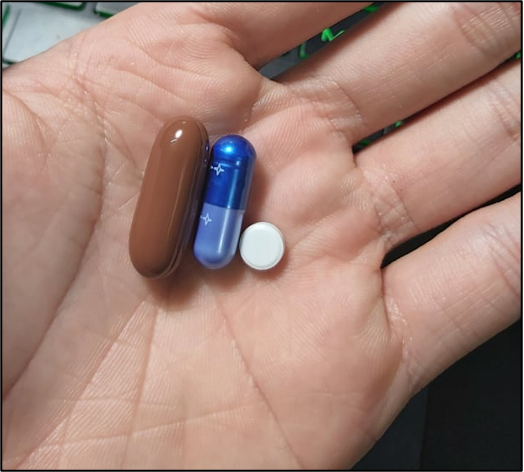}
            \label{fig:original}
        }
        \hspace{-0.15em}%
        \centering
        \subfloat[LIVOLIN-FORTE]{ %
            \includegraphics[width=0.21\columnwidth,valign=c]{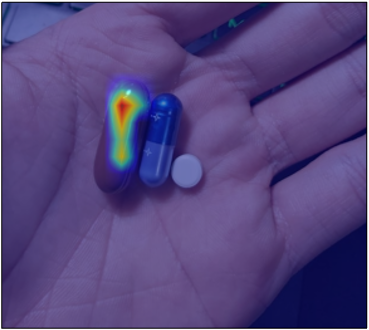}
            \label{fig:exp_1}
        }
        \hspace{-0.15em}%
        \subfloat[Hapenxin]{ %
            \includegraphics[width=0.21\columnwidth,valign=c]{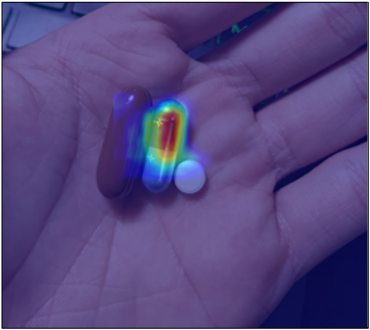}
            \label{fig:exp_2}
        }
        \hspace{-0.15em}%
        \subfloat[Hexinvon-8mg]{ %
            \includegraphics[width=0.21\columnwidth,valign=c]{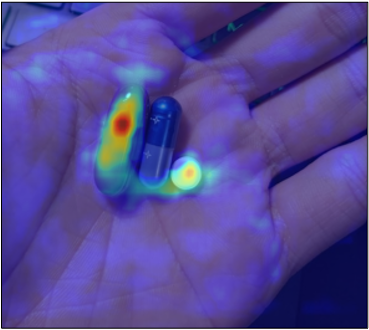}
            \label{fig:exp_3}
        }
        \caption{The saliency maps for each of the groundtruth labels included in the image instance. For simple samples (LIVOLIN-FORTE and Hapenxin), the classifier focuses on the exact location of the tablets to determine their identity. In contrast, for the hard case (Hexivon-8mg), information on both Hexivon and LIVOLIN-FORTE served as evidence. \label{fig:XAI_heatmap} }
\end{minipage}
\vspace{-0.5cm}
\end{figure*}


This section is dedicated to analyzing the results produced by PGPNet through a specific sample. This example demonstrates that the operation of PGPNet is very congruent with our initial motivation and that our designed architecture can materialize this motivation.



\subsubsection*{Experiment settings.} 
In this experiment, we choose a hard sample, namely \emph{Hexinvon-8mg}, with a relatively common appearance, for investigation. Figure \ref{fig:misclassified} visualizes \emph{Hexinvon-8mg} together with other pills in our dataset with almost identical visual appearance (round shape, white tint, etc.). 
As illustrated, these pills are readily confused with \emph{Hexinvon-8mg}.
Indeed, Fig.~\ref{fig:kg_robust_instance} depicts an example in which  \emph{Hexinvon-8mg} is miscategorized as \emph{Alpha-Chymotrypsine} by Faster R-CNN. 
Our PGPNet can, however, successfully distinguish \emph{Hexinvon-8mg} with a high confidence score.
In the following, we applied several Explainable AI techniques to explain the results inferred by our PGPNet. The image of interest consists of three pills: LIVOLIN-FORTE,  Hapenxin, and Hexivon as shown in Fig.~\ref{fig:kg_robust_instance}. 



\subsubsection*{Explanation of the Prediction Results}
We adopt the Excitation Backpropagation technique proposed by Zhang \cite{exb} to construct the saliency maps (Fig.\ref{fig:XAI_heatmap}), which indicate what the classifier has learned to produce the final results.
Firstly, for the easy samples, i.e., LIVOLIN-FORTE and Hapenxin, our model focuses precisely on those pill regions to make the prediction decision. 
In contrast, in the case of the hard sample, i.e., \emph{Hexinvon-8mg}, however, two regions are highlighted: one at the position of \emph{Hexinvon-8mg} and the other at the location of LIVOLIN-FORTE.
It indicates that the classifier solely requires information about LIVOLIN-FORTE and Hapenxin to identify these pills. Nevertheless, for \emph{Hexinvon-8mg}, the classifier must additionally incorporate information about its neighbor, i.e., LIVOLIN-FORTE.
This hypothesis is also supported by the Probabilistic score matrix shown in Fig.~\ref{fig:pseudo_scores}.
The probabilistic score matrix represents the prediction results generated by our Pseudo Classifier, which relies mainly on the pill's visual characteristics.
As demonstrated, Pseudo Classifier can accurately detect the proper labels of two simple samples, with their prediction scores approaching $1$,  and boost up their neighbors' probabilities (label ID $7$, $17$, etc.). However, with the case of \emph{Hexinvon-8mg}, the probability scores are relatively low, with all RoIs being investigated achieving scores of only about $0.3$.

Now, we utilize another explainable AI technique named GNNExplainer \cite{gnn_explainer} to investigate further the reason for identifying the hard sample, \emph{Hexinvon-8mg}. GNNExplainer is a model-agnostic architecture that can provide interpretable explanations for predictions of graph-based models. Specifically, GNNExplainer may identify a subgraph and a subset of node features that have a significant role in the prediction outcomes.
In our experiment, we treat our Graph Transformer Network as a module that produces regression output, i.e., the context vectors corresponding to all RoIs. For a more comprehensible result, we set the number of RoIs selected from the RPN module to ten, consisting of the five RoIs with the greatest \emph{objectness} scores and the other five with the lowest score. 
We utilize GNNExplainer to identify the sub-graph that contributes the most in recognizing \emph{Hexinvon-8mg}. 
The results are demonstrated in Fig.~\ref{fig:gnn_explain}. 
In this figure, the white box depicts the RoI of \emph{Hexinvon-8mg}, the two orange boxes and blue boxes represent the RoIs of LIVOLIN-FORTE, and Hapenxin, respectively, while the five gray boxes indicate the RoIs of noise.
The black edges represent the vital connections, whose weights are proportionate to the width of the edges.
First, there are almost no edges between the nodes representing \emph{Hexinvon-8mg} and those of the noise RoIs.
It implies that the noise RoIs do not cue the prediction of \emph{Hexinvon-8mg}.
In contrast, there are bolded linkages between the RoIs of LIVOLIN-FORTE, Hapenxin, and \emph{Hexinvon-8mg}. 
These findings, along with the saliency map (Fig.\ref{fig:XAI_heatmap}), interpret that PGPNet has learned both the visual characteristic of the pill itself and the relationship between that pill and the others to make the final decision.

\begin{figure*}[!t]
\begin{minipage}{0.4\linewidth}
        \centering
        \includegraphics[width=0.9\columnwidth]{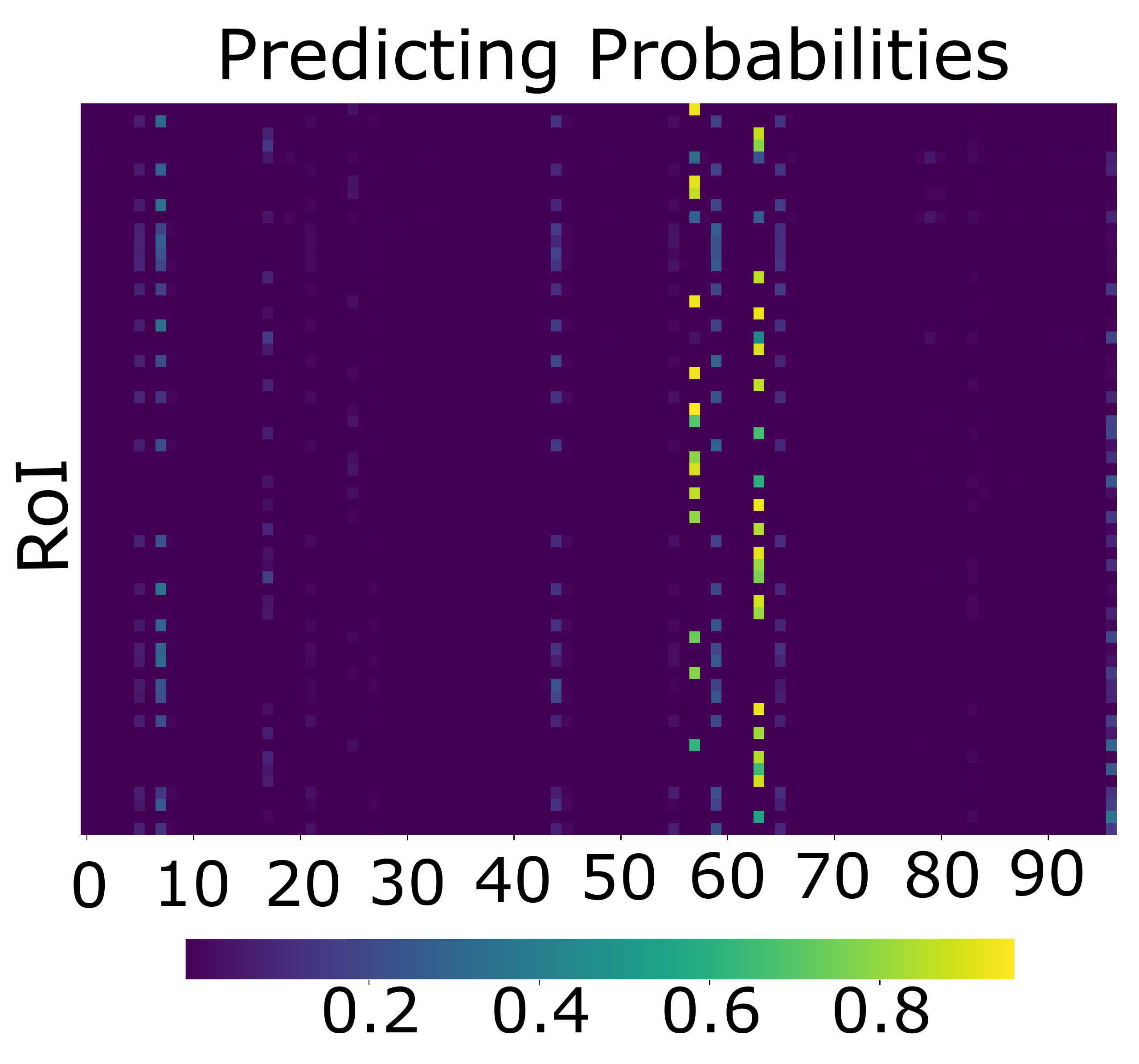}
        \caption{Probabilistic scores produced by PGPNet's Pseudo Classifier. \label{fig:pseudo_scores}}
\end{minipage}
\hspace{0.1cm}
\begin{minipage}{0.55\linewidth}
        \centering \small
        \subfloat[Bounding boxes of the RoIs.]{ %
            \includegraphics[width=0.4\columnwidth,valign=c]{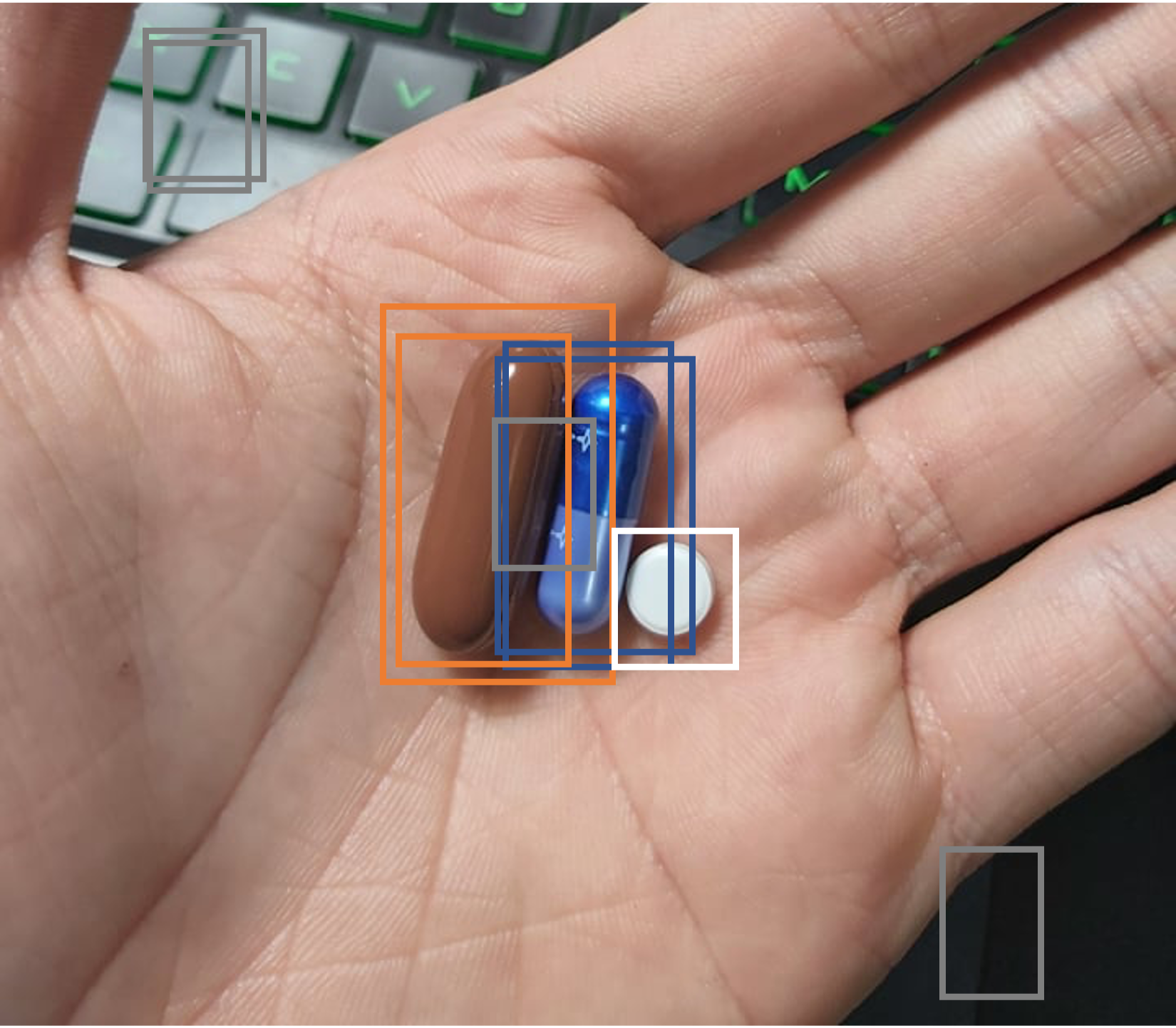}
            \label{fig:rpn_choose10}
        }
        \hfill%
        \centering
        \subfloat[Sub-graph identified by GNNExplainer.]{ %
            \includegraphics[width=0.5\columnwidth,valign=c]{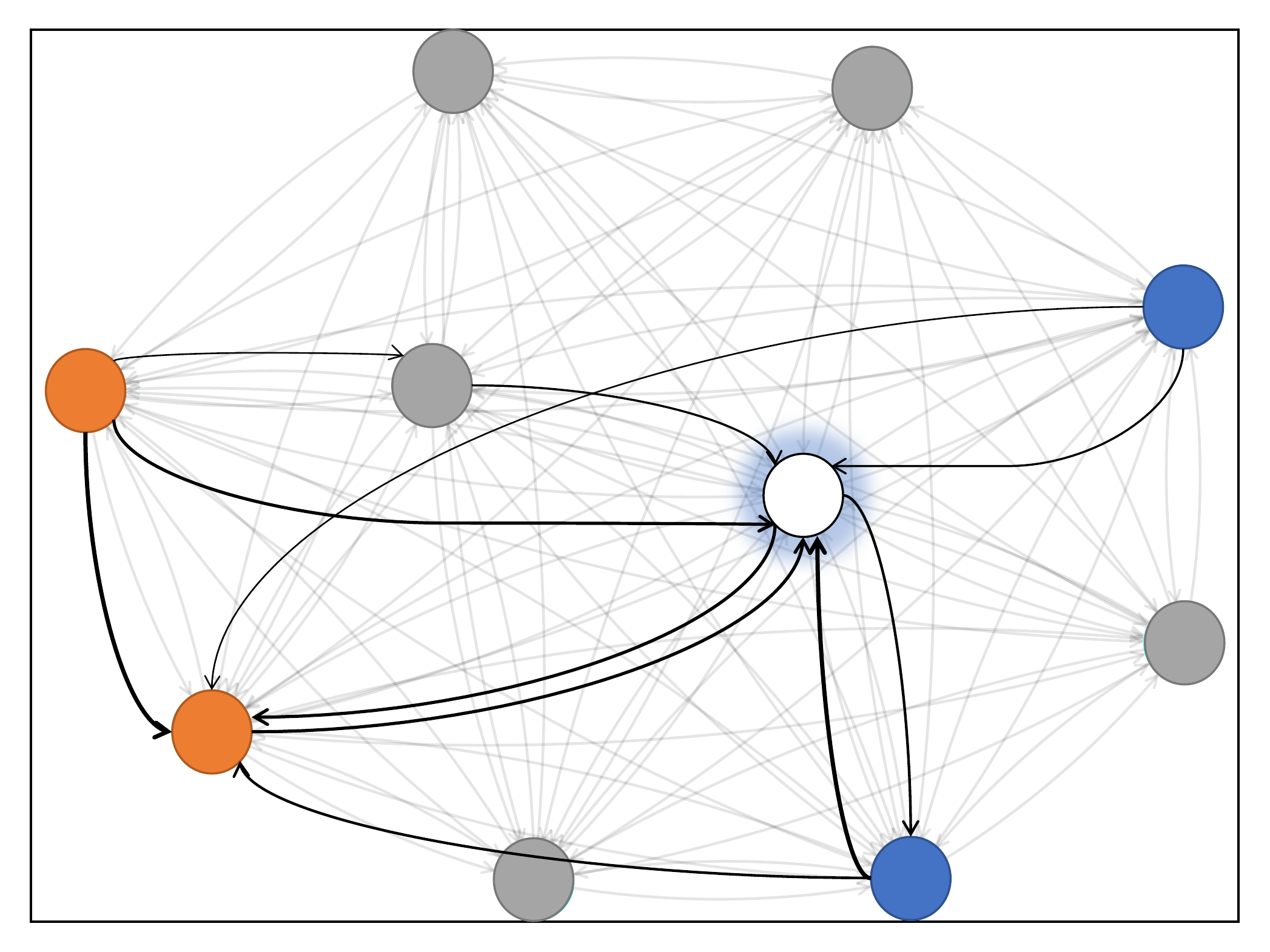}
            \label{fig:gnn_saliency}
        }
         \caption{Interpretation of the prediction result for using GNNExplainer.} 
    
        
   
\end{minipage}
\vspace{-0.3cm}
\end{figure*}

\begin{figure}[!tb]
        \centering
        \includegraphics[width=0.5\columnwidth]{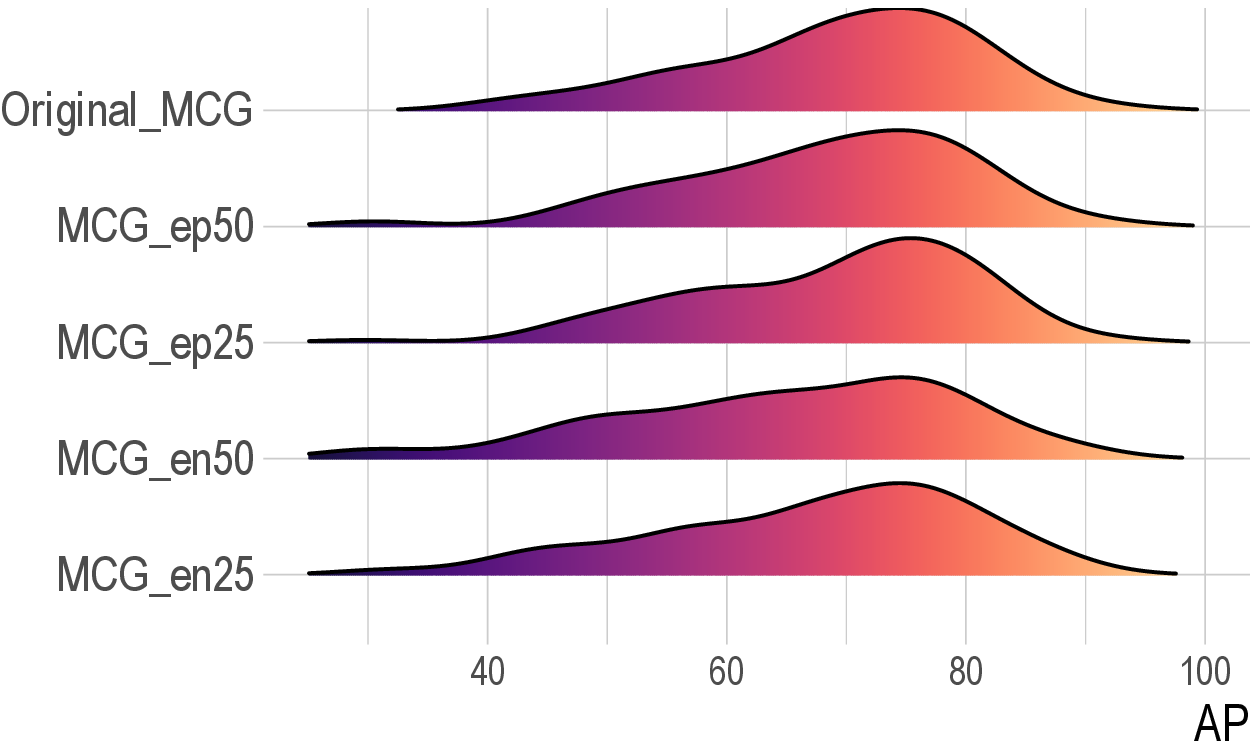}
       \caption{Distributions of Average Precision recorded over all classes produced by PGPNet with different MCG versions. \label{fig:edge_modify}}
    \end{figure}

    \begin{figure}[!tb]
        \centering
        \includegraphics[width=0.5\columnwidth]{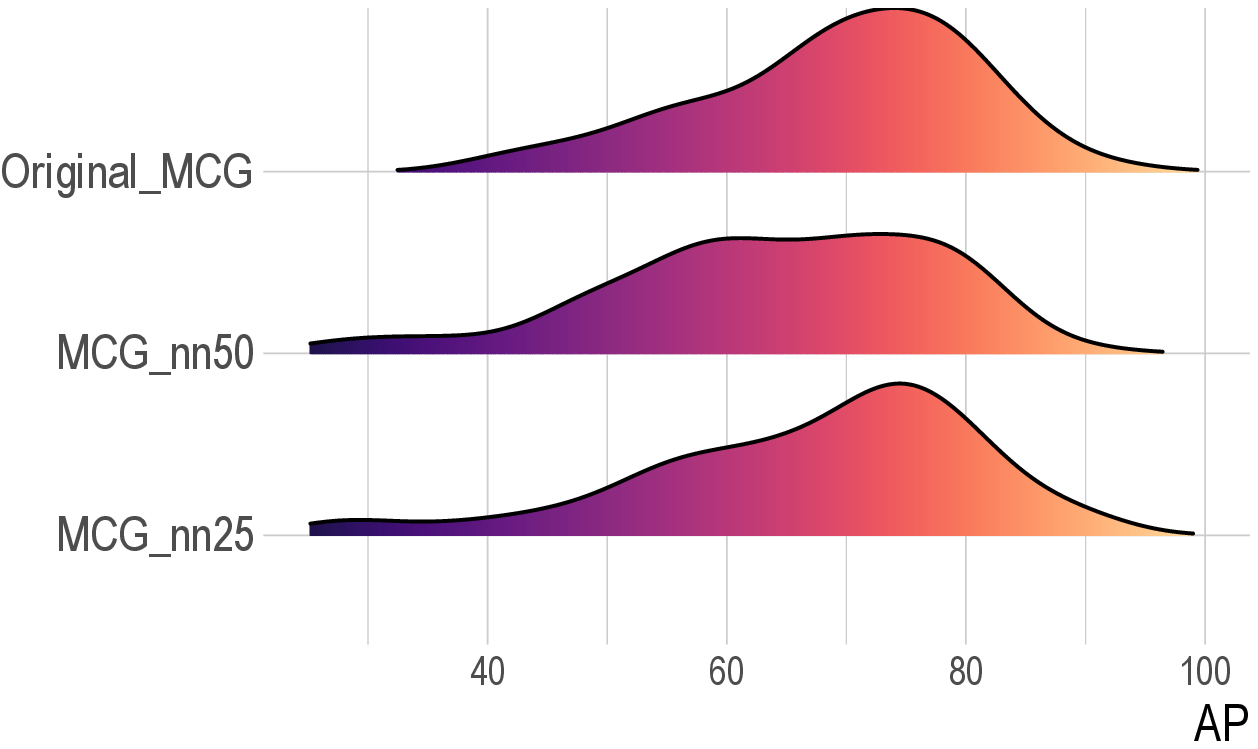}
       \caption{Distributions of Average Precision recorded over the classes in $N_A$ set produced by PGPNet with different MCG versions. \label{fig:node_modify}}
    \end{figure}

\subsection*{Ablation Studies}
In this section, we perform extensive ablation studies to investigate the impacts of the main techniques proposed in our PGPNet and to investigate how each component in the proposed method helps to improve learning performance. Specifically, we alter the Co-occurrence Graph and observe how it affects the detection results in Section \nameref{sec:ablation_mcg}. We then assess the effects of using the relational graphs, the Graph transformer network, and the proposed auxiliary loss in Sections \nameref{sec:ablation_relational graphs}, \nameref{sec:ablation_gtn}, 
respectively.

\subsubsection*{Effect of Co-occurrence Graph's Quality}
\label{sec:ablation_mcg}
In this section, we perform two experiments to observe how the performance is changed when the nodes set and edges set of MCG are modified respectively.

\textbf{Edge Set Modification.}
We first observe the behavior of our PGPNet when adding noise edges and removing actual edges. We set up four scenarios which are the combinations of removing $25\%$ and $50\%$ of the edges in the set $E_1$, and adding a number of synthesized edges corresponding to $25\%$ and $50\%$ of the cardinality of $E_1$.

Figure \ref{fig:edge_modify} illustrates the performances of PGPNet with all Medical Co-occurrence Graph variances when being put into comparison with the original one. The performance here is denoted by the general metrics AP. As indicated by AP density, PGPNet with original MCG generates a more concentrated density with a smaller variance and a higher mean than other variances. In addition, when $50\%$ of edges are eliminated, the performance is clearly inferior to when $25\%$ of edges are eliminated. The figure concludes with the intriguing observation that eliminating edges at random would result in a greater performance decrease than adding noisy edges. This is because, even with the addition of noisy edges, PGPNet could still filter out unnecessary information through the training process. When excluding edges, the situation is different because the framework cannot learn the external knowledge contained in the eliminated edges.

    

\begin{table*}[!t]
    \captionof{table}{Performance of PGPNet with the diferent combination of its components, i.e., when removing (marked as $\times$) / keeping (marked as \checkmark) the relational graph, GTN and auxiliary loss. Numbers inside the (.) represent the gap in percentage compared to the full version of PGPNet.\label{tab:ablation_result}}
    \centering
    \small
    \resizebox{\columnwidth}{!}{%
    \begin{tabular}{l|ccccc|l|l|l|l|l|l}
    \toprule
    \multicolumn{1}{l|}{} & \multicolumn{5}{c|}{\textbf{Component}}  & \multicolumn{6}{c}{\textbf{Performance}}
    \\
    & $\mathcal{G}_c$ & $\mathcal{G}_s$ & $\mathcal{G}_v$ & GTN & $\mathcal{L}_{aux}$ &   mAP & AP50 & AP75 & APs & APm & APl
    \\ \midrule
    Faster R-CNN & $\times$  & $\times$   & $\times$   & $\times$    & $\times$    &
      \begin{tabular}[c]{@{}r@{}}63.7 (-8.6)\end{tabular} &
      \begin{tabular}[c]{@{}r@{}}86.7 (-8.4)\end{tabular} &
      \begin{tabular}[c]{@{}r@{}}76.9 (-7.9)\end{tabular} &
      \begin{tabular}[c]{@{}r@{}}71.3 (-20.8)\end{tabular} &
      \begin{tabular}[c]{@{}r@{}}58.1 (-10.6)\end{tabular} &
      \begin{tabular}[c]{@{}r@{}}64.6 (-7.8)\end{tabular}  \\
    PGPNet-v1    & \checkmark  & $\times$   & $\times$   & $\times$    & $\times$     &
      \begin{tabular}[c]{@{}r@{}}65.9 (-5.5)\end{tabular} &
      \begin{tabular}[c]{@{}r@{}}91.9 (-2.9)\end{tabular} &
      \begin{tabular}[c]{@{}r@{}}79.6 (-4.6)\end{tabular} &
      \begin{tabular}[c]{@{}r@{}}72.5 (-19.4)\end{tabular} &
      \begin{tabular}[c]{@{}r@{}}62.3 (-4.3)\end{tabular} &
      \begin{tabular}[c]{@{}r@{}}66.0 (-5.8)\end{tabular}  \\
    PGPNet-v2    & \checkmark  & $\times$  & \checkmark  & \checkmark   & \checkmark   &
      \begin{tabular}[c]{@{}r@{}}66.9 (-3.9)\end{tabular} &
      \begin{tabular}[c]{@{}r@{}}92.1 (-2.7)\end{tabular} &
      \begin{tabular}[c]{@{}r@{}}81.1 (-2.8)\end{tabular} &
      \begin{tabular}[c]{@{}r@{}}80.0 (-11.1)\end{tabular} &
      \begin{tabular}[c]{@{}r@{}}61.3 (-5.9)\end{tabular} &
      \begin{tabular}[c]{@{}r@{}}67.6 (-3.6)\end{tabular} \\
    PGPNet-v3    & \checkmark  & \checkmark  & $\times$  & \checkmark   & \checkmark   &
      \begin{tabular}[c]{@{}r@{}}67.8 (-2.8)\end{tabular} &
      \begin{tabular}[c]{@{}r@{}}92.9 (-1.9)\end{tabular} &
      \begin{tabular}[c]{@{}r@{}}82.3 (-1.5)\end{tabular} &
      \begin{tabular}[c]{@{}r@{}}82.5 (-8.3)\end{tabular} &
      \begin{tabular}[c]{@{}r@{}}62.7 (-3.5)\end{tabular} &
      \begin{tabular}[c]{@{}r@{}}68.1 (-2.8)\end{tabular} \\
    PGPNet-v4    & \checkmark  & \checkmark  & \checkmark  & $\times$   & \checkmark  &
      \begin{tabular}[c]{@{}r@{}}68.4 (-1.9)\end{tabular} &
      \begin{tabular}[c]{@{}r@{}}92.6 (-2.2)\end{tabular} &
      \begin{tabular}[c]{@{}r@{}}81.7 (-2.1)\end{tabular} &
      \begin{tabular}[c]{@{}r@{}}80.0 (-11.1)\end{tabular} &
      \begin{tabular}[c]{@{}r@{}}64.4 (-1.0)\end{tabular} &
      \begin{tabular}[c]{@{}r@{}}68.8 (-1.8)\end{tabular}  \\
    PGPNet-v5    & \checkmark  & \checkmark & \checkmark  & \checkmark   & $\times$  &
      \begin{tabular}[c]{@{}r@{}}67.2 (-3.6)\end{tabular} &
      \begin{tabular}[c]{@{}r@{}}91.3 (-3.5)\end{tabular} &
      \begin{tabular}[c]{@{}r@{}}80.9 (-3.1)\end{tabular} &
      90.0 (+0.0) &
      \begin{tabular}[c]{@{}r@{}}62.2 (-4.3)\end{tabular} &
      \begin{tabular}[c]{@{}r@{}}67.3 (-3.9)\end{tabular}  \\
    PGPNet    & \checkmark  & \checkmark  & \checkmark  & \checkmark   & \checkmark   &
      69.7 &
      94.7 &
      83.5 &
      90.0 &
      65.0 &
      70.0 \\ \bottomrule
    \end{tabular}
    }
    \end{table*}

\textbf{Node Set Modification.}
To observe PGPNet's performance when the Medical Co-occurrence Graph lacks information on some specific nodes - classes, we design two different scenarios. In the first one, $25\%$ nodes are removed in the original graph, this set is denoted as $N_A$. For the latter, $50\%$ of nodes are eliminated, and the corresponding set $N_B$ is ensured to be a superset of $N_A$. The performances of PGPNet in two circumstances are compared with itself when having the full MCG, considering only the classes appeared in the set $N_A$.

Figure \ref{fig:node_modify} depicts the outcome of this experiment. The AP across all $N_A$ classes is used to evaluate performance here. As indicated by the graph, node removals also result in a significant decrease in model performance. More interestingly, the more nodes being eliminated, the greater drop is captured. Specifically, the AP density in case MCG contains only $50\%$ of remaining nodes has a great variance, with the mean value only around $60\%$. 

In the following, we study the effectiveness of the relational graphs, Graph Transformer Network (GTN) block, and auxiliary loss. 
The detailed configurations are presented in Table \ref{tab:ablation_result}. The $+$ sign indicates the presence of a component in a specific version, while $-$ denotes the opposite.  

\subsubsection*{Effects of the Relational Graphs}
\label{sec:ablation_relational graphs}
In this section, we study the effectiveness of the Size-graph and visual-based graph. 
To this end, we implement two simplified versions of PGPNet, namely PGPNet-v2 and  PGPNet-v3, in which we remove the Size-graph and visual-based graph, respectively. 
As shown in Table \ref{tab:ablation_result}, eliminating the Size-graph causes a decrease in performance from 3.9\% to 11.1\%, while omitting the visual-based graph reduces the accuracy from 2.8\% to 8.3\%.
An interesting finding is that the deterioration gap when removing the size graph is more significant than those when eliminating the visual-based graph in terms of all evaluation metrics. 
These findings imply the effectiveness of the Size-graph over the visual-based graph.
Moreover, it can be observed that mAP is the most impacted when the relational graphs are removed, followed by AP50, when comparing mAP, AP50, and AP75.
This can be explained as follows.
In AP75, we measure the precision of RoIs with the IoU beyond 75\%, which presumably has a high degree of confidence regarding the objective.
In contrast, when we reduce the IoU threshold, such as AP50 and mAP, the overlap area of the objective drops, resulting in a model with a significant degree of uncertainty.
In this case, integrating relational graphs provides additional data that reduces uncertainty, thereby boosting detection accuracy.
\subsubsection*{Effects of the Multi-modal Data Fusion Block and Auxiliary Loss}
\label{sec:ablation_gtn}
To investigate the effectiveness of the GTN, we implement PGPNet-v4, omitting the GTN block and relying solely on the GCN to learn the node representation. 
Results in Table~\ref{tab:ablation_result} reveal that GTN enhances the model's accuracy from $1.0\%$ to $11.1\%$. 
Comparing mAP, AP50, and AP75, AP50, and AP75 are slightly more influenced by GTN than mAP, but the gaps are trivial.
We employ PGPNet-v5, which eliminates the proposed auxiliary loss and compare its performance with the original PGPNet. 
As illustrated in Table \ref{tab:ablation_result}, adopting our auxiliary loss may result in a 3 to 4 percent performance gain for most evaluation metrics. 
In the final ablation study, we implement PGPNet-v1, which retains only the co-occurrence graph and removes all the other components.
As depicted in Table \ref{tab:ablation_result}, the detection accuracy degrades significantly, with a gap ranging from $2.9\%$ to $19.4\%$. However, even with this version, PGPNet is still superior to Faster RCNN, with a performance margin of up to $7.1\%$.

In conclusion, the PGPNet version with all components exhibits its superiority in all evaluation metrics. In addition, all versions of PGPNet are superior to the Faster R-CNN backbone, demonstrating the contribution of each component to the overall performance of PGPNet.

\section*{Conclusion}
\label{sec:conclusion}
\textbf{Contributions.} We proposed PGPNet, a reliable and explainable pill detection framework in real-world settings. To deal with hard samples, PGPNet leveraged external knowledge, including co-occurrence likelihood, relative pill size, and visual semantic correlation during the training process. We implemented PGPNet into two popular object detectors and evaluated the proposed method on a real-world multiple pill detection dataset. The experimental results demonstrated that it could improve these models by considerable margins. Moreover, our comprehensive ablation studies proved the robustness, reliability, and explainability of the proposed framework. Future work will aim to evaluate PGPNet under a federated learning setting as well as other image-text understanding datasets.

\noindent \textbf{Limitations and Future Works.} Although the effectiveness of PGPNet largely relies on external information from a graph, we recognize that this external knowledge may not always be practical or feasible in all hospitals and locations. Additionally, our co-occurrence graph building process currently relies on prescription data tied to the VAIPE dataset. However, in order for PGPNet to be suitable for practical settings, this graph needs to be expanded to include more pills and relationships, which may make it impractical to deploy the framework to actual devices and applications. To address this, we plan to collect more prescriptions and construct more general medical knowledge graphs in our future work. Additionally, we plan to optimize the computational requirements when scaling up our framework.

\section*{Acknowledgments}

This work was funded by Vingroup Joint Stock Company and supported by Vingroup Innovation Foundation (VINIF) under project code $VINIF.2021.DA00128$. We thank all our collaborators who participated in the collection and annotation of the VAIPE dataset.

\bibliographystyle{abbrv}
\bibliography{references}
\end{document}